\documentclass[10pt,twocolumn,letterpaper]{article}

\usepackage{cvpr}              

\usepackage[dvipsnames]{xcolor}

\definecolor{cvprblue}{rgb}{0.21,0.49,0.74}
\usepackage[pagebackref,breaklinks,colorlinks,citecolor=cvprblue]{hyperref}

\def\paperID{16418} 
\def\confName{CVPR}
\def\confYear{2025}
\newcommand{\indep}{\perp \!\!\! \perp}
\title{Causal Composition Diffusion Model for Closed-loop Traffic Generation}

\author{First Author\\
Institution1\\
Institution1 address\\
{\tt\small firstauthor@i1.org}
\and
Second Author\\
Institution2\\
First line of institution2 address\\
{\tt\small secondauthor@i2.org}
}

\DeclareMathAlphabet\mathbfcal{OMS}{cmsy}{b}{n}

\usepackage{pifont}

\usepackage[utf8]{inputenc} 
\usepackage[T1]{fontenc}    
\usepackage{hyperref}       
\usepackage{url}            
\usepackage{booktabs}       
\usepackage{amsfonts}       
\usepackage{nicefrac}       
\usepackage{microtype}      
\usepackage{xcolor}         
\usepackage{amsthm}
\usepackage{thmtools}
\usepackage{bbm}
\usepackage{mathtools}
\usepackage{algorithm}
\usepackage{algpseudocode}
\usepackage{multirow}
\usepackage{hhline}
\usepackage{xcolor,colortbl}
\usepackage{caption}
\usepackage{subcaption}
\usepackage{wrapfig}
\usepackage{enumitem,kantlipsum}
\usepackage{amssymb}
\usepackage{subcaption}

\usepackage{amsmath,amsfonts,bm}
\usepackage{xspace}

\def\eqref#1{equation~(\ref{#1})}

\def\1{\bm{1}}

\mathchardef\mhyphen="2D

\DeclareMathAlphabet{\mathsfit}{\encodingdefault}{\sfdefault}{m}{sl}
\SetMathAlphabet{\mathsfit}{bold}{\encodingdefault}{\sfdefault}{bx}{n}

\DeclareMathOperator*{\argmax}{arg\,max}

\usepackage{fontawesome}
\usepackage{threeparttable} 

\usepackage{bbding}

\theoremstyle{plain}

\theoremstyle{definition}
\newtheorem{definition}{Definition}

\theoremstyle{remark}

\usepackage{graphicx}

\newcommand{\method}{\textit{CCDiff}}

\author{%
  Haohong Lin${}^1$\thanks{This work was done when Haohong was an intern at Cruise LLC.}\ , Xin Huang${}^2$, Tung Phan${}^2$, David Hayden${}^2$, \\
  Huan Zhang${}^3$, Ding Zhao${}^1$, Siddhartha Srinivasa${}^2$, Eric Wolff${}^2$, Hongge Chen${}^2$ \\
  ${}^1$CMU, ${}^2$Cruise LLC, ${}^3$UIUC\\
  \small{\texttt{haohongl@cmu.edu}, \texttt{hongge.chen@getcruise.com}} \\
}

\begin{document}

\maketitle

\vspace{-8mm}
\begin{abstract}
\noindent Simulation is critical for safety evaluation in autonomous driving, particularly in capturing complex interactive behaviors. 
However, generating \textbf{realistic} and \textbf{controllable} traffic scenarios in long-tail situations remains a significant challenge. 
Existing generative models suffer from the conflicting objective between user-defined controllability and realism constraints, which is amplified in safety-critical contexts. 
In this work, we introduce the \textbf{C}ausal \textbf{C}ompositional \textbf{Diff}usion Model~(\textbf{\method}), a structure-guided diffusion framework to address these challenges. 
We first formulate the learning of controllable and realistic closed-loop simulation as a constrained optimization problem. 
Then, CCDiff maximizes controllability while adhering to realism by automatically identifying and injecting causal structures directly into the diffusion process, providing structured guidance to enhance both realism and controllability. 
Through rigorous evaluations on benchmark datasets and in a closed-loop simulator, CCDiff demonstrates substantial gains over state-of-the-art approaches in generating realistic and user-preferred trajectories. 
Our results show CCDiff’s effectiveness in extracting and leveraging causal structures, showing improved closed-loop performance based on key metrics such as collision rate, off-road rate, FDE, and comfort. For more details, welcome to check our~\href{https://sites.google.com/view/ccdiff/}{project website}. 
\end{abstract}
\vspace{-5mm}

\section{Introduction}
Reliable closed-loop traffic simulation is essential for assessing autonomous vehicle (AV) safety in diverse and complex scenarios~\cite{feng2023dense, ding2023survey, zhong2023guided, zhong2023language, ding2023realgen}. 
Simulations must be both \textit{realistic}, capturing the intricacies of real-world driving, and \textit{controllable}, allowing customization aligned with user preferences. 
However, balancing realism with controllability remains a significant challenge. 
Previous works often prioritize one aspect, optimizing either realism or user-specified objectives~\cite{zhong2023guided, ding2023realgen}. How to jointly achieve both objectives under safety-critical conditions remains fruitful yet unresolved. 

\begin{figure}
    \centering
    \includegraphics[width=1.0\linewidth]{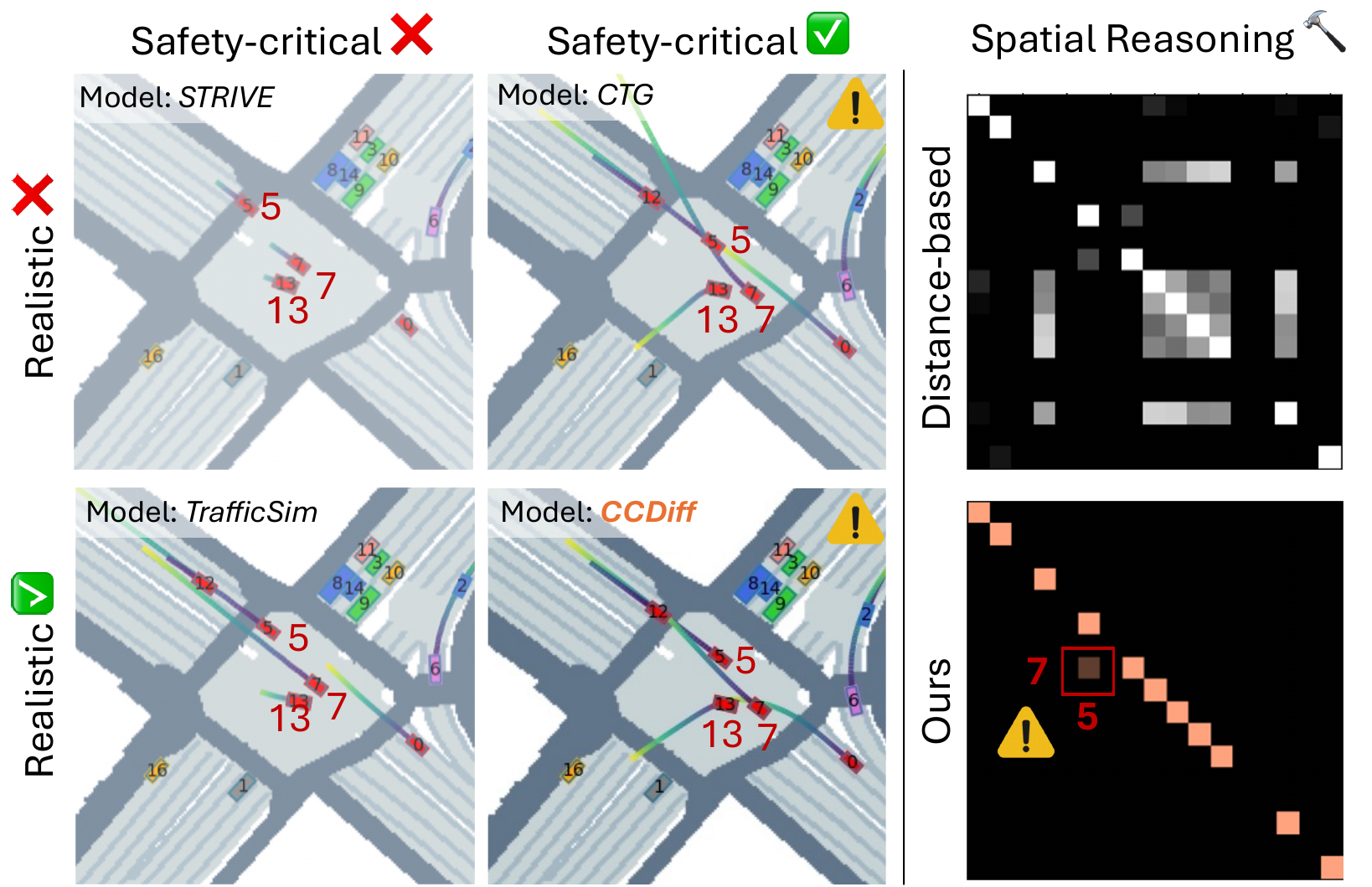}
    \caption{Comparison of safety-critical scenario generation methods, featuring \method\ alongside existing methods (STRIVE, CTG, and TrafficSim). The illustrated scenario involves Car 13 executing an unprotected left turn, prompting Car 7 to change lanes and interfere with Car 5. Unlike other methods, \method\ successfully achieves both realism and controllability in generating this safety-critical scenario. In the right column, \method's spatial reasoning method is compared to a distance-based baseline approach. \method\ accurately captures the causal relationships between key agents, identifying crucial interactions with greater precision and spatial alignment than distance-based reasoning.}
    \label{fig:teaser}
\end{figure}

Traffic agent simulation often resorts to either (i) {data-driven} approaches that generate the most probable trajectories based on scene context or (ii) {rule-based} approaches that maximize alignment with a user’s control.
However, both approaches face key limitations for effective scenario generation. 

{Data-driven scenario generation} faces two primary challenges. 
First, the rarity of collision and near-miss events in public datasets limits the ability of data-driven methods to generate safety-critical scenarios. 
As shown in prior studies~\cite{ding2023survey, hu2023simulation}, even a small domain mismatch, such as changes in road structure or the behavior of surrounding vehicles, can cause significant regressions. 
Second, closed-loop simulation requires that generated trajectories continuously interact with the simulated environment, so current predictions influence future predictions. 
This feedback loop often creates compounding errors, leading to distributional shifts that challenge the generation of both controllable and realistic behaviors over long horizons. 

On the other hand, rule-based approaches to simulation~\cite{dosovitskiy2017carla, lopez2018microscopic} offer precise user control, but often fail to capture the nuanced, adaptive behaviors of real-world driving, especially in unpredictable scenarios. 
Their rigidity can make generated behaviors feel scripted and unrealistic, particularly in closed-loop simulations where each action influences future states. 
This lack of adaptability often leads to compounding errors and a drift from realistic behavior distributions, limiting their effectiveness in complex, long-horizon interactions. 

Recent advances in deep generative models have enabled scalable traffic behavior simulation~\cite{wang2021advsim, tan2021scenegen}, facilitating realistic scenario generation from massive offline datasets. 
Notably, prior works compose explicit rules into scenario generation, such as causal graphs~(CG), signal temporal logic~(STL), or large language models~(LLM), which act as structured constraints to improve the controllability~\cite{ding2023causalaf, zhong2023guided, patrikar2024rulefuser, tan2023language, zhong2023language}. 
However, interactive driving scenarios cannot be fully encapsulated by explicit rules alone. 
Rule-based models struggle to generalize effectively in many safety-critical corner cases, where certain rules may need to be adapted.

Our key insight in driving scenarios is that interactions between agents follow an inherent causal structure: each agent’s actions depend primarily on the states of a subset of nearby agents.
Following this observation, we frame the problem as a Constrained Factored Markov Decision Process (MDP), shaped by these causal dependencies to mirror real-world interactions. 
Unlike previous approaches that manage conflicts between controllability and realism through reweighting, we directly utilize the causal structure by selectively masking agents with conflicting behaviors. 
This structure enables our model to uphold both realism and controllability constraints simultaneously, even in safety-critical situations. 
To implement this approach, we introduce the \textbf{C}ausal \textbf{C}omposition \textbf{Diff}usion model~(\textbf{\method})—a structure-enhanced diffusion model that combines structure-aware classifier-free guidance with compositional classifier-based guidance. 
By integrating these elements, our model achieves a balanced, flexible generation of realistic and controllable driving scenarios, as demonstrated in Figure~\ref{fig:teaser}.
Our contributions can be summarized as follows: 
\begin{itemize}[leftmargin=0.5cm]
    \item We formulate the learning of controllable and realistic closed-loop simulation as a constrained optimization problem, which aims to maximize the user's control preferences while satisfying realism constraints. 
    \item We propose \method, a principled algorithm to solve the constrained optimization problem by identifying the causal structure and injecting it as a structured guidance to the diffusion model. 
    \item We systematically evaluate the performance of \method\ with state-of-the-art in closed-loop scenario generation on the nuScenes dataset~\cite{caesar2019nuscenes}, showing benefits in the controllability and realism in generating safety-critical driving scenarios.
\end{itemize}


\section{Related Work}

\paragraph{Causal reasoning for behavior models}
Causal reasoning has seen extensive applications in trajectory modeling, with previous studies leveraging causal structures to enhance the robustness and generalizability of open-loop behavior prediction models. These efforts have included causal representation learning~\cite{liu2022towards}, backdoor adjustment~\cite{ge2023causal}, counterfactual analysis~\cite{chen2021human}, and realistic causal interventions~\cite{pourkeshavarz2024cadet, ding2023causalaf}. 
Despite these advances, applying a causal approach to broader trajectory prediction tasks~\cite{ettinger2021large} often demands substantial human annotation efforts~\cite{roelofs2022causalagents}. To address the challenges of automating spatiotemporal reasoning in traffic scenarios, state-of-the-art methods~\cite{ngiam2021scene, zhou2022hivt, zhou2023query, nayakanti2023wayformer} employ factorized attention mechanisms. 
However, while previous work applies causal structured reasoning primarily in open-loop settings, the efficacy of causal behavior modeling in closed-loop scenarios for autonomous driving remains under-explored.
\vspace{-2mm}
\paragraph{Generative models for traffic simulation}
Prior arts balance the trade-off between \textit{realism} and \textit{controllability} in safety-critical scenario generation by incorporating various constraints, such as inference-time sampling strategies~\cite{tan2021scenegen}, retrieval-augmented generation~\cite{ding2023realgen}, low-rank fine-tuning~\cite{dyro2024realistic}, and language-conditioned generation~\cite{tan2023language}. In closed-loop simulation methods, compositional constraints in the training loss are often integrated into the simulation pipeline. For instance, TrafficSim~\cite{suo2021trafficsim} achieves a balance between realism and common sense using a time-adaptive multi-task loss design; SimNet~\cite{bergamini2021simnet} factorizes trajectory sequences using Markov processes; STRIVE~\cite{rempe2022generating} imposes structured priors to constrain samples, avoiding unrealistic outcomes; and BITS~\cite{xu2023bits} optimizes closed-loop performance via bi-level imitation. 
Yet these prior methods struggle to resolve conflicts between \textit{controllability} and \textit{realism} objectives when these are at odds during inference.


\vspace{-2mm}
\paragraph{Diffusion model for sequential decision making}
Diffusion models~\cite{song2019generative, ho2020denoising, ho2022classifier} have shown strong controllability in density estimation and generation tasks. 
Scenario Diffusion~\cite{pronovost2023scenario} adopts latent diffusion, utilizing multi-source conditioning to generate realistic scenarios. In closed-loop traffic simulation, several prior works incorporate compositional classifier-based guidance to steer the diffusion model's sampling process~\cite{jiang2023motiondiffuser, jiangscenediffuser}, including signal temporal logic~(STL) guidance, language-based guidance~\cite{zhong2023language}, adversarial guidance~\cite{xu2023diffscene, chang2023controllable, xie2024advdiffuser}, and game-theoretic guidance~\cite{huang2024versatile}. 

Appendix Table~\ref{tab:paper_review} systematically compares the key features among the prior works and our proposed approach. 
While related works often focus on generating rule-compliant normal scenarios or enhancing safety-critical scenarios purely through classifier guidance, a fundamental challenge of achieving a balance between \textit{controllability} and \textit{realism} under safety-critical conditions remains unresolved.


\section{Problem Formulation}
\begin{figure*}
    \centering
    \includegraphics[width=1.0\linewidth]{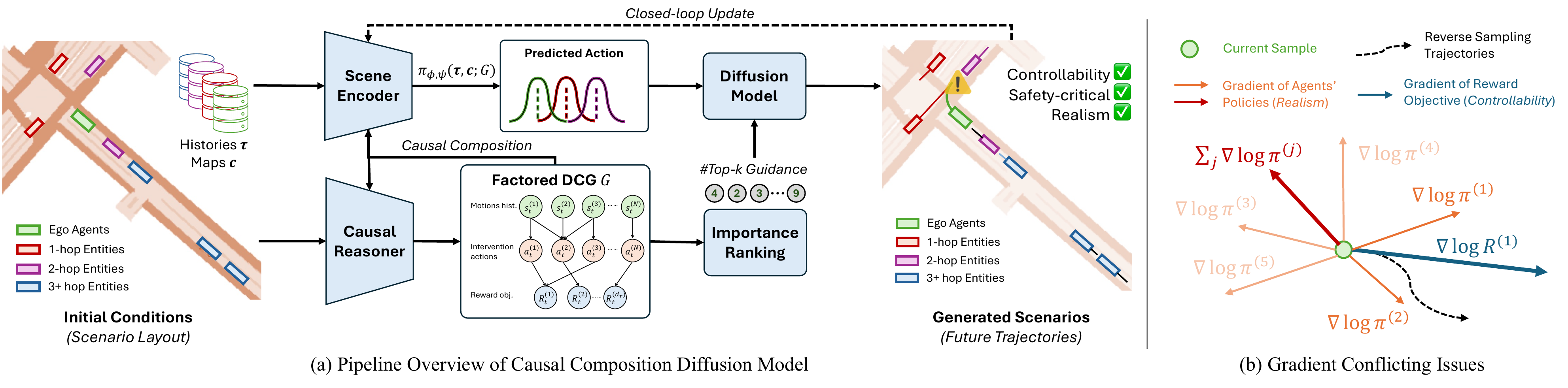}
    \caption{\textbf{(a)}: Overview of Causal Composition Diffusion Model. The scene encoder encodes the history and then uses causal reasoning for a structured scene encoding and causal ranking. Finally, we exert guidance only to the top-K agents and eliminate the non-causal agents that would not contribute to the guidance objective to maintain better realism. \textbf{(b)}: Summing up the score functions over all the agents achieves sub-optimal performance due to the conflict between the gradients of realism and controllability objectives.}
    \label{fig:ccdiff-diagram}
    \vspace{-5mm}
\end{figure*}


\subsection{Constrained Factored Markov Decision Process}
\label{section: problem formulation}

We formulate the closed-loop traffic simulation as an MDP problem, then utilize diffusion model for sequential modeling to learn a controllable simulation policy $\pi$.
Since we would like to exploit the causal structure between the state, action, and reward space, we define the Constrained Factored MDP as follows: 
\begin{definition}[Constrained Factored MDP]
\label{def:factored-mdp}
A \textit{Constrained Factored Markov Decision Process} (CFMDP) is a Markov Decision Process where the state space $\mathbfcal{S}$ and reward function $R$ are factorized to exploit the structure of the problem. 
A CFMDP is defined by the tuple: $\mathcal{M}_F = \big( \mathbfcal{S}, \mathbfcal{A}, P, R, C, \bm{s}_0 \big)$. 
\\
The factored state space, denoted as $\mathbfcal{S} = \mathcal{S}^{(1)} \times \mathcal{S}^{(2)} \times \dots \times \mathcal{S}^{(N)}$, represents the motion trajectory space at the current step for each agent $i$. The factored action space, $\mathbfcal{A} = \mathcal{A}^{(1)} \times \mathcal{A}^{(2)} \times \dots \times \mathcal{A}^{(N)}$, consists of interventions on the subsequent driving behaviors for each agent in the scenario. 
The joint transition dynamics $P(\bm{s}_{t} | \bm{s}_{t-1}, \bm{a}_{t-1}) = \prod_{i=1}^{N} p_i({s}_{t}^{(i)} | \bm{s}_{t-1}, {a}_{t-1}^{(i)})$, are defined over the state $s\in \mathbfcal{S}$ and action $a\in \mathbfcal{A}$ pairs. In our case, $P$ is the \textit{deterministic} vehicle dynamics for each agent in our setting. 
The reward objective $R(\bm{s}, \bm{a}) = \sum_{j=1}^{d_r} R^{(j)}(s^{(I_j)}, \bm{a})$ include collision, off-road events, over-speed, or other objectives, where each subset $I_j$ specifies the state factors impacting the $j$-th reward. For a learned policy $\pi$, the constraint function $C(\bm{s}, \bm{a}) = \mathbb{D}\big(\pi_\beta(\cdot|\bm{s}_t) \| {\pi}(\cdot | \bm{s}_t) \big)$ indicates the realism level of generated trajectories with respect to the dataset policies $\pi_\beta$, where a lower constraint value implies greater realism. The initial state $\bm{s}_0$ lies in the factored state space $\mathbfcal{S}$. 
\end{definition}

We then formulate the closed-loop scenario generation problem as a constrained optimization problem that aims to find an intervention policy ${\pi}$ that maximizes the controllability $R(\bm{\tau})$ while maintaining an acceptable deviation in the realism $C(\bm{\tau})$: 
\begin{equation*}
\begin{aligned}
 \max_{\pi} \mathbb{E}_{\bm{\tau}\sim (P, \pi)} \big[R(\bm{\tau}) \big],\quad  s.t. \ \mathbb{E}_{\bm{\tau}\sim (P, \pi)} \left[C(\bm{\tau}) \right] \leq \kappa,
   \label{eq:moo}
\end{aligned}
\end{equation*}
where the cumulative reward, $R(\bm{\tau})$, represents the total reward accumulated from individual reward factors along the trajectory $\bm{\tau}$:
$R(\bm{\tau}) = \sum_{j=1}^{d_r} R^{(j)}(\bm{\tau}^{(I_j)})=\mathbb{E} \left[\sum_{t=1}^{T} \sum_{j=1}^{d_r} R^{(j)}(s^{(I_j)}_t, a_t^{(I_j)}) \right]$. 
The cumulative cost $C(\bm{\tau})$ quantifies the realism constraints, measuring how closely the generated trajectory $\bm{\tau}$ resembles the ground-truth trajectory $\bm{\tau}^*$. Following the approach in~\cite{zhong2023guided, xu2023bits}, we use the Total Variation (TV) distance between the estimated intervention policy ${\pi}(\bm{a}_t | \bm{s}_{t})$ and the dataset policy $\pi_\beta(\bm{a}_t | \bm{s}_{t})$: 
$
C(\bm{\tau}) \triangleq\sum_{t=1}^T \mathbb{D}\big(\pi_\beta(\bm{a}_{t}| \bm{s}_t) \parallel {\pi}(\bm{a}_{t}| \bm{s}_t) \big).
$

 To further incorporate the structure in this multi-agent decision-making problem~\cite{grimbly2021causal}, we define the Decision Causal Graph~(DCG) below. 
\begin{definition}[Decision Causal Graph]
    For every timestep $t$, we define a causal graph $G\in \mathbb{R}^{N\times N}$, where $G_{ij}=0$ if and only if the future action of agent $j$ is conditionally independent with the  $i$-th agent's history: $a_t^{(j)}\indep s_{t}^{(i)} | s_{t}^{(-i)}$. And $G_{ij}=1$ means there exists a causal edge $s_{t}^{(i)}\to a_t^{(j)}$. 
\end{definition}
 Following the definition above, we can also define a set of policy $\bm{\pi}(a_t^{(1)}, \cdots, a_t^{(N)} | \bm{s}_t) = \prod_{i=1}^N \pi^{(i)} (a_t^{(i)} | \textbf{PA}_t^G(i))$, where $\textbf{PA}_t^G(i)\in \{s^{(1)}, s^{(2)}, \cdots, s^{(N)}\}$ is the causal parents to the $i$-th agents in graph $G$ when making decisions. 
A diagram of Factored DCG is illustrated in Figure~\ref{fig:ccdiff-diagram}(a).

\subsection{Diffusion Model for Sequence Modeling}
We then solve the traffic simulation inspired by the recent advancement in diffusion-guided sequential data generation~\cite{janner2022planning, ajay2022conditional, zhong2023guided}.
Denote $\bm{\tau}(k) \triangleq \{(\bm{s}_t(k), \bm{a}_t(k))\}_{t=1}^T$ represent the joint state-action trajectory at the $k$-th diffusion step, $k\in \{0,1,\cdots, K\}$, where $\bm{\tau}(0)$ denotes the original clean trajectory. The forward diffusion process, acting on $\bm{\tau}(0)$, gradually corrupts it with Gaussian noise:
{\small
\begin{equation*}
\setlength{\abovedisplayskip}{1pt}
\setlength{\belowdisplayskip}{1pt}
\begin{aligned}
\label{eq:diffusion_q}
& q(\bm{\tau}(1:K)|\bm{\tau}(0)) \triangleq \prod_{k=1}^{K} q(\bm{\tau}(k)|\bm{\tau}(k-1)), \\
& q(\bm{\tau}(k)|\bm{\tau}(k-1)) \triangleq  \mathcal{N}\left(\bm{\tau}(k); \sqrt{1-\beta_k} \, \bm{\tau}(k-1), \beta_k \bm{I}\right),
\end{aligned}
\end{equation*}
}
where $\beta_1, \dots, \beta_K$ are pre-defined variance schedules at each diffusion step. 
Over the forward process, the trajectory is transformed into a standard Gaussian distribution:
$q(\bm{\tau}(K)) \approx \mathcal{N}(\bm{0}, \bm{I})$. For scenario generation, the reverse diffusion process iteratively denoises from noise to recover the original trajectories. 
Given a context $\bm{c}$ (e.g., map features), the reverse process is:
{
\small
\begin{equation*}
\begin{aligned}
\label{eq:denoise_p}
& p_{\phi, \psi}(\bm{\tau}({0:K}) | \bm{c}) = p(\bm{\tau}(K)) \prod_{k=1}^{K} p_{\theta}(\bm{\tau}({k-1}) | \bm{\tau}(k), \bm{c}), \\
& p_{\phi, \psi}(\bm{\tau}(k-1) | \bm{\tau}(k), \bm{c}) = \mathcal{N}\left(\bm{\tau}(k-1); \bm{\pi}_{\phi, \psi}(\bm{\tau}(k), k, \bm{c}), \sigma_k^2 \bm{I}\right),
\end{aligned}
\end{equation*}
}
where \(p(\bm{\tau}(K)) = \mathcal{N}(\bm{0}, \bm{I})\) is the Gaussian prior, and $\bm{\pi}_{\phi, \psi}$ is the scene encoder parameterized by $\phi, \psi$, which will be covered in later sections. 


\section{Methodology}



\subsection{Realism Constrained Score Matching}

We denote a factored optimality of time-step $t$, a set of binary random variables as $\mathbfcal{O}_t = \{\mathcal{O}_t^{(j)}\}_{j=1}^{d_r}$~\cite{levine2018reinforcement, janner2022planning}. The joint optimality in all reward objectives can be written as $p(\mathcal{O}_t^{(j)}=1|\bm{\tau}_{t})\propto \exp\big( R^{(j)}(\bm{\tau}_t^{(j)}) \big)$. We slightly exploit the notations $\bm{\tau} \triangleq \{(\bm{s}_t, \bm{a}_t)\}_{t=1}^T $ as the trajectories of state action pairs, where $\bm{s}_t\in \mathbb{R}^{N\times d_s}$ is the state trajectories for all $N$ agents. Given the CFMDP in Definition~\ref{def:factored-mdp}, with known transition (vehicle) dynamics, we can factorize the objective of the optimal closed-loop scenario generation as follows: 

\begin{equation}
\small
\begin{aligned}
    \max & P(\mathbfcal{O}_t=1,  \bm{\tau}_t |   \bm{\tau}_{t-1}) \Leftrightarrow  \max  P(\mathbfcal{O}_t=1 | \bm{\tau}_t) P(  \bm{\tau}_t |\bm{\tau}_{t-1}) \Leftrightarrow \\
   \max_{\pi} & P(\mathbfcal{O}_t=1 | \bm{s}_{t}, \bm{a}_t)  \pi(\bm{a}_t | \bm{s}_t) P(\bm{s}_t | \bm{s}_{t-1},\bm{a}_{t-1}) \Leftrightarrow \\
    \max_{\pi} &  \underbrace{\prod_{j=1}^{d_r} \exp\big(R^{(j)} (s_t^{(I_j)}, \pi(\bm{s}_t)) \big)}_{\text{Controllability}} \underbrace{\prod_{i=1}^{N} \pi^{(i)}\big(a_t^{(i)} | \bm{s}_t \big)}_{\text{Realism}},  \\
\end{aligned}
\label{eq:mle}
\end{equation}
where the first term corresponds to \textit{controllability}, i.e. the likelihood of optimality specified by some user-specified reward objective, and the second term corresponds to the \textit{realism}, the likelihood of generated behaviors. 
We denote $\nabla \log P\triangleq \nabla \log  P(\mathbfcal{O}_t=1,  \bm{\tau}_t |   \bm{\tau}_{t-1})$ The score function of the maximum likelihood objective in~\eqref{eq:mle} can be written as~\cite{song2019generative, janner2022planning}: 
{\small
\begin{equation}
\begin{aligned}
    \nabla \log P =  \sum_{j=1}^{d_r} \nabla_{\bm{\tau}} R^{(j)}({s}_t^{(I_j)}, \pi(\bm{s}_t)) + \sum_{i=1}^N \nabla_{\bm{\tau}} \log \pi^{(i)}  \big(a_t^{(i)} | \bm{s}_t \big) \\
\end{aligned}
\end{equation}
}
Unlike the normal scenarios where optimizing the imitation basically adheres with the rule compliance reward, safety-critical guidance $R^{(j)}$ can suffer from gradient conflict~\cite{yu2020gradient, dinh2023pixelasparam, yao2024gradient, ma2024decouple}.  Namely, for some $i\in [1, N], j\in [1, d_r]$, if
\begin{equation*}
\big\langle \nabla_{\bm{\tau}} \log \pi^{(i)}  \big(a_t^{(i)} | \bm{s}_t \big), \nabla_{\bm{\tau}} R^{(j)}(\bm{\tau}^{(I_j)}) \big\rangle < 0, 
\end{equation*}
using a weighted sum of all the objectives as classifier-based guidance would achieve sub-optimal performance, as illustrated in Figure~\ref{fig:ccdiff-diagram}(b). 
In order to resolve this gradient conflicting issues, we need to prioritize to control the agents index $i\in [N]$ that could maximize the reward while maintaining a high likelihood of the learned policies, i.e., a lower realism gap between the learned policies and behavior policies $\pi_\beta$: $\mathbb{D}(\pi_\beta \| \widehat{\pi})$. 
We denote the flag of controllable agents as ${\bm{\rho}}\in [0, 1]^{N}$, the target simulation policies: 
{\small{
\begin{equation*}
    \pi^{(i)}(a_t^{(i)} | \bm{s}_t) = \begin{cases}
        \pi_{\rho, \psi}^{(i)}(a_t^{(i)} | \bm{s}_t), & {\bm{\rho}_i}=1 \\
        {\pi}_\beta^{(i)}(a_t^{(i)} | \bm{s}_t), & {\bm{\rho}_i}=0 \\
    \end{cases}
\end{equation*}
}
}
We can use the Lagrangian multiplier~\cite{achiam2017constrained} and structured projected gradient descent~\cite{bahmani2016learning} to solve the constrained optimization with the following maximum likelihood estimation problems: 
{\small
\begin{equation}
\begin{aligned}
    \max_{\substack{\pi \in {\Pi},{\bm{\rho}} \in \{0, 1\}^{N}\\ {G}\in \{0,1\}^{N\times N}} } & 
   \prod_{j=1}^{d_r} \exp\big(R^{(j)} (\bm{\tau}_t^{I_j}; {\bm{\rho}}) \big)
 \prod_{i\in [N], {\bm{\rho}_i}=1}  \pi^{(i)}\big(a_t^{(i)} | \textbf{PA}_t^{{G}}(i) \big) \\
    s.t. \quad  & |{G}| \leq C_{\text{sparsity}}, \quad \sum_i {\bm{\rho}_i} \leq N_c. 
\end{aligned}
\label{eq:new_moo}
\end{equation}
}
 We can then control the realism level by changing the constraint level of $N_c, C_{\text{sparsity}} \in \mathbb{Z}^+$.



\subsection{Proposed Method: \method}
We hereby introduce \method\ to optimize the simulation policy $\pi$ of~\eqref{eq:new_moo} in a scalable and efficient way by decomposing the constrained optimization problem into several small components. 
We illustrate the pipeline in Figure~\ref{fig:ccdiff-diagram}(a). 
To promote \textit{realism}, \method\ first encodes the motion histories of different agents based on the {spatial attention}, then discovers the {decision causal graph} ${G}$ based on the factorized attention masks and kinematic factors. \method\ then utilizes causal interactive patterns in ${G}$ to extract the importance rank ${\bm{\rho}}$. Finally, \method\ optimizes its \textit{controllability} by masking out those unimportant agents based on ${\bm{\rho}}$ to guide the diffusion reverse sampling process in a structured way. We zoom into the details below. 

\begin{figure}
    \centering
    \includegraphics[width=1.0\linewidth]{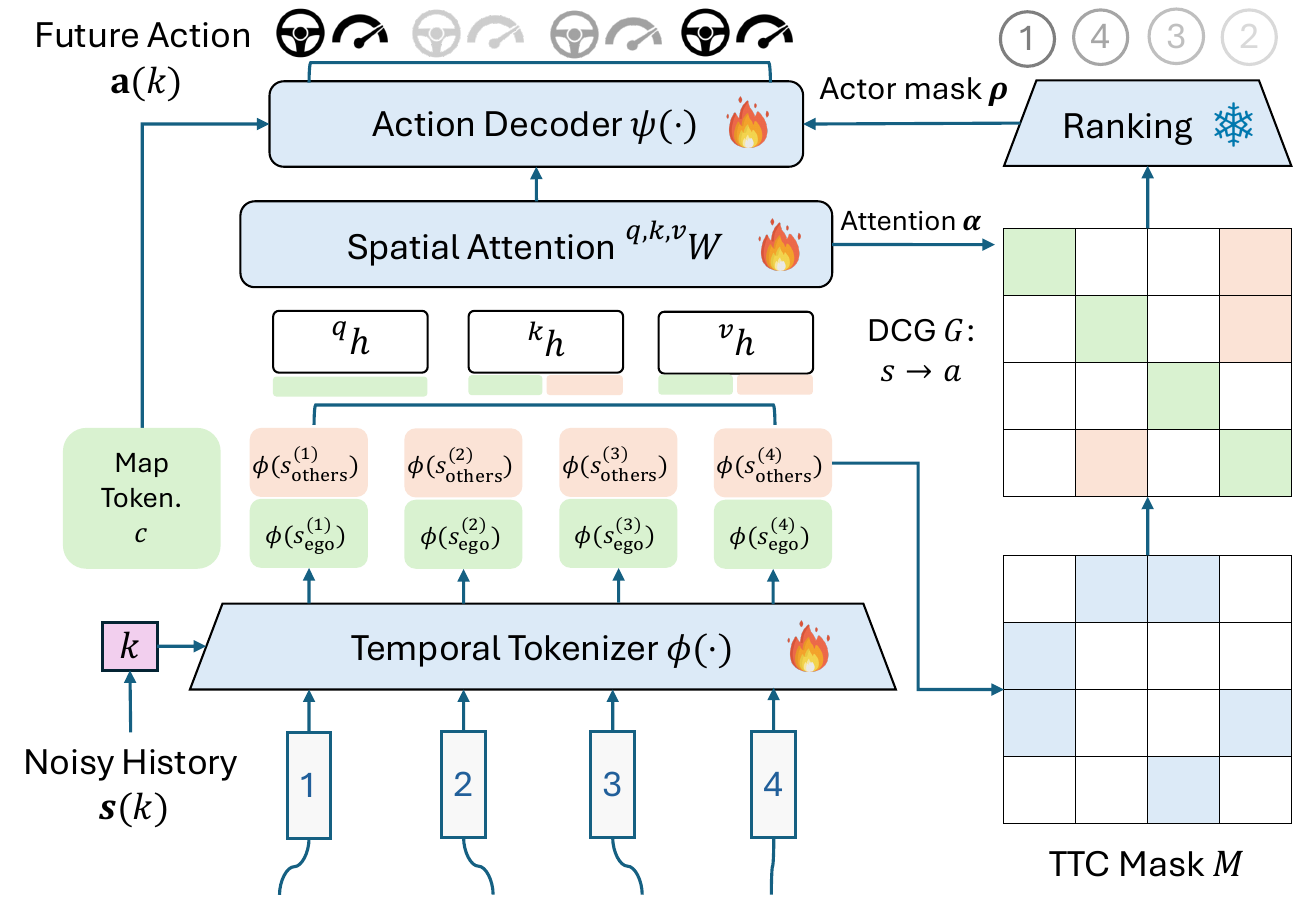}
    \caption{Detailed model structure of~\method, which incorporates temporal tokenizer, spatial attention, and action decoding. The decision causal graph helps to extract the spatial patterns to identify the most relevant actions, then use the ranking outputs to mask the output of the action. \textcolor{red}{\faFire} means trainable modules, and \textcolor{blue}{\Snowflake} means non-trainable parts during training. }
    \label{fig:model_structure}
    \vspace{-5mm}
\end{figure}

\paragraph{Causal Composition Scene Encoder}
Inspired by~\cite{brouillard2020differentiable, scherrer2021learning}, the goal of causal composition scene encoder is to generate the most likely action under a parsimonious decision causal graph $G$ with Lagrangian multiplier $\lambda_{\text{sparsity}}$: 
\begin{equation*}
\begin{aligned}
    \max_{{{G}}\in [0, 1]^{N\times N}} \quad  & \prod_{i\in [N]} \pi^{(i)} (a_t^{(i)} | \textbf{PA}_t^{{G}} (i)) + \lambda_{\text{sparsity}}\cdot |G|
\end{aligned}
\end{equation*}
We parameterize our model $\pi_{\phi,\psi}(\bm{a}_t | \bm{s}_t, \bm{c}, k; {G})$ for scenario generation with a transformer-based structures for temporal attention $\phi$, spatial attention modules $\psi$, as well as some decision causal graph ${G}$. The model output $\bm{a}_t$ is conditioned on the agents' history $\bm{s}_t$, map context $\bm{c}$, and diffusion sample step $k$. 
Similar to the scene transformer structure in~\cite{zhou2022hivt, zhong2023language}, we first embed the history of ego and surrounding agents with temporal attention layer: $\phi_{\text{ego}}({s}_{t}^{(i)}), \phi_{\text{others}}( \bm{s}_{t}^{(-i)\to (i)})]$, here ${s}_{t}^{(i)}$ is the history of the $i$-th agents, and $\bm{s}_{t}^{(-i)\to (i)}$ are the relative history of all the other agents than $i$. 
To facilitate the relational reasoning, we incorporate both the absolute and relative features in $\phi_{\text{others}}(\cdot)$, including the position, velocity, distance, and time-to-collision~(TTC). 
Then we can aggregate all the temporal information into the following spatial cross-attention layer: 
\begin{equation*}
\begin{aligned}
    & {}^q h_t^{(i)} = \phi_{\text{ego}}({s}_{t}^{(i)}),  \\
    & {}^k h_t^{(i j )} ={}^v h_t^{(ij)} = [\phi_{\text{ego}}({s}_{t}^{(j)}), \phi_{\text{others}}( \bm{s}_{t}^{(i)\to (j))})]]. 
\end{aligned}
\end{equation*}
\\
In order to further discover useful spatial parent-to-child relationships, we design a two-step causal reasoning to identify the DCG in the spatial-temporal interaction of the traffic agents. First, we set a hard constraint over the neighborhood perception field by trimming down the unnecessary causal connection between agents' states and corresponding actions at time-step $t$.  
Second, we apply the first tunable hard constraint as a memory mask to the attention weights: 

\begin{equation}
    {G_{ij}}(\bm{\tau}_t) = M_{ij}(\bm{\tau}_t)\cdot \text{softmax}\big(\frac{({}^q W {}^q h_t^{(i)})^T ({}^k W {}^k h_t^{(ij)})}{\sqrt{d_k}}\big), 
\end{equation}
where the memory mask $M$ is extracted with relative TTC features $f_{\text{TTC}}(\cdot)$ with the surrounding agents given the threshold $C_{\text{ttc}}$ of causal graph $G$: 
\begin{equation*}
    M_{ij}(\bm{\tau}_t) = \begin{cases}
    1, & f_{\text{TTC}}(\phi_{\text{others}}(\bm{s}_t^{(j)\to (i)})) \leq C_{\text{ttc}} \\
    0, & \text{otherwise}
    \end{cases}
\end{equation*}
In practice, we can tune the threshold of $C_{\text{ttc}}$ here to control the sparsity of the final causal graph. We then aggregate the map information $\bm{c}$ into the decoder. 
The output layer aggregate the state of causal parental agents $\textbf{PA}_t^G(i)$ to get the action: ${a}_{t}^{(i)}(k)= \psi\big(\phi(\textbf{PA}_t^{{G}}(i)), \bm{c}, k \big)$.
\vspace{-4mm}
\paragraph{Causal Ranking}
We then use the identified DCG to rank the agents' importance to the safety-critical objectives $R^{(j)}(\bm{\tau})$: 
\begin{equation*}
    \bm{\rho}_i = \argmax_{i\in [N]} \big\langle \nabla_{\bm{\tau}} \log \pi^{(i)}  \big(a_t^{(i)} | \bm{s}_t \big), \nabla_{\bm{\tau}} R^{(j)}(\bm{\tau}^{(I_j)}) \big\rangle
\end{equation*}
\vspace{-5mm}
\\
To automate the ranking process, we resort to the estimated causal graph ${G}$ above. The causal composition scene encoder gives us ${G}$ and a policy network $\pi_{\phi,\psi}(\bm{a}_t | \bm{s}_t; {G})$. 
Then we design a graph-based community detector on the DCG ${G}$, then sort the time of occurrences in any cliques for all the nodes from $1$ to $N$~\cite{chen2021human, liu2022towards, pourkeshavarz2024cadet}.
After sorting, we have the ranked id sequence $\{\rho_i(\tau)\}_{i=1}^N$, then we can pick the top $N_c$ \textit{key} agents $\{\rho_i(\tau)\}_{i=1}^{N_c}$ at the scene, which represent the most densely interactive with the other agents. 
This ranking process empirically helps identify the most interactive and influential agents for the safety-critical objective.
We further discuss in the appendix with more details about the specific design of relational features $\phi$ and the community detection algorithms we used.
\\

\vspace{-8mm}
\paragraph{Causal Composition Guidance}
For the diffusion guidance process, similar to the inpainting technique~\cite{janner2022planning}, we aim to trim down the controllable space by reducing the number of controllable agents with cause-and-effect ranking. 
With the causal reasoner and importance ranker modules, we sorted out the key agents $\{\rho_k(\tau)\}_{k=1}^K$. We then apply both classifier-based and classifier guidance. 

In \method, we derive a special form of classifier-free guidance~\cite{ho2022classifier} as a combination of unconditional scene encoding and causal interventional encoding. At timestep $t$, for the top-$N_c$ controllable agents $i\in \{{\bm{\rho}_{N_c}(\tau_t)}\}_{i=1}^{N_c}$, the classifier-free guidance is:
\begin{equation*}    
(1 - w) \nabla_{\bm{a}} \log \pi_{\phi,\psi}i(a_t^{(i)}| s_t^{(i)}) + w \nabla_{\bm{a}} \log \pi_{\phi,\psi}(a_t^{(i)}| \textbf{PA}_t^G(i)),
\end{equation*}
where \( w \) is the guidance scale. For agent $i$, $\pi_{\phi,\psi}(a_t^{(i)}| s_t^{(i)})$ is the unconditional distribution that only considered the ego histories $\phi(s_{\text{ego}})$, and $\pi_{\phi,\psi}(a_t^{(i)}| \textbf{PA}_t^G(i))$ is the intervened encoded results given some parental agents in causal graph $G$. This formulation implies that the guided distribution corresponds to a geometric mixture:
\begin{equation*}
{\pi}(a_t^{(i)}) \propto \underbrace{\pi(a_t^{(i)}| \textbf{PA}_t^{{I}_N}(i))^{1 - w}}_{\text{Original}} \cdot \underbrace{\pi(a_t^{(i)}| \textbf{PA}_t^{{G}}(i))^{w}}_{\text{Intervened}}.
\end{equation*}
Thus, classifier-free guidance in diffusion models can be viewed as a do-intervention~\cite{pearl2009causality} by specifying the causal parents of $i$-th agents in DCG during the generative process. The guidance scale $w\in [1,2)$ acts analogously to the strength of intervention, extrapolating the original and intervened distributions. 
With the causal ranking, we mask out the agents with conflicted gradients as a reweighted classifier-based guidance: 
\begin{equation}
    \sum_{j=1}^{d_r} \nabla_{\bm{a}^{(I_j)}} R^{(j)}(\tau) \approx \sum_{j=1}^{d_r} {\bm{\rho}_j(\tau)} \odot  \nabla_{\bm{a}} [R^{(j)}(\tau)]
\end{equation}
\vspace{-4mm}
\\
In practice, we use the distance-based guidance objective over the trajectories, including the map collision guidance and the agent collision guidance~\cite{zhong2023guided}. We also use the same classifier function for all the baseline methods, see detailed description in the appendix~\ref{app:baseline}. 


\vspace{-4mm}
\paragraph{Training and Inference} 
We train the model with using the classical DDPM~\cite{ho2020denoising} diffusion with classifier-free guidance~\cite{ho2022classifier}. Specifically, the loss we solve is $\min_{\phi,\psi} \mathbb{{G}} \left\|\pi_{\phi, \psi}(\bm{\tau}(k), \bm{c}, k; {G}(\tau)) - \bm{a}(0) \right\|^2$. We introduce counterfactual conditions by randomly dropping the DCG $G$ of the scene transformer by replacing the decision causal graph $G$ as a diagonal matrix, so all agents' actions are only conditioned on the ego history. At inference time, we combine both classifier-based and classifier-free guidance to facilitate better controllability, see algorithm~\ref{alg:inference}. 

\begin{algorithm}
\caption{\method\ for Scenario Generation}\label{alg:inference}
\begin{algorithmic}
\Require Dropout $p_{\text{uncond}}$, threshold $C_\rho, C_{\text{ttc}}$
\Require Guidance loss $\{\mathcal{J}_i\}_{i=1}^N$, trajectories $\bm{\tau}$, map $\bm{c}$. 

\While{$k = K, \dots, 1$}  \Comment{\textit{Inference Sampling}}
    \State $\pi_{\text{uncond}}, \bm{\alpha}_\text{attn}\gets \pi_{\phi,\psi}(\bm{\tau}(k), \bm{c}, k; \varnothing) $
    \State ${G}(\bm{\tau}) \gets M(\bm{\tau})\cdot \bm{\alpha}_\text{attn}$  
    \Comment{Causal masking}
    \State ${\bm{\rho}(\bm{\tau})} \gets \text{ranking}({G})$  
    \Comment{Importance ranking}
    \State {$ \widehat{\pi} \gets(1 - \omega) \pi_{\phi,\psi}(\bm{\tau}(k), \bm{c}, k; {G}) + \omega \pi_{\text{uncond}} $}
    \State $\bm{a}{(0)} \sim \widehat{\pi} (\cdot | \bm{\tau}(k), \bm{c}; {G})$
    \State $\bm{a}(k-1) \gets \bm{a}(k-1) + \sum_{i=1}^{d_r} {\bm{\rho}_i(\tau)} \cdot \nabla R(\bm{s}, \bm{a}{(k-1)})$
    \State $\bm{\tau}({k-1}) \gets f_{\text{dyn}}\big(\bm{s}, \bm{a}(k-1)\big)$ \Comment{Vehicle dynamics}
\EndWhile
\\
\Return Generated trajectory $\bm{\tau}(0)$. 
\end{algorithmic}
\end{algorithm}
\vspace{-5mm}

\section{Experiment}

\begin{figure*}[htbp]
    \centering
    \includegraphics[width=0.95\linewidth]{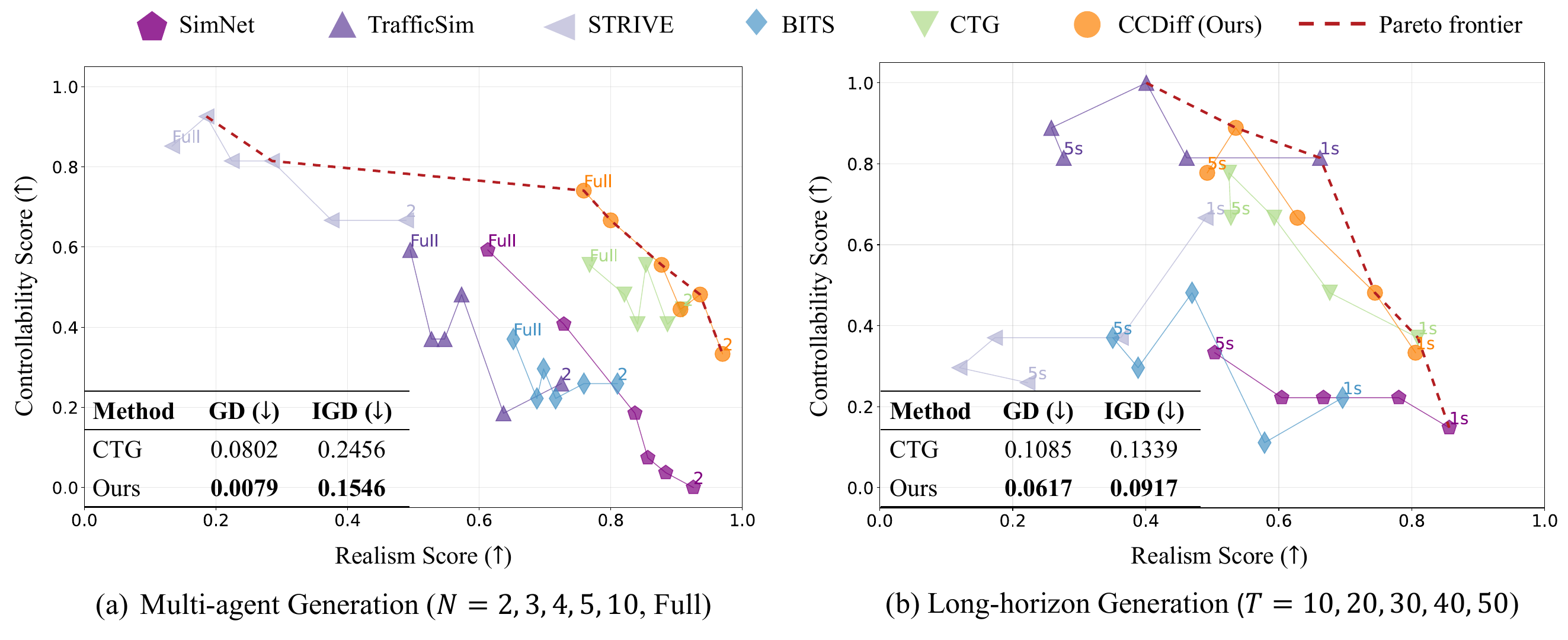}
    \caption{Plot of the controllability v.s. realism in the multi-agent and long-horizon generation settings. \method\ outperforms baselines in both the Generational Distance~(GD) and Inverted Generational Distance~(IGD), with better proximity to the Pareto frontier, and better coverage of the optimal solution along the frontier in this multi-objective optimization. Our method is more realistic and controllable compared to other approaches consistently in both multi-agent scenario generation and long-horizon scenario generation. }
    \label{fig:cr_plot}
    \vspace{-5mm}
\end{figure*}


In the following parts of the experiments, we aim to answer the following three research questions~(\textbf{RQ}s): \textbf{RQ1}: Under different sizes of total controllable agents, how are the realism and controllability of the safety-critical scenarios generated by \method\ compared to the baselines? \textbf{RQ2}: With a longer generation horizon and lower frequency, how are the realism and controllability of the safety-critical scenarios generated by \method\ compared to the baselines? \textbf{RQ3}: How much does the causal reasoning module in \method\ contribute to the overall performance? 

The remaining parts of the experiment section first introduce our experiment settings, then compare our methods with baselines in the \textit{controllability} and \textit{realism} to answer the research questions. Finally, we conduct ablation studies to show the effect of individual modules in~\method. 

\subsection{Experiment Settings}
\paragraph{Datasets}
We use the nuScenes dataset~\cite{caesar2019nuscenes} and traffic behavior simulation~(tbsim)~\cite{xu2023bits} for model training and evaluation. 
We train all models on scenes from the train split and evaluate on 100 scenes randomly sampled from the validation split. 
During evaluation phase, we initialize all the models with the same set of initial layouts and initial history trajectories of 3 seconds, the model is responsible of generating the future 10 seconds of trajectories for the driving agents in a closed-loop manner. 
\vspace{-5mm}

\paragraph{Baselines}
We implement the following baselines in the above platform settings. To systematically illustrate the effectiveness of \method, we include the following SOTAs for comparison: \textbf{SimNet}~\cite{bergamini2021simnet}, \textbf{TrafficSim}~\cite{suo2021trafficsim}, \textbf{BITS}~\cite{xu2023bits}, \textbf{Strive}~\cite{rempe2022generating}, and \textbf{CTG}~\cite{zhong2023guided}. 
We compare all the baselines with \method\ in the publicly available nuScenes dataset~\cite{caesar2019nuscenes} and baseline implementations\footnote{\href{https://github.com/NVlabs/CTG}{{https://github.com/NVlabs/CTG}}}. 
For a fair comparison with all the baselines, we use the rasterized map used in the previous works and encode them with ResNet-18 for the map conditioning $\bm{c}$ for all the methods. 

\vspace{-3mm}

\paragraph{Metrics} We compare the performance of \method\ and all the baselines with the following categories of metrics: 
\begin{itemize}
    \item \textbf{Controllability Score~(CS)}: we use the scenario-wise collision rate~(SCR) used in~\cite{suo2021trafficsim, tan2023language} as the controllability metrics. Among all the testing scenarios, we calculate the proportion of the scenarios where at least one collision event occurred between different agents. We then standardize SCR among all the methods to get the \textbf{CS}, a higher-the-better score between 0 and 1. 
    \item \textbf{Realism Score~(RS)}: How to quantify realism is an open problem in evaluating traffic scenarios. In order to get a more interpretable and direct way to quantify realism, we adopt three widely-used quantitative metrics to evaluate the realism of the scenarios: (i) scenario off-road rate~(ORR) used in~\cite{tan2023language, ding2023realgen}, (ii) final displacement error~(FDE, $\mathrm{m}$) and (iii) comfort distance~(CFD), which is used in in~\cite{xu2023bits,zhong2023guided} to quantify the realism of the similarity in the smoothness of agents' trajectories in the generated scenarios. We standardize all the metrics among all the methods respectively and average them to get the \textbf{RS}, a higher-the-better score between 0 and 1. 
    \item \textbf{Multi-objective optimization metrics}: with the \textbf{RS} and \textbf{CS},  we further quantify the optimality of the solution based on generational distance~(\textbf{GD}) and inverted generational distance~(\textbf{IGD}), the average minimum distance between the methods and Pareto frontier~\cite{coello2007evolutionary, yu2020gradient}.
\end{itemize}

\subsection{Multi-agent Scenario Generation~(RQ1)}
To address~\textbf{RQ1}, we train our model on the training split of the nuScenes dataset and vary the number of controllable agents from 2 agents to the full sets of agents for all baselines at inference time by running a closed-loop generation at 2$\mathrm{Hz}$~(0.5$\mathrm{s}$). 
We then evaluate the CS and RS of generated scenarios under different numbers of controllable agents and report the comparison between \method\ with SOTAs in Table~\ref{tab:spatial_cr_value} and Figure~\ref{fig:cr_plot}(a). 

From the table, we can see that \method\ outperforms SimNet, TrafficSim, and BITS in both controllability and realism for almost all the cases, whereas TrafficSim only outperforms \method\ in one controllability score. 
STRIVE shows some good controllability in generating safety-critical scenarios, yet its realism score is the poorest among all. 
The closest competitor, CTG, shows comparable performance in realism, yet \method\ outperforms CTG in the controllability metrics, especially when the size of controllable agents goes larger as we gradually scale up the number of controllable agents from 2 to 5 and eventually to the full size of agents at the scene.
From Figure~\ref{fig:cr_plot}(a), we can also see that \method\ enjoys significantly better realism and controllability score on the most upper right side,  Pareto front 
More detailed results for the SCR, ORR, FDE, and CFD are illustrated in the appendix Table~\ref{tab:combined_evaluation}. 
The consistent benefits of~\method\ in RS and CS compared to other baselines confirm that composing causal structure in the diffusion model facilitates the algorithm to generate reasonable safety-critical scenarios. 
We also show qualitative studies for long-horizon generation in the Appendix Figure~\ref{fig:qualitative}.

\begin{table}[h!]
\centering
\caption{Comparison in Controllability~(\textbf{CS}) and Realism~(\textbf{RS}) among all the baselines in $K$-agent scenario generation. \colorbox{red!15}{Red} means \method\ outperforms the baseline in the corresponding metrics, \colorbox{green!15}{green} means the baseline is better. \method\ has \textbf{best} or \textcolor{blue}{\textbf{second best}} performance in 10 out of 12 metrics. }
\resizebox{1.\linewidth}{!}{
\begin{tabular}{l l c c c c c c}
\toprule
\textbf{Method} & \textbf{Metric} & $K$=2 & 3 & 4 & 5 & 10 & Full \\
\midrule
SimNet     & CS~($\uparrow$) & \cellcolor{red!15}0.00 & \cellcolor{red!15}0.04 & \cellcolor{red!15}0.07 & \cellcolor{red!15}0.19 & \cellcolor{red!15}0.41 & \cellcolor{red!15}0.59 \\
           & RS~($\uparrow$) & \cellcolor{red!15}\textbf{\textcolor{blue}{0.93}} & \cellcolor{red!15}0.88 & \cellcolor{red!15}0.86 & \cellcolor{red!15}0.84 & \cellcolor{red!15}0.73 & \cellcolor{red!15}0.61 \\
\midrule
TrafficSim & CS~($\uparrow$) & \cellcolor{red!15}0.26 & \cellcolor{red!15}0.19 & \cellcolor{green!15}0.48 & \cellcolor{red!15}0.37 & \cellcolor{red!15}0.37 & \cellcolor{red!15}0.59 \\
           & RS~($\uparrow$) & \cellcolor{red!15}0.72 & \cellcolor{red!15}0.64 & \cellcolor{red!15}0.57 & \cellcolor{red!15}0.55 & \cellcolor{red!15}0.53 & \cellcolor{red!15}0.50 \\
\midrule
STRIVE     & CS~($\uparrow$) & \cellcolor{green!15}\textbf{0.67} & \cellcolor{green!15}\textbf{0.67} & \cellcolor{green!15}\textbf{0.81} & \cellcolor{green!15}\textbf{0.81} & \cellcolor{green!15}\textbf{0.93} & \cellcolor{green!15}\textbf{0.85} \\
           & RS~($\uparrow$) & \cellcolor{red!15}0.49 & \cellcolor{red!15}0.38 & \cellcolor{red!15}0.29 & \cellcolor{red!15}0.22 & \cellcolor{red!15}0.19 & \cellcolor{red!15}0.13 \\
\midrule
BITS       & CS~($\uparrow$) & \cellcolor{red!15}0.26 & \cellcolor{red!15}0.26 & \cellcolor{red!15}0.22 & \cellcolor{red!15}0.30 & \cellcolor{red!15}0.22 & \cellcolor{red!15}0.37 \\
           & RS~($\uparrow$) & \cellcolor{red!15}0.81 & \cellcolor{red!15}0.76 & \cellcolor{red!15}0.72 & \cellcolor{red!15}0.70 & \cellcolor{red!15}0.69 & \cellcolor{red!15}0.65 \\
\midrule
CTG        & CS~($\uparrow$) & \cellcolor{green!15}\textcolor{blue}{\textbf{0.44}} & \cellcolor{red!15}0.41 & \cellcolor{green!15}\textcolor{blue}{\textbf{0.56}} & \cellcolor{red!15}0.41 & \cellcolor{red!15}0.48 & \cellcolor{red!15}0.56 \\
           & RS~($\uparrow$) & \cellcolor{red!15}0.91 & \cellcolor{red!15}\textbf{\textcolor{blue}{0.89}} & \cellcolor{red!15}\textbf{\textcolor{blue}{0.85}} & \cellcolor{red!15}\textbf{\textcolor{blue}{0.84}} & \cellcolor{green!15}{\textbf{0.82}} & \cellcolor{green!15}\textbf{{0.77}} \\
\midrule
Ours       & CS~($\uparrow$) &  0.33 &  \textbf{\textcolor{blue}{0.48}} &  {0.44} &  \textbf{\textcolor{blue}{0.56}} &  \textbf{\textcolor{blue}{0.67}} &  \textbf{\textcolor{blue}{0.74}} \\
           & RS~($\uparrow$) &  \textbf{0.97} &  \textbf{0.94} &  \textbf{0.91} &  \textbf{0.88} &  \textbf{\textcolor{blue}{0.80}} &  \textbf{\textcolor{blue}{0.76}} \\
\bottomrule
\end{tabular}
}
\label{tab:spatial_cr_value}
\end{table}

\subsection{Long-horizon Closed-loop Generation~(RQ2)}
To address~\textbf{RQ2}, we evaluate \method's performance in long-horizon safety-critical scenario generation by changing the simulation frequency in $T$ seconds. 
We test the generation results with $T\in \{0.5\mathrm{s}, 1\mathrm{s}, 2\mathrm{s}, 3\mathrm{s}, 4\mathrm{s}, 5\mathrm{s}\}$ , which corresponds to a closed-loop simulation frequency between 0.2$\mathrm{Hz}$ to 2$\mathrm{Hz}$. 
We demonstrate the comparison of realism and controllability results in Table~\ref{tab:temporal_cr_value} and Figure~\ref{fig:cr_plot}(b). \method\ consistently outperform BITS and STRIVE in both \textbf{CS} and \textbf{RS} as the planning horizon enlarges. TrafficSim outperforms all the other baselines with the best controllability, while its realism in long-horizon generation is second-worst and only better than BITS. SimNet marginally outperforms \method\ in realism, yet its controllability is the worst among all. \method\ outperforms CTG with a comparable realism score and higher controllability at longer horizons. Figure~\ref{fig:cr_plot}(b) confirms our approach has the best proximity to the multi-objective Pareto frontier and has best coverage in the realistic zone. 
We also show qualitative studies for long-horizon generation in Appendix Figure~\ref{fig:first-qualitative}-\ref{fig:last-qualitative}. 

\begin{table}
\centering
\caption{Comparison in \textbf{RS} and \textbf{CS} in long-horizon scenario generation of 5 controllable agents over $T$ seconds. \colorbox{red!15}{Red} means \method\ outperforms the baseline in the corresponding metrics, \colorbox{green!15}{green} means the baseline is better, and \colorbox{yellow!30}{yellow} means a tie. \method\ has \textbf{best} and \textcolor{blue}{\textbf{second best}} performance in 10 out of 12 metrics. }
\resizebox{1.\linewidth}{!} {
\begin{tabular}{llccccc}
\toprule
\textbf{Method} & \textbf{Metric} &    {$T$=1$\mathrm{s}$} &    {2$\mathrm{s}$} &    {3$\mathrm{s}$} &    {4$\mathrm{s}$} &    {5$\mathrm{s}$} \\
\midrule
SimNet     & CS~($\uparrow$) & \cellcolor{red!15}0.15 & \cellcolor{red!15}0.22 & \cellcolor{red!15}0.22 & \cellcolor{red!15}0.22 & \cellcolor{red!15}0.33 \\
           & RS~($\uparrow$) & \cellcolor{green!15}\textbf{0.86} & \cellcolor{green!15}\textbf{0.78} & \cellcolor{green!15}\textbf{0.67} & \cellcolor{green!15}\textbf{0.60} & \cellcolor{green!15}\textcolor{blue}{\textbf{0.50}} \\
\midrule
TrafficSim & CS~($\uparrow$) & \cellcolor{green!15}\textbf{0.81} & \cellcolor{green!15}\textbf{0.81} & \cellcolor{green!15}\textbf{1.00} & \cellcolor{yellow!30}\textbf{0.89} & \cellcolor{green!15}\textbf{0.81} \\
           & RS~($\uparrow$) & \cellcolor{red!15}0.66 & \cellcolor{red!15}0.46 & \cellcolor{red!15}0.40 & \cellcolor{red!15}0.26 & \cellcolor{red!15}0.28 \\
\midrule
STRIVE     & CS~($\uparrow$) & \cellcolor{green!15}0.67 & \cellcolor{red!15}0.37 & \cellcolor{red!15}0.37 & \cellcolor{red!15}0.30 & \cellcolor{red!15}0.26 \\
           & RS~($\uparrow$) & \cellcolor{red!15}0.49 & \cellcolor{red!15}0.36 & \cellcolor{red!15}0.17 & \cellcolor{red!15}0.12 & \cellcolor{red!15}0.22 \\
\midrule
BITS       & CS~($\uparrow$) & \cellcolor{red!15}0.22 & \cellcolor{red!15}0.11 & \cellcolor{red!15}0.48 & \cellcolor{red!15}0.30 & \cellcolor{red!15}0.37 \\
           & RS~($\uparrow$) & \cellcolor{red!15}0.70 & \cellcolor{red!15}0.58 & \cellcolor{red!15}0.47 & \cellcolor{red!15}0.39 & \cellcolor{red!15}0.35 \\
\midrule
CTG        & CS~($\uparrow$) & \cellcolor{green!15}0.37 & \cellcolor{yellow!30}\textcolor{blue}{\textbf{0.48}} & \cellcolor{yellow!30}\textcolor{blue}{\textbf{0.67}} & \cellcolor{red!15}0.78 & \cellcolor{red!15}0.67 \\
           & RS~($\uparrow$) & \cellcolor{yellow!30}\textcolor{blue}{\textbf{0.81}} & \cellcolor{red!15}0.68 & \cellcolor{red!15}0.59 & \cellcolor{red!15}0.53 & \cellcolor{green!15}\textbf{0.53} \\
\midrule
CCDiff     & CS~($\uparrow$) & 0.33 & \textcolor{blue}{\textbf{0.48}} & \textcolor{blue}{\textbf{0.67}} & \textbf{0.89} & \textcolor{blue}{\textbf{0.78}} \\
           & RS~($\uparrow$) & \textcolor{blue}{\textbf{0.81}} & \textcolor{blue}{\textbf{0.74}} & \textcolor{blue}{\textbf{0.63}} & \textcolor{blue}{\textbf{0.54}} & 0.49 \\
\bottomrule
\end{tabular}
}
\label{tab:temporal_cr_value}
\end{table}

\newcommand{\best}[1]{\textbf{{#1}}}
\newcommand{\second}[1]{\textbf{\textcolor{blue}{#1}}}

\begin{table}[h!]
\caption{Ablation study on CCDiff's variants. Evaluation of Controllability (CO, OR) and Realism (FDE and CFD) over different agent scales. For each metric we highlight the \textbf{best} and the \textcolor{blue}{\textbf{second best}} results.}
\label{tab:ablation_variants}
\centering
\resizebox{1.\linewidth}{!} {
\begin{tabular}{cccccccccc}
\toprule
\textbf{Enc.} & \textbf{Guide}  & \textbf{Rank} & \textbf{Metrics} & $K$=1 & 2 & 3 & 4 & 5 \\
\midrule
& &  & SCR ($\uparrow$) & \textbf{0.32} & \textbf{0.43} & \textbf{0.44} & \textbf{0.43} & \second{0.42} \\
            & \checkmark & \checkmark & ORR ($\downarrow$) & 0.53 & 1.10 & 0.98 & \textbf{0.91} & \textbf{0.91} \\
           &            &  & FDE ($\downarrow$) & \second{2.18} & \textbf{4.00} & 5.41 & \second{5.87} & \textbf{5.79} \\
           &            &  & CFD ($\downarrow$) & \textbf{1.09} & \textbf{1.00} & \textbf{1.14} & \textbf{1.22} & \textbf{1.22} \\
\midrule   
&   &  & SCR ($\uparrow$) & 0.31 & 0.38 & 0.45 & 0.40 & 0.40 \\
 \checkmark &  & \checkmark  & ORR ($\downarrow$) & \textbf{0.33} & \second{0.81} & \second{0.76} & 1.00 & 1.06 \\
           &            &  & FDE ($\downarrow$) & 2.21 & 4.33 & 5.28 & 6.13 & 6.82 \\
           &            &  & CFD ($\downarrow$) & {1.51} & {1.81} & \second{1.60} & \second{1.84} & \second{1.94} \\
\midrule      
& & & SCR ($\uparrow$) & \textbf{0.32} & 0.33 & 0.34 & 0.36 & 0.37 \\
           \checkmark & \checkmark & \text{Dist} & ORR ($\downarrow$) & 0.63 & 1.38 & 1.50 & 1.59 & \second{1.49} \\
           &  &  & FDE ($\downarrow$) & 2.80 & \second{4.15} & \textbf{5.15} & 5.79 & 5.96 \\
           &  &  & CFD ($\downarrow$) & \second{1.24} & \second{1.79} & 2.44 & 2.03 & 2.34 \\
\midrule
 &  &  & SCR ($\uparrow$) & 0.28 & 0.34 & 0.35 & 0.33 & 0.31 \\
 \checkmark  & \checkmark & \text{Human} & ORR ($\downarrow$) & 0.67 & 1.66 & 1.65 & 1.73 & 1.93 \\
 &           &  & FDE ($\downarrow$) & 3.13 & 5.80 & 6.74 & 7.40 & 7.84 \\
 &           &  & CFD ($\downarrow$) & {1.31} & 2.21 & 2.51 & 2.83 & 3.14 \\
\midrule                              
& & & SCR ($\uparrow$) & 0.29 & \second{0.40} & \textbf{0.44} & \textbf{0.43} & \textbf{0.46} \\
           \checkmark & \checkmark & \checkmark    & ORR ($\downarrow$) & \second{0.37} & \textbf{0.61} & \textbf{0.72} & \second{0.99} & \second{1.02} \\
           &            &            & FDE ($\downarrow$) & \textbf{2.16} & {4.17} & \second{5.22} & \textbf{5.99} & 6.59 \\
           &            &            & CFD ($\downarrow$) & 1.70 & 1.88 & 1.92 & 1.93 & 2.25 \\
\bottomrule
\end{tabular}
}
\vspace{-5mm}
\end{table}

\subsection{Ablation study~(RQ3)}
To answer~\textbf{RQ3}, we evaluate our methods with different ablation variants related to the causal composition, including (i) \method\ w/o encoder, which removes sparsity constraints $\lambda_{\text{sparsity}}$ of the causal composition scene encoder, (ii) \method\ w/o factored guide, which replaces the factorized guidance with the whole state space guidance, (iii) \method\ w/ human and (iv) \method\ w/ distance which replace the causal ranking algorithms with distance-based ranking and human ranking. We demonstrate the quantitative results in Table~\ref{tab:ablation_variants}. 
The non-causal guidance variants and non-causal encoder variants show a performance drop in the collision rate~(controllability), yet the w/o encoder variants outperform in kinematic comfort. 
Among all the ablation variants, different ranking strategies result in the \textit{largest} performance drop for the multi-agent controllable generation settings. Compared to our causal ranking, human ranking and distance-based ranking strategy suffers from a performance drop with 5 to 10\% in the collision rate and, more than 0.5\% in the off-road rate, 1$\mathrm{m}$ for FDE, and also larger CFD. This signifies the importance of applying the proper guidance to the correct agents at the traffic scene when doing safety-critical generation.

\section{Conclusion}
In this paper, we propose \method, a causal composition diffusion model that aims to improve the \textit{controllability} and \textit{realism} in closed-loop safety-critical scenario generation for autonomous driving. Based on the formulation of constrained factored MDP, \method\ promotes realism by first identifying the underlying causal structure between agents, then incorporating it in the scene encoder and ranking the importance of agents based on causal knowledge. 
\method\ uses both interventional classifier-free guidance and masked classifier guidance to improve controllability in safety-critical scenario generation. 
In multi-agent generation and long-horizon generation settings, \method\ outperforms SOTA methods over nuScenes data in closed-loop evaluation. One limitation of the current work is that the design of the causal reasoning pipeline relies on hyperparameter tuning and it is hard to directly evaluate. It would be interesting to construct a traffic reasoning benchmark and incorporate a foundation model to further scale up the traffic reasoning and generation process.

\clearpage
\section*{Acknowledgment}
We acknowledge Paul Vernaza for his insightful and valuable discussion with the authors. 

\bibliography{reference}

\begin{thebibliography}{10}

\bibitem{feng2023dense}
Shuo Feng, Haowei Sun, Xintao Yan, Haojie Zhu, Zhengxia Zou, Shengyin Shen, and Henry~X Liu.
\newblock Dense reinforcement learning for safety validation of autonomous vehicles.
\newblock {\em Nature}, 615(7953):620--627, 2023.

\bibitem{ding2023survey}
Wenhao Ding, Chejian Xu, Mansur Arief, Haohong Lin, Bo~Li, and Ding Zhao.
\newblock A survey on safety-critical driving scenario generation—a methodological perspective.
\newblock {\em IEEE Transactions on Intelligent Transportation Systems}, 2023.

\bibitem{zhong2023guided}
Ziyuan Zhong, Davis Rempe, Danfei Xu, Yuxiao Chen, Sushant Veer, Tong Che, Baishakhi Ray, and Marco Pavone.
\newblock Guided conditional diffusion for controllable traffic simulation.
\newblock In {\em 2023 IEEE International Conference on Robotics and Automation (ICRA)}, pages 3560--3566. IEEE, 2023.

\bibitem{zhong2023language}
Ziyuan Zhong, Davis Rempe, Yuxiao Chen, Boris Ivanovic, Yulong Cao, Danfei Xu, Marco Pavone, and Baishakhi Ray.
\newblock Language-guided traffic simulation via scene-level diffusion.
\newblock In {\em Conference on Robot Learning}, pages 144--177. PMLR, 2023.

\bibitem{ding2023realgen}
Wenhao Ding, Yulong Cao, Ding Zhao, Chaowei Xiao, and Marco Pavone.
\newblock Realgen: Retrieval augmented generation for controllable traffic scenarios.
\newblock {\em arXiv preprint arXiv:2312.13303}, 2023.

\bibitem{hu2023simulation}
Xuemin Hu, Shen Li, Tingyu Huang, Bo~Tang, Rouxing Huai, and Long Chen.
\newblock How simulation helps autonomous driving: A survey of sim2real, digital twins, and parallel intelligence.
\newblock {\em IEEE Transactions on Intelligent Vehicles}, 2023.

\bibitem{dosovitskiy2017carla}
Alexey Dosovitskiy, German Ros, Felipe Codevilla, Antonio Lopez, and Vladlen Koltun.
\newblock Carla: An open urban driving simulator.
\newblock In {\em Conference on robot learning}, pages 1--16. PMLR, 2017.

\bibitem{lopez2018microscopic}
Pablo~Alvarez Lopez, Michael Behrisch, Laura Bieker-Walz, Jakob Erdmann, Yun-Pang Fl{\"o}tter{\"o}d, Robert Hilbrich, Leonhard L{\"u}cken, Johannes Rummel, Peter Wagner, and Evamarie Wie{\ss}ner.
\newblock Microscopic traffic simulation using sumo.
\newblock In {\em 2018 21st international conference on intelligent transportation systems (ITSC)}, pages 2575--2582. IEEE, 2018.

\bibitem{wang2021advsim}
Jingkang Wang, Ava Pun, James Tu, Sivabalan Manivasagam, Abbas Sadat, Sergio Casas, Mengye Ren, and Raquel Urtasun.
\newblock Advsim: Generating safety-critical scenarios for self-driving vehicles.
\newblock In {\em Proceedings of the IEEE/CVF Conference on Computer Vision and Pattern Recognition}, pages 9909--9918, 2021.

\bibitem{tan2021scenegen}
Shuhan Tan, Kelvin Wong, Shenlong Wang, Sivabalan Manivasagam, Mengye Ren, and Raquel Urtasun.
\newblock Scenegen: Learning to generate realistic traffic scenes.
\newblock In {\em Proceedings of the IEEE/CVF Conference on Computer Vision and Pattern Recognition}, pages 892--901, 2021.

\bibitem{ding2023causalaf}
Wenhao Ding, Haohong Lin, Bo~Li, and Ding Zhao.
\newblock Causalaf: Causal autoregressive flow for safety-critical driving scenario generation.
\newblock In {\em Conference on robot learning}, pages 812--823. PMLR, 2023.

\bibitem{patrikar2024rulefuser}
Jay Patrikar, Sushant Veer, Apoorva Sharma, Marco Pavone, and Sebastian Scherer.
\newblock Rulefuser: Injecting rules in evidential networks for robust out-of-distribution trajectory prediction.
\newblock {\em arXiv preprint arXiv:2405.11139}, 2024.

\bibitem{tan2023language}
Shuhan Tan, Boris Ivanovic, Xinshuo Weng, Marco Pavone, and Philipp Kraehenbuehl.
\newblock Language conditioned traffic generation.
\newblock In {\em Conference on Robot Learning}, pages 2714--2752. PMLR, 2023.

\bibitem{caesar2019nuscenes}
H~Caesar, V~Bankiti, AH~Lang, S~Vora, VE~Liong, Q~Xu, A~Krishnan, Y~Pan, G~Baldan, and O~Beijbom.
\newblock nuscenes: A multimodal dataset for autonomous driving. arxiv.
\newblock 2019.

\bibitem{liu2022towards}
Yuejiang Liu, Riccardo Cadei, Jonas Schweizer, Sherwin Bahmani, and Alexandre Alahi.
\newblock Towards robust and adaptive motion forecasting: A causal representation perspective.
\newblock In {\em Proceedings of the IEEE/CVF Conference on Computer Vision and Pattern Recognition}, pages 17081--17092, 2022.

\bibitem{ge2023causal}
Chunjiang Ge, Shiji Song, and Gao Huang.
\newblock Causal intervention for human trajectory prediction with cross attention mechanism.
\newblock In {\em Proceedings of the AAAI Conference on Artificial Intelligence}, volume~37, pages 658--666, 2023.

\bibitem{chen2021human}
Guangyi Chen, Junlong Li, Jiwen Lu, and Jie Zhou.
\newblock Human trajectory prediction via counterfactual analysis.
\newblock In {\em Proceedings of the IEEE/CVF International Conference on Computer Vision}, pages 9824--9833, 2021.

\bibitem{pourkeshavarz2024cadet}
Mozhgan Pourkeshavarz, Junrui Zhang, and Amir Rasouli.
\newblock Cadet: a causal disentanglement approach for robust trajectory prediction in autonomous driving.
\newblock In {\em Proceedings of the IEEE/CVF Conference on Computer Vision and Pattern Recognition}, pages 14874--14884, 2024.

\bibitem{ettinger2021large}
Scott Ettinger, Shuyang Cheng, Benjamin Caine, Chenxi Liu, Hang Zhao, Sabeek Pradhan, Yuning Chai, Ben Sapp, Charles~R Qi, Yin Zhou, et~al.
\newblock Large scale interactive motion forecasting for autonomous driving: The waymo open motion dataset.
\newblock In {\em Proceedings of the IEEE/CVF International Conference on Computer Vision}, pages 9710--9719, 2021.

\bibitem{roelofs2022causalagents}
Rebecca Roelofs, Liting Sun, Ben Caine, Khaled~S Refaat, Ben Sapp, Scott Ettinger, and Wei Chai.
\newblock Causalagents: A robustness benchmark for motion forecasting using causal relationships.
\newblock {\em arXiv preprint arXiv:2207.03586}, 2022.

\bibitem{ngiam2021scene}
Jiquan Ngiam, Benjamin Caine, Vijay Vasudevan, Zhengdong Zhang, Hao-Tien~Lewis Chiang, Jeffrey Ling, Rebecca Roelofs, Alex Bewley, Chenxi Liu, Ashish Venugopal, et~al.
\newblock Scene transformer: A unified architecture for predicting multiple agent trajectories.
\newblock {\em arXiv preprint arXiv:2106.08417}, 2021.

\bibitem{zhou2022hivt}
Zikang Zhou, Luyao Ye, Jianping Wang, Kui Wu, and Kejie Lu.
\newblock Hivt: Hierarchical vector transformer for multi-agent motion prediction.
\newblock In {\em Proceedings of the IEEE/CVF Conference on Computer Vision and Pattern Recognition}, pages 8823--8833, 2022.

\bibitem{zhou2023query}
Zikang Zhou, Jianping Wang, Yung-Hui Li, and Yu-Kai Huang.
\newblock Query-centric trajectory prediction.
\newblock In {\em Proceedings of the IEEE/CVF Conference on Computer Vision and Pattern Recognition}, pages 17863--17873, 2023.

\bibitem{nayakanti2023wayformer}
Nigamaa Nayakanti, Rami Al-Rfou, Aurick Zhou, Kratarth Goel, Khaled~S Refaat, and Benjamin Sapp.
\newblock Wayformer: Motion forecasting via simple \& efficient attention networks.
\newblock In {\em 2023 IEEE International Conference on Robotics and Automation (ICRA)}, pages 2980--2987. IEEE, 2023.

\bibitem{dyro2024realistic}
Robert Dyro, Matthew Foutter, Ruolin Li, Luigi Di~Lillo, Edward Schmerling, Xilin Zhou, and Marco Pavone.
\newblock Realistic extreme behavior generation for improved av testing.
\newblock {\em arXiv preprint arXiv:2409.10669}, 2024.

\bibitem{suo2021trafficsim}
Simon Suo, Sebastian Regalado, Sergio Casas, and Raquel Urtasun.
\newblock Trafficsim: Learning to simulate realistic multi-agent behaviors.
\newblock In {\em Proceedings of the IEEE/CVF Conference on Computer Vision and Pattern Recognition}, pages 10400--10409, 2021.

\bibitem{bergamini2021simnet}
Luca Bergamini, Yawei Ye, Oliver Scheel, Long Chen, Chih Hu, Luca Del~Pero, B{\l}a{\.z}ej Osi{\'n}ski, Hugo Grimmett, and Peter Ondruska.
\newblock Simnet: Learning reactive self-driving simulations from real-world observations.
\newblock In {\em 2021 IEEE International Conference on Robotics and Automation (ICRA)}, pages 5119--5125. IEEE, 2021.

\bibitem{rempe2022generating}
Davis Rempe, Jonah Philion, Leonidas~J Guibas, Sanja Fidler, and Or~Litany.
\newblock Generating useful accident-prone driving scenarios via a learned traffic prior.
\newblock In {\em Proceedings of the IEEE/CVF Conference on Computer Vision and Pattern Recognition}, pages 17305--17315, 2022.

\bibitem{xu2023bits}
Danfei Xu, Yuxiao Chen, Boris Ivanovic, and Marco Pavone.
\newblock Bits: Bi-level imitation for traffic simulation.
\newblock In {\em 2023 IEEE International Conference on Robotics and Automation (ICRA)}, pages 2929--2936. IEEE, 2023.

\bibitem{song2019generative}
Yang Song and Stefano Ermon.
\newblock Generative modeling by estimating gradients of the data distribution.
\newblock {\em Advances in neural information processing systems}, 32, 2019.

\bibitem{ho2020denoising}
Jonathan Ho, Ajay Jain, and Pieter Abbeel.
\newblock Denoising diffusion probabilistic models.
\newblock {\em Advances in neural information processing systems}, 33:6840--6851, 2020.

\bibitem{ho2022classifier}
Jonathan Ho and Tim Salimans.
\newblock Classifier-free diffusion guidance.
\newblock {\em arXiv preprint arXiv:2207.12598}, 2022.

\bibitem{pronovost2023scenario}
Ethan Pronovost, Meghana~Reddy Ganesina, Noureldin Hendy, Zeyu Wang, Andres Morales, Kai Wang, and Nick Roy.
\newblock Scenario diffusion: Controllable driving scenario generation with diffusion.
\newblock {\em Advances in Neural Information Processing Systems}, 36:68873--68894, 2023.

\bibitem{jiang2023motiondiffuser}
Chiyu Jiang, Andre Cornman, Cheolho Park, Benjamin Sapp, Yin Zhou, Dragomir Anguelov, et~al.
\newblock Motiondiffuser: Controllable multi-agent motion prediction using diffusion.
\newblock In {\em Proceedings of the IEEE/CVF Conference on Computer Vision and Pattern Recognition}, pages 9644--9653, 2023.

\bibitem{jiangscenediffuser}
Chiyu~Max Jiang, Yijing Bai, Andre Cornman, Christopher Davis, Xiukun Huang, Hong Jeon, Sakshum Kulshrestha, John~Wheatley Lambert, Shuangyu Li, Xuanyu Zhou, et~al.
\newblock Scenediffuser: Efficient and controllable driving simulation initialization and rollout.
\newblock In {\em The Thirty-eighth Annual Conference on Neural Information Processing Systems}.

\bibitem{xu2023diffscene}
Chejian Xu, Ding Zhao, Alberto Sangiovanni-Vincentelli, and Bo~Li.
\newblock Diffscene: Diffusion-based safety-critical scenario generation for autonomous vehicles.
\newblock In {\em The Second Workshop on New Frontiers in Adversarial Machine Learning}, 2023.

\bibitem{chang2023controllable}
Wei-Jer Chang, Francesco Pittaluga, Masayoshi Tomizuka, Wei Zhan, and Manmohan Chandraker.
\newblock Controllable safety-critical closed-loop traffic simulation via guided diffusion.
\newblock {\em arXiv preprint arXiv:2401.00391}, 2023.

\bibitem{xie2024advdiffuser}
Yuting Xie, Xianda Guo, Cong Wang, Kunhua Liu, and Long Chen.
\newblock Advdiffuser: Generating adversarial safety-critical driving scenarios via guided diffusion.
\newblock {\em arXiv preprint arXiv:2410.08453}, 2024.

\bibitem{huang2024versatile}
Zhiyu Huang, Zixu Zhang, Ameya Vaidya, Yuxiao Chen, Chen Lv, and Jaime~Fern{\'a}ndez Fisac.
\newblock Versatile scene-consistent traffic scenario generation as optimization with diffusion.
\newblock {\em arXiv preprint arXiv:2404.02524}, 2024.

\bibitem{grimbly2021causal}
St~John Grimbly, Jonathan Shock, and Arnu Pretorius.
\newblock Causal multi-agent reinforcement learning: Review and open problems.
\newblock {\em arXiv preprint arXiv:2111.06721}, 2021.

\bibitem{janner2022planning}
Michael Janner, Yilun Du, Joshua~B Tenenbaum, and Sergey Levine.
\newblock Planning with diffusion for flexible behavior synthesis.
\newblock {\em arXiv preprint arXiv:2205.09991}, 2022.

\bibitem{ajay2022conditional}
Anurag Ajay, Yilun Du, Abhi Gupta, Joshua Tenenbaum, Tommi Jaakkola, and Pulkit Agrawal.
\newblock Is conditional generative modeling all you need for decision-making?
\newblock {\em arXiv preprint arXiv:2211.15657}, 2022.

\bibitem{levine2018reinforcement}
Sergey Levine.
\newblock Reinforcement learning and control as probabilistic inference: Tutorial and review.
\newblock {\em arXiv preprint arXiv:1805.00909}, 2018.

\bibitem{yu2020gradient}
Tianhe Yu, Saurabh Kumar, Abhishek Gupta, Sergey Levine, Karol Hausman, and Chelsea Finn.
\newblock Gradient surgery for multi-task learning.
\newblock {\em Advances in Neural Information Processing Systems}, 33:5824--5836, 2020.

\bibitem{dinh2023pixelasparam}
Anh-Dung Dinh, Daochang Liu, and Chang Xu.
\newblock Pixelasparam: A gradient view on diffusion sampling with guidance.
\newblock In {\em International Conference on Machine Learning}, pages 8120--8137. PMLR, 2023.

\bibitem{yao2024gradient}
Yihang Yao, Zuxin Liu, Zhepeng Cen, Peide Huang, Tingnan Zhang, Wenhao Yu, and Ding Zhao.
\newblock Gradient shaping for multi-constraint safe reinforcement learning.
\newblock In {\em 6th Annual Learning for Dynamics \& Control Conference}, pages 25--39. PMLR, 2024.

\bibitem{ma2024decouple}
Qianli Ma, Xuefei Ning, Dongrui Liu, Li~Niu, and Linfeng Zhang.
\newblock Decouple-then-merge: Towards better training for diffusion models.
\newblock {\em arXiv preprint arXiv:2410.06664}, 2024.

\bibitem{achiam2017constrained}
Joshua Achiam, David Held, Aviv Tamar, and Pieter Abbeel.
\newblock Constrained policy optimization.
\newblock In {\em International Conference on Machine Learning}, pages 22--31. PMLR, 2017.

\bibitem{bahmani2016learning}
Sohail Bahmani, Petros~T Boufounos, and Bhiksha Raj.
\newblock Learning model-based sparsity via projected gradient descent.
\newblock {\em IEEE Transactions on Information Theory}, 62(4):2092--2099, 2016.

\bibitem{brouillard2020differentiable}
Philippe Brouillard, S{\'e}bastien Lachapelle, Alexandre Lacoste, Simon Lacoste-Julien, and Alexandre Drouin.
\newblock Differentiable causal discovery from interventional data.
\newblock {\em Advances in Neural Information Processing Systems}, 33:21865--21877, 2020.

\bibitem{scherrer2021learning}
Nino Scherrer, Olexa Bilaniuk, Yashas Annadani, Anirudh Goyal, Patrick Schwab, Bernhard Sch{\"o}lkopf, Michael~C Mozer, Yoshua Bengio, Stefan Bauer, and Nan~Rosemary Ke.
\newblock Learning neural causal models with active interventions.
\newblock {\em arXiv preprint arXiv:2109.02429}, 2021.

\bibitem{pearl2009causality}
Judea Pearl.
\newblock {\em Causality}.
\newblock Cambridge university press, 2009.

\bibitem{coello2007evolutionary}
Carlos A~Coello Coello.
\newblock {\em Evolutionary algorithms for solving multi-objective problems}.
\newblock Springer, 2007.

\end{thebibliography}
\bibliographystyle{unsrt}

\clearpage
\pagenumbering{arabic}
\setcounter{page}{1}

\onecolumn
\appendix
\section*{\centering {\Large \textit{CCDiff}: Causal Composition Diffusion Model for Closed-loop Traffic Generation \\
(Supplementary Materials)}}
\section{Additional Related Works}
\begin{table*}[htbp]
\vspace{-3mm}
    \caption{Key features of related works in scenario generation for autonomous vehicles. }
    \vspace{-3mm}
    \centering
    \begin{tabular}{ m{3cm} | c c c c c }
        \toprule
        \textbf{Paper} & \textbf{Controllability} & \textbf{Realism} & \textbf{Closed-loop} & \textbf{Safety-Critical} & \textbf{Compositionality} \\ \midrule
        TrafficSim~\cite{suo2021trafficsim} & \checkmark & \checkmark & \checkmark & \ding{55} & \ding{55} \\
        BITS~\cite{xu2023bits} & \checkmark & \checkmark & \checkmark & \ding{55} & \ding{55} \\
        SimNet~\cite{bergamini2021simnet} & \checkmark & \checkmark & \checkmark & \ding{55} & \ding{55} \\
        STRIVE~\cite{rempe2022generating} & \checkmark & \checkmark & \checkmark & \checkmark & \ding{55} \\
        CTG~\cite{zhong2023guided} & \checkmark & \checkmark & \checkmark & \ding{55} & STL \\
        SceneGen~\cite{tan2021scenegen} & \checkmark & \checkmark & \ding{55} & \ding{55} & \ding{55} \\
        RealGen~\cite{ding2023realgen} & \checkmark & \checkmark & \ding{55} & \checkmark & \ding{55} \\
        CTG++~\cite{zhong2023language} & \checkmark & \checkmark & \checkmark & \checkmark & LLM \\
        LCTGen~\cite{tan2023language} & \checkmark & \checkmark & \ding{55} & \ding{55} & LLM \\
        CausalAF~\cite{ding2023causalaf} & \checkmark & \checkmark & \ding{55} & \checkmark & CG \\
        Ours & \checkmark & \checkmark & \checkmark & \checkmark & CG \\
        \bottomrule
    \end{tabular}
    \label{tab:paper_review}
\end{table*}

\vspace{-5mm}
\section{Additional Algorithm Details}
Algorithm~\ref{alg:training} presents the training of \method\ similar to DDPM and outputs denoising scene encoder $\pi_{\phi, \psi}(\cdot | \bm{s}, \bm{c}; G)$. 
Algorithm~\ref{alg:causal_cluster} presents the causal discovery and ranking algorithm


\begin{algorithm}
\caption{Training of \method}\label{alg:training}
\begin{algorithmic}
\Require Dropout $p_{\text{uncond}}$, threshold $C_\rho, C_{\text{ttc}}$
\Require Guidance loss $\{\mathcal{J}_i\}_{i=1}^N$, trajectories $\bm{\tau}$, map $\bm{c}$. 

\While{$M \leq M_{\text{max}}$} 
    \State $M\gets M+1$
    \State $(\bm{\tau}(0), \bm{c}) \sim \mathcal{D}$ 
    \State $\textcolor{blue}{G}\gets \textcolor{blue}{G}(\bm{\tau}(0))$ with probability $1-p_{\text{uncond}}$
    \State $\textcolor{blue}{G}\gets \bm{I}_N$ with probability $p_{\text{uncond}}$
    \State $k \sim \text{Unif}[K]$ 
    \State $\bm{\tau}(k) = \sqrt{\overline{\alpha}_k} \bm{\tau}(0) + \sqrt{1 - \overline{\alpha}_k} \bm{\epsilon}$ 
    \State Update $\pi_{\phi,\psi}$ with $\nabla_{\phi,\psi} \left\|\pi_{\phi, \psi}(\bm{\tau}(k), \bm{c}, k; \textcolor{blue}{G}) - \bm{a}(0) \right\|^2$
\EndWhile
\\
\Return Denoising scene encoder $\pi_{\phi, \psi}(\cdot | \bm{s}, \bm{c}; G)$
\end{algorithmic}
\end{algorithm}

\vspace{-5mm}
\begin{algorithm}
\caption{Causal discovery and Ranking for \method}\label{alg:causal_cluster}
\begin{algorithmic}
\Require History trajectories $\bm{\tau}$, TTC Graph $M$, attention matrix $\bm{\alpha}$, Top-K agents $k$
\State $G = M\cdot \bm{\alpha}\triangleq (V, E, w)$
\ForAll{$v_i\in G$}
    \State $C_i\gets \{v_i\}, w_i = 0$
    \ForAll{$v_j\in V\backslash C_i$}
        \If{$(v_j, v)\in E, \forall v\in C_i$} 
            \State $w_i \gets w_i + \sum_{v\in C_i}w(v_j, v)$
            \State $C_i\gets C_i\cup \{v_j\}$
        \EndIf
    \EndFor
\EndFor
\State $\bm{\rho} \gets \text{argsort}(C, w)$[:k]
\\
\Return Importance ranking $\bm{\rho}$
\end{algorithmic}
\end{algorithm}

\clearpage
\newpage
\section{Additional Experiment Details}
\subsection{Additional Quantitative Results}
\label{app:quantitative_results}


\begin{table*}[h!]
\caption{Evaluation of Controllability and Realism across different scales of editable agents ($N$) and planning horizons ($T$). For each metric, we report the \textbf{best} and \textcolor{blue}{\textbf{second best}} performance among all the methods. CCDiff has the best overall performance presented in the main text. }
\centering
{\small
\begin{tabular}{lc|cccccc|ccccc}
\toprule
\textbf{Methods} & \textbf{Metrics} & $K$=2 & {3} & {4} & {5} & {10} & {Full} & $T$=1$\mathrm{s}$ & 2$\mathrm{s}$ & 3$\mathrm{s}$ & 4$\mathrm{s}$ & 5$\mathrm{s}$ \\
\midrule
\multirow{4}{*}{SimNet} 
    & SCR ($\uparrow$) & 0.31 & 0.32 & 0.33 & 0.36 & 0.42 & 0.47 & 0.35 & 0.37 & 0.37 & 0.37 & 0.40 \\
    & ORR ($\downarrow$) & 1.76 & 2.19 & 2.62 & 2.67 & 2.90 & 3.17 & \second{2.09} & \second{3.87} & \second{6.16} & \second{8.36} & \second{9.93} \\
    & FDE ($\downarrow$) & \second{3.76} & \second{4.34} & \second{4.98} & \second{5.26} & \second{6.63} & 8.03 & \textbf{4.11} & \textbf{3.78} & \textbf{4.90} & \textbf{4.83} & \textbf{3.83} \\
    & CFD ($\downarrow$) & 2.56 & 2.95 & 2.86 & 3.16 & 5.00 & 7.00 & 4.02 & 5.03 & 5.51 & 5.57 & 8.04 \\
\midrule
\multirow{4}{*}{TrafficSim} 
    & SCR ($\uparrow$) & 0.38 & 0.36 & 0.44 & 0.41 & 0.41 & 0.47 & \textbf{0.53} & \textbf{0.53} & \textbf{0.58} & \textbf{0.55} & \textbf{0.53} \\
    & ORR ($\downarrow$) & 2.09 & 2.25 & 2.45 & 2.48 & 2.66 & 2.73 & 3.56 & 6.36 & 8.98 & 10.96 & 12.21 \\
    & FDE ($\downarrow$) & 4.25 & 5.06 & 5.77 & 6.23 & 6.79 & \second{7.13} & 8.32 & 6.48 & 8.61 & 8.66 & 7.32 \\
    & CFD ($\downarrow$) & 7.76 & 9.53 & 10.64 & 10.99 & 10.96 & 11.57 & 5.00 & 10.06 & 7.89 & 10.38 & 9.90 \\
\midrule
\multirow{4}{*}{STRIVE} 
    & SCR ($\uparrow$) & \textbf{0.49} & \textbf{0.49} & \textbf{0.53} & \textbf{0.53} & \textbf{0.56} & \textbf{0.54} & 0.49 & 0.41 & 0.41 & 0.39 & 0.38 \\
    & ORR ($\downarrow$) & 5.70 & 6.45 & 7.13 & 7.50 & 8.04 & 8.53 & 5.75 & 4.98 & 6.64 & 8.40 & 10.02 \\
    & FDE ($\downarrow$) & 9.01 & 10.79 & 12.13 & 13.00 & 13.76 & 14.52 & 11.48 & 11.20 & 14.56 & 15.00 & 12.41 \\
    & CFD ($\downarrow$) & 7.72 & 8.93 & 9.91 & 10.67 & 10.72 & 11.21 & 5.60 & 10.21 & 11.59 & 11.32 & 9.11 \\
\midrule
\multirow{4}{*}{BITS} 
    & SCR ($\uparrow$) & 0.38 & 0.38 & 0.37 & 0.39 & 0.37 & 0.41 & 0.37 & 0.34 & 0.44 & 0.39 & 0.41 \\
    & ORR ($\downarrow$) & \textbf{0.53} & \textbf{0.51} & \textbf{0.56} & \textbf{0.63} & \textbf{0.56} & \textbf{0.60} & \textbf{1.44} & \textbf{3.68} & \textbf{5.63} & \textbf{7.56} & \textbf{9.39} \\
    & FDE ($\downarrow$) & \textbf{3.20} & \textbf{3.95} & \textbf{4.42} & \textbf{4.67} & \textbf{5.05} & \textbf{5.35} & \second{4.68} & \second{4.69} & \second{6.10} & \second{6.36} & \second{5.44} \\
    & CFD ($\downarrow$) & 7.43 & 8.32 & 9.15 & 9.42 & 9.46 & 10.23 & 8.79 & 10.35 & 10.75 & 11.30 & 11.65 \\
\midrule
\multirow{4}{*}{CTG} 
    & SCR ($\uparrow$) & \second{0.43} & 0.42 & \second{0.46} & 0.42 & 0.44 & 0.46 & \second{0.41} & \second{0.44} & \second{0.49} & 0.52 & 0.49 \\
    & ORR ($\downarrow$) & 1.00 & 1.04 & 1.10 & 1.09 & \second{1.12} & \second{1.23} & 1.91 & 4.58 & 7.13 & 9.04 & 10.71 \\
    & FDE ($\downarrow$) & 5.32 & 6.18 & 6.83 & 7.40 & 8.10 & 9.19 & 7.58 & 7.91 & 10.26 & 10.30 & 8.27 \\
    & CFD ($\downarrow$) & \second{2.37} & \second{2.31} & \second{2.68} & \second{2.59} & \textbf{2.57} & \textbf{3.13} & 2.68 & \textbf{4.06} & \textbf{2.43} & \textbf{2.80} & \textbf{3.00} \\
\midrule
\multirow{4}{*}{Ours} 
    & SCR ($\uparrow$) & 0.40 & \second{0.44} & 0.43 & \second{0.46} & \second{0.49} & \second{0.51} & 0.40 & \second{0.44} & \second{0.49} & \second{0.55} & \second{0.52} \\
    & ORR ($\downarrow$) & \second{0.61} & \second{0.72} & \second{0.99} & \second{1.02} & 1.80 & 2.05 & 2.92 & 4.52 & 7.10 & 9.35 & 10.51 \\
    & FDE ($\downarrow$) & 4.17 & 5.22 & 5.99 & 6.59 & 7.84 & 8.26 & 7.06 & 5.54 & 6.86 & 7.00 & 5.71 \\
    & CFD ($\downarrow$) & \textbf{1.88} & \textbf{1.92} & \textbf{1.93} & \textbf{2.25} & \second{2.83} & \second{3.47} & \textbf{2.37} & \second{4.08} & \second{4.25} & \second{4.97} & \second{6.33} \\
\bottomrule
\end{tabular}
}
\label{tab:combined_evaluation}
\end{table*}

\begin{table}[h!]
\caption{Ablation study on CCDiff's variants. Evaluation of Controllability (CO, OR) and Realism (FDE and CFD) over different agent scales. For each metric, we highlight the \textbf{best} and the \textcolor{blue}{\textbf{second best}} results. Causal ranking has the greatest impact to the final performance. }
\label{tab:ablation_variants}
\centering
{\scriptsize
\begin{tabular}{cccc|cccccc|ccccc}
\toprule
\textbf{Enc.} & \textbf{Guide}  & \textbf{Rank} & \textbf{Metrics} & $K$=2 & 3 & 4 & 5 & 10 & Full & $T$=1s & 2s & 3s & 4s & 5s \\
\midrule
& &  & SCR ($\uparrow$) & \textbf{0.43} & \textbf{0.44} & \textbf{0.43} & \second{0.42} & \second{0.42} &  \second{0.48} & \textbf{0.41} & \textbf{0.48} & \second{0.46} & \second{0.50} & 0.44 \\
& \checkmark & \checkmark & ORR ($\downarrow$) & 1.10 & 0.98 & \textbf{0.91} & \textbf{0.91} & \textbf{1.39} & \textbf{1.43} & \textbf{2.45} & \second{4.54} & \textbf{6.85} & \second{9.46} & \textbf{10.38} \\
& & & FDE ($\downarrow$) & \textbf{4.00} & 5.41 & \second{5.87} & \textbf{5.79} & \textbf{7.65} & \textbf{8.22} & \textbf{6.33} & 5.96 & 7.17 & \second{7.01} & 5.73 \\
& & & CFD ($\downarrow$) & \textbf{1.00} & \textbf{1.14} & \textbf{1.22} & \textbf{1.22} & \textbf{1.78} & \textbf{1.73} & 2.47 & 3.86 & \second{4.11} & \second{4.77} & \textbf{5.41} \\
\midrule   
& & & SCR ($\uparrow$) & 0.38 & 0.45 & 0.40 & 0.40 & 0.39 & 0.40 & \textbf{0.41} & \second{0.44} & \second{0.46} & 0.48 & \second{0.48} \\
\checkmark & & \checkmark & ORR ($\downarrow$) & \second{0.81} & \second{0.76} & 1.00 & 1.06 & \second{1.47} & \second{1.60} & 2.78 & 4.83 & 7.39 & 9.40 & 10.44 \\
& & & FDE ($\downarrow$) & 4.33 & 5.28 & 6.13 & 6.82 & 8.65 & 9.20 & 7.03 & 6.14 & 8.02 & 6.99 & \second{5.55} \\
& & & CFD ($\downarrow$) & 1.81 & \second{1.60} & \second{1.84} & \second{1.94} & 2.92 & \second{2.62} & 2.87 & \second{3.64} & 4.27 & 5.37 & 6.46 \\
\midrule      
& & & SCR ($\uparrow$) & 0.33 & 0.34 & 0.36 & 0.37 & 0.39 & 0.36 & 0.34 & 0.35 & 0.41 & 0.41 & 0.39 \\
\checkmark & \checkmark & \text{Dist} & ORR ($\downarrow$) & 1.38 & 1.50 & 1.59 & \second{1.49} & 1.56 & 1.74 & 3.06 & 5.21 & 7.41 & 10.14 & \second{10.43} \\
& & & FDE ($\downarrow$) & \second{4.15} & \textbf{5.15} & 5.79 & 5.96 & 8.01 & 9.69 & \second{6.51} & \second{5.73} & \textbf{6.82} & \second{7.01} & \textbf{5.38} \\
& & & CFD ($\downarrow$) & \second{1.79} & 2.44 & 2.03 & 2.34 & 3.09 & 3.30 & \textbf{1.94} & \textbf{2.92} & \textbf{3.88} & \textbf{4.44} & \second{5.95} \\
\midrule
& & & SCR ($\uparrow$) & 0.34 & 0.35 & 0.33 & 0.31 & 0.33 & 0.33 & 0.33 & 0.34 & 0.40 & 0.37 & 0.40 \\
\checkmark & \checkmark & \text{Human} & ORR ($\downarrow$) & 1.66 & 1.65 & 1.73 & 1.93 & 1.66 & 1.75 & 3.10 & 5.25 & 7.44 & 10.37 & 10.51 \\
& & & FDE ($\downarrow$) & 5.80 & 6.74 & 7.40 & 7.84 & 8.63 & 8.99 & 8.12 & 7.25 & 8.70 & 9.16 & 7.01 \\
& & & CFD ($\downarrow$) & 2.21 & 2.51 & 2.83 & 3.14 & \second{2.60} & 2.96 & 3.39 & 5.20 & 6.17 & 6.65 & 8.43 \\
\midrule                              
& & & SCR ($\uparrow$) & \second{0.40} & \textbf{0.44} & \textbf{0.43} & \textbf{0.46} & \textbf{0.49} & \textbf{0.51} & 0.40 & \second{0.44} & \textbf{0.49} & \textbf{0.55} & \textbf{0.52} \\
\checkmark & \checkmark & \checkmark & ORR ($\downarrow$) & \textbf{0.61} & \textbf{0.72} & \second{0.99} & \second{1.02} & 1.80 & 2.05 & 2.92 & \textbf{4.52} & \second{7.10} & \textbf{9.35} & 10.51 \\
& & & FDE ($\downarrow$) & {4.17} & \second{5.22} & \textbf{5.99} & 6.59 & \second{7.84} & \second{8.26} & 7.06 & 5.54 & \second{6.86} & \textbf{7.00} & 5.71 \\
& & & CFD ($\downarrow$) & 1.88 & 1.92 & 1.93 & 2.25 & 2.83 & 3.47 & \second{2.37} & 4.08 & 4.25 & 4.97 & 6.33 \\
\bottomrule
\end{tabular}
}\vspace{-5mm}
\end{table}

We also extend our experiments to over-speed scenarios by incorporating an over-speed guidance function. We compare the Scene Overspeed Rate~(\textbf{SOR}) with CTG in Table~\ref{tab:gradient} (upper). CCDiff demonstrates better realism~(ORR, CFD) with comparable controllability~(SOR, SCR). This confirms that CCDiff is extensible to diverse safety-critical events under corresponding controllability guidance objectives.

We then analyze gradient conflicts in CTG and CCDiff, focusing on two aspects: (i) negative average cosine similarity among conflicted gradients and (ii) the percentage of agents with gradient conflicts (inner product $< 0$). Table~\ref{tab:gradient}~(lower) shows CCDiff reduces conflicting agents from $\scriptsize{\sim}$9.1\% (CTG) to $\scriptsize{\sim}$4.8\% and lowers negative average cosine similarity, demonstrating its effectiveness in mitigating gradient conflicts.

\begin{table}[htbp]
    \centering
    \scriptsize
    \caption{\textbf{Upper}: additional controllability experiments with over-speed guidance. \textbf{Lower:} Gradient conflict statistics. In both cases, CCDiff outperforms CTG in both metrics by a clear margin.}
    \resizebox{0.75\linewidth}{!} {
    \begin{tabular}{c c c c c c} \hline
    \textbf{Metric} & \textbf{CTG} & \textbf{Ours} & \textbf{Metric} & \textbf{CTG} & \textbf{Ours} \\ \hline
    SOR~($\uparrow$) & 0.68 & \textbf{0.73} & SCR~($\uparrow$) & \textbf{0.35} & 0.33 \\
    ORR~($\downarrow$) & 4.23 & \textbf{0.89} & CFD~($\downarrow$) & 15.81 & \textbf{9.65} \\ \hline
    \begin{tabular}[c]{@{}c@{}}Neg. grad. cosine\\ similarity (1e-2, $\downarrow$)\end{tabular} & 1.85 & \textbf{1.29}  & \begin{tabular}[c]{@{}c@{}}The \% of agents w/\\ grad conflict (\%, $\downarrow$)\end{tabular} & 9.12 & \textbf{4.79} \\  \hline
    \end{tabular}
    }
    \label{tab:gradient}
\end{table}

\begin{figure}[htbp]
    \centering
    \includegraphics[width=0.8\textwidth]{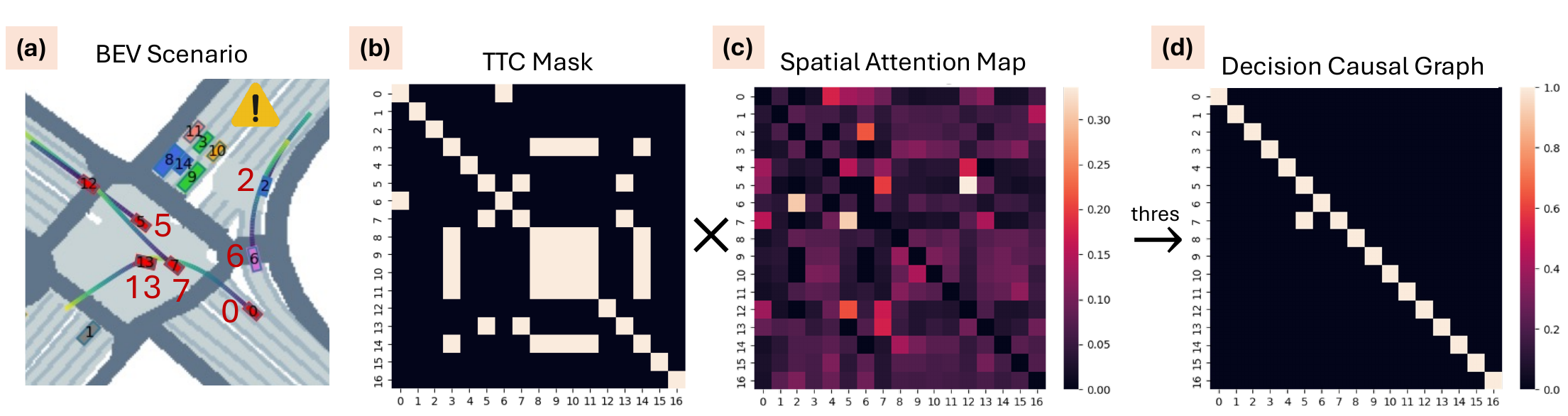}
    \caption{\textbf{(a)} Lane-changing at an intersection; \textbf{(b, c, d)} Interpretable computation of DCG from TTC mask and attention map. }
    \label{fig:interpretability}
\end{figure}

We further illustrate Decision Causal Graph~(DCG) computation using attention and time-to-collision~(TTC) masks in Figure~\ref{fig:interpretability}. As is shown in Figure~\ref{fig:interpretability}(a), Agent 7 tends to change lanes and interact with Agent 5. 
resulting in non-diagonal elements in the DCG matrix between Agents 5 and 7 in Figure~\ref{fig:interpretability}(d). 
This is computed by the TTC mask in Figure~\ref{fig:interpretability}(b) and attention map in Figure~\ref{fig:interpretability}(c). We've included more qualitative results in our qualitative examples in the following subsection. 

\begin{figure}[htbp]
    \centering
    \includegraphics[width=0.6\textwidth]{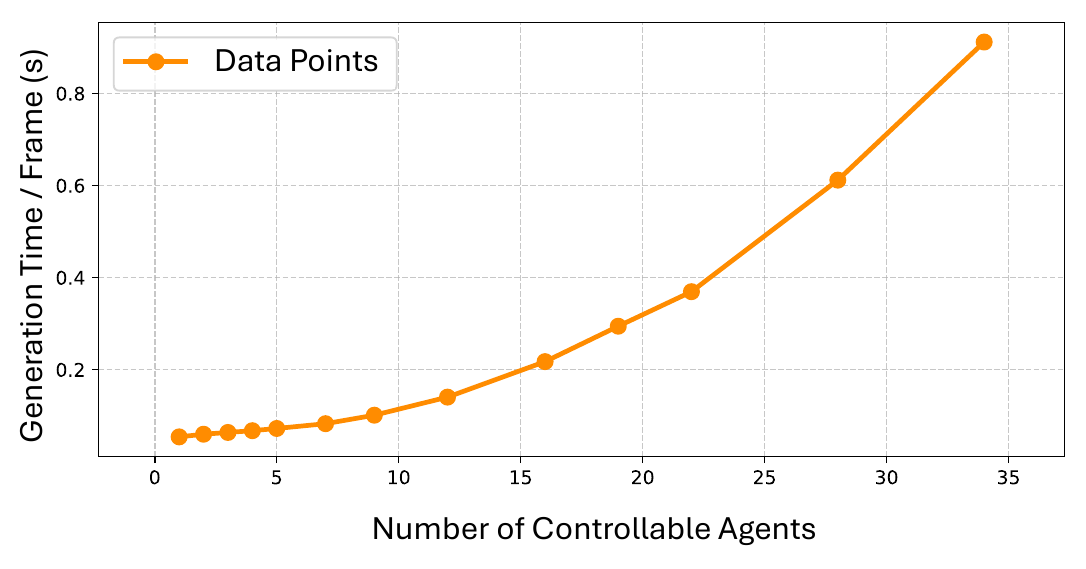}
    \caption{Inference speed with respect to the number of agents. }
    \label{fig:inference_speed}
\end{figure}

The CCDiff model has 15.4M parameters, including a CNN-based map encoder and a transformer-based trajectory encoder. Its inference speed is comparable to CTG at~$\scriptsize{\sim}$20 ms per frame per agent on an NVIDIA V100. Figure~\ref{fig:inference_speed} illustrates full-scene generation time across agent scales.

\newpage
\subsection{Additional Qualitative Results}
\subsubsection{Long-horizon Generation}

We evaluate the long-horizon generation with different planning cycle for the scenarios with same length between \method\ and all the baselines. We illustrate the qualitative examples below. The results demonstrate that \method\ can consistently generate realistic cross-traffic violation scenarios for $1s\leq T\leq 5s$. In contrast, CTG baseline can only generate an opposite-lane collision when $T=1s$. 

\begin{figure}[htbp]
    \centering
    \includegraphics[width=1.0\linewidth]{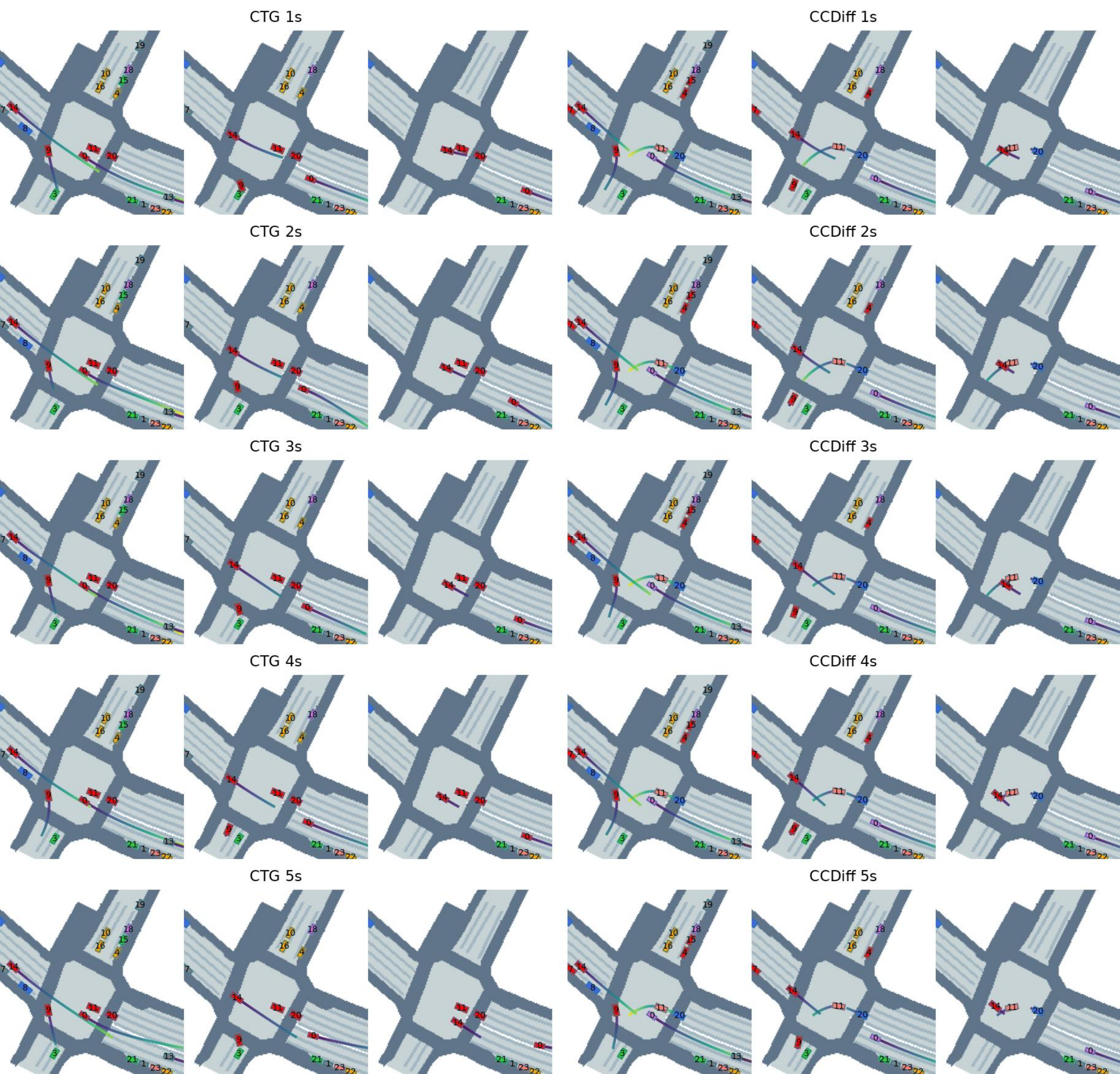}
    \caption{Comparison of \method\ and CTG on the controllability and realism under different sizes of controllable agents. We can see that \method\ can consistently generate realistic cross-traffic violation scenarios, yet CTG can only generate one with shorter planning cycle in 1s. }
    \label{fig:qualitative}
\end{figure}

\subsubsection{Multi-agent Generation}
We evaluate the multi-agent generation with different sizes of controllable agents $K$. We illustrate the qualitative examples of unprotected left turn scenarios below. The results demonstrate that with abundant controllable access to the agents at the scene ($K\geq 2$ in this case),  \method\ can consistently generate realistic unprotected left-turn scenarios compared to the CTG baseline. 

\begin{figure}[htbp]
    \centering
    \includegraphics[width=1.0\linewidth]{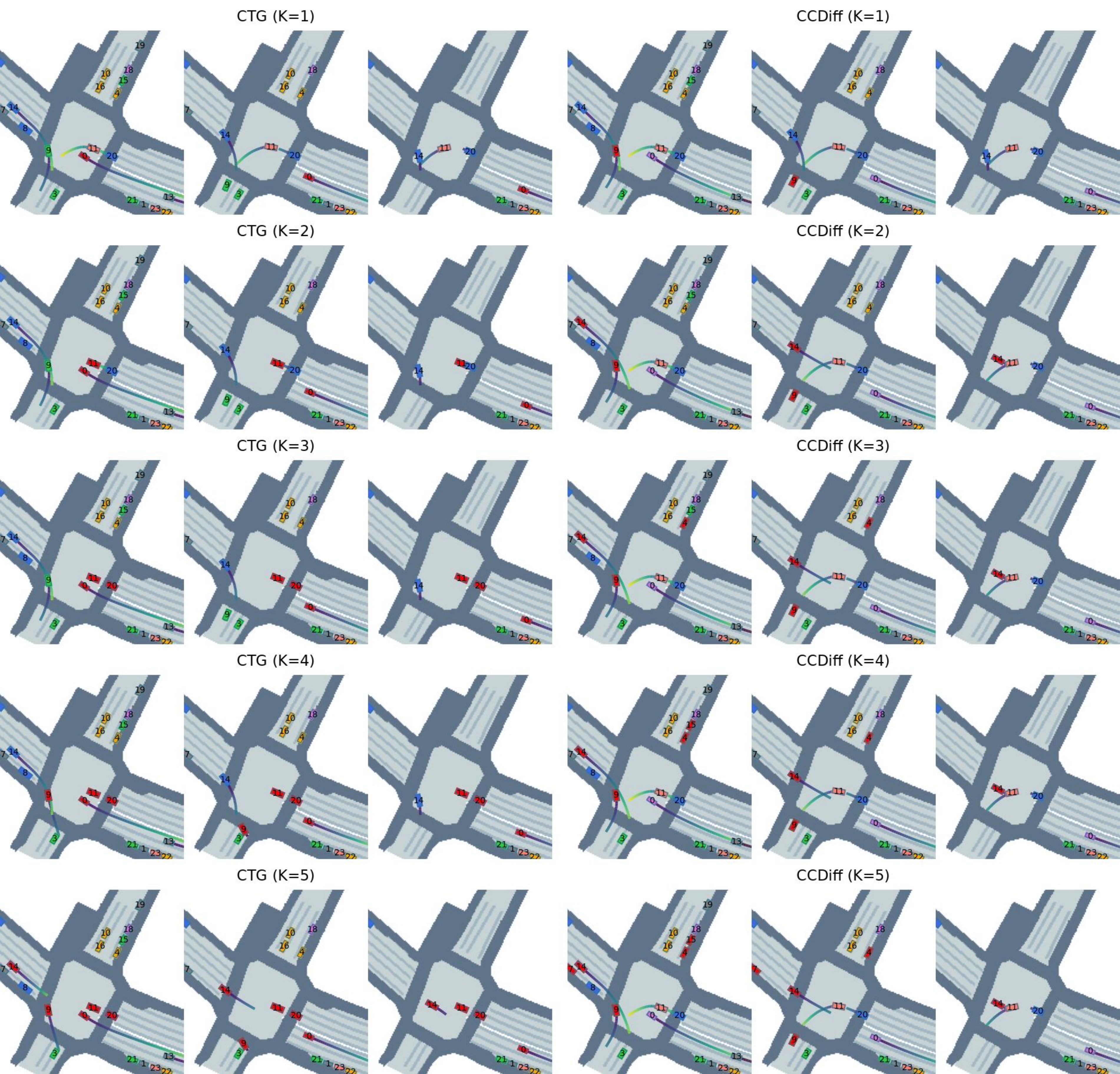}
    \caption{Comparison of \method\ and CTG on the controllability and realism under different sizes of controllable agents. We can see that when the number of controllable agents is greater than 1,  \method\ can consistently generate realistic unprotected left-turn violations, yet CTG can only generate one unrealistic right turn collision with 5 controllable agents. }
    \label{fig:qualitative}
\end{figure}

\clearpage
\newpage
\subsection{Detailed description of baselines}
\label{app:baseline}

\paragraph{SimNet}~\cite{bergamini2021simnet}: SimNet frames the problem as a Markov Process, and models state distributions and transitions directly from raw observational data, eliminating the need for handcrafted models. Trained on 1,000 hours of driving logs, it dynamically generates novel and adaptive scenarios that enable closed-loop evaluations. The system reveals subtle issues, such as causal confusion, in state-of-the-art planning models that traditional non-reactive simulations fail to detect. 

\paragraph{TrafficSim}~\cite{suo2021trafficsim}: TrafficSim uses an implicit latent variable model like conditional variational autoencoder~(CVAE). The system parameterizes a joint actor policy that simultaneously generates plans for the agents in a scene. The model is jointly trained with (i) ELBO objective inspired by CVAE and (ii) common-sense following with agents' pair-wise collision loss. TrafficSim generates diverse, realistic traffic scenarios and can serve as effective data augmentation for improving autonomous motion planners.

\paragraph{STRIVE}~\cite{rempe2022generating}: STRIVE employs a graph-based conditional variational autoencoder (VAE) to model realistic traffic motions and formulates scenario generation as an optimization problem in the latent space of this model. By perturbing real-world traffic data, STRIVE generates scenarios that stress-test planners. A subsequent optimization step ensures that the scenarios are useful for improving planner performance by being solvable and challenging. STRIVE has been successfully applied to attack two planners, showing its ability to produce diverse, accident-prone scenarios and improve planner robustness through hyperparameter tuning. 

\paragraph{BITS}~\cite{xu2023bits}: BITS (Bi-level Imitation for Traffic Simulation) framework leverages the hierarchical structure of driving behaviors by decoupling the simulation into two levels: high-level intent inference and low-level driving behavior imitation. This structure enhances sample efficiency, behavior diversity, and long-horizon stability. BITS also integrates a planning module to ensure consistency over extended scenarios.

\paragraph{CTG}~\cite{zhong2023guided}: CTG is a novel framework combining controllability and realism in traffic simulation by leveraging conditional diffusion models and Signal Temporal Logic (STL). The approach allows fine-grained control over trajectory properties, such as speed and goal-reaching, while maintaining realism and physical feasibility through enforced dynamics. Extending to multi-agent settings, the model incorporates interaction-based rules, such as collision avoidance, to simulate realistic agent interactions in traffic. 

We list implementation details of all the methods are listed below with important hyperparameters and model structures information in Table~\ref{app:parameters}.  

\begin{table}[!b]
\renewcommand\arraystretch{1.1}
\caption{Hyper-parameters of models used in experiments of CCDiff and baselines}
\label{app:parameters}
\centering
\begin{threeparttable}
  \begin{tabular}{c|c|c|c}
    \toprule
    Parameter Name & Value & Parameter Name & Value \\
    \midrule
    Step length & 0.1s & Map Encoder & ResNet-18 \\
    History steps & 31 & Map feature dim. & 256 \\
    Generation steps & 52 & Trajectory Encoder & MLP \\
    Learning rate & 0.0001 & Trajectory feature dim. & 128 \\
    Optimizer & \text{Adam} & Transformer decoder head & 16 \\
    Batch size & 100 & Transformer decoder layers & 2 \\
    Trajectory prediction loss weight & 1.0 & Guidance gradient Steps & 30 \\
    Yaw regularization weight & 0.1 & Guidance constraint norm & 100 \\
    EMA step & 1 & Guidance learning rate & 0.001 \\
    EMA decay & 0.995 & Guidance optimizer & Adam \\
    Denoising Steps & 100 & Guidance weight: off-road & 1.0 \\
    Guidance discount factor & 0.99 & Guidance weight: collision & -50.0 \\
    Planning steps & 10, 20, 30, 40, 50 & TTC threshold & 3.0 $\mathrm{s}$ \\
    Controllable Agents & 1, 2, 3, 4, 5, 10, Full & Distance threshold & 50 $\mathrm{m}$ \\
    \bottomrule
  \end{tabular}
 \end{threeparttable}
\end{table}

\paragraph{Training and Inference Resources } We conduct training and inference of all the models on 4x NVIDIA Tesla V100 with 16GB GPU memory each, and 48-core CPU Intel(R) Xeon(R) CPU @ 2.30GHz. 
The training of one model takes 3 hours per epoch on nuScenes training split, and we train 10 epochs for each baseline model and CCDiff. 
At inference time, the parallel evaluation takes an average of 3 minutes on each closed-loop testing scenario for all the methods under the same configuration (controllable agents and generation frequencies).

\subsection{Detailed description of evaluation metrics}

\begin{figure}[htbp]
    \centering
    \includegraphics[width=0.65\linewidth]{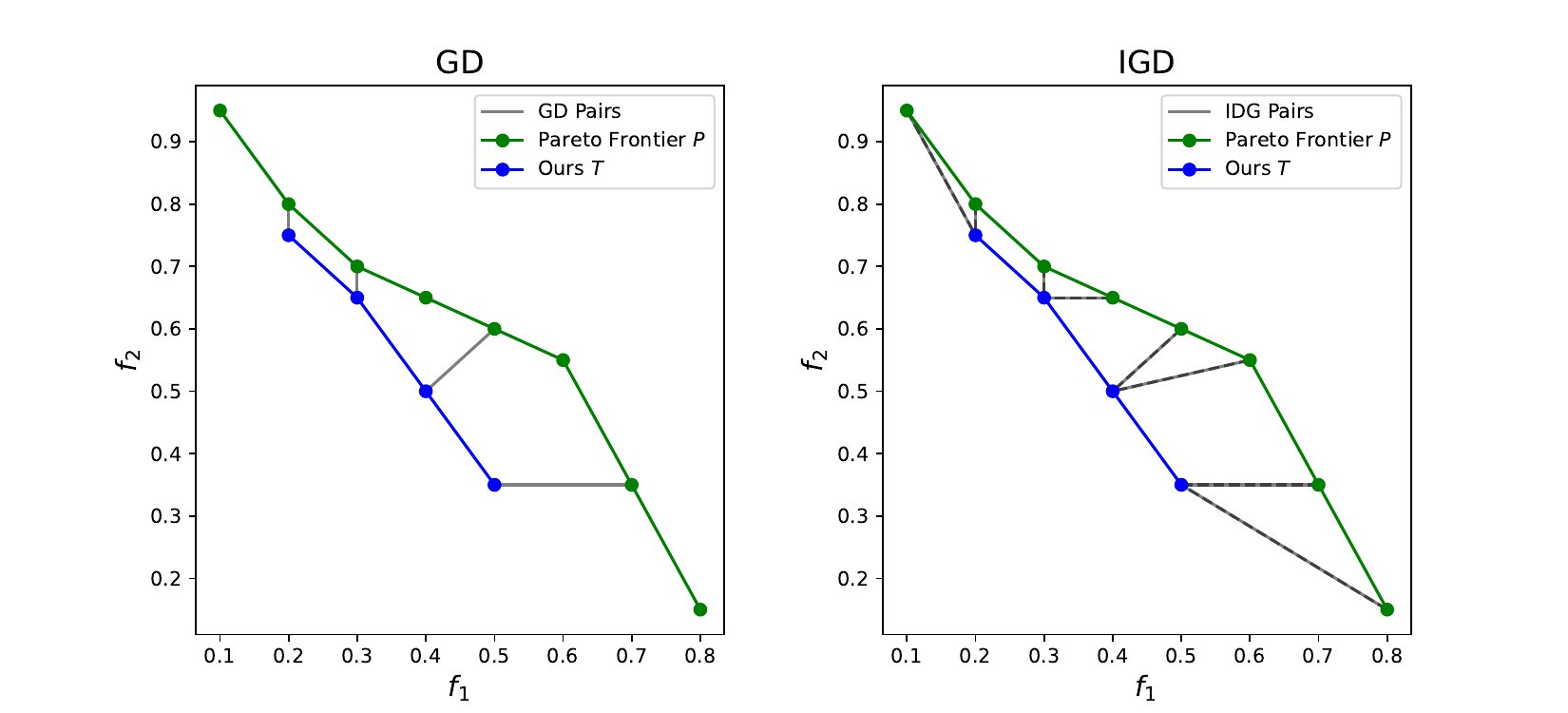}
    \vspace{-3mm}
    \caption{Examples of GD and IGD used to evaluate the multi-objective optimization. Two axes $f_1, f_2$ represent two objectives. }
    \vspace{-5mm}
    \label{fig:moo_objective}
\end{figure}

\begin{itemize}
    \item \textbf{Controllability Score~(CS)}: The computation of CS standardizes the scenario-wise collision rate~(SCR) used in~\cite{suo2021trafficsim, tan2023language}: 
    \begin{equation*}
        \text{CS} = \frac{\text{SCR} - \min(\text{SCR})}{\max(\text{SCR})-\min(\text{SCR})}
    \end{equation*}
    We then standardize SCR among all the methods to get the \textbf{CS}, a higher-the-better score between 0 and 1. 
    \item \textbf{Realism Score~(RS)}: We average over three widely-used quantitative metrics to evaluate the realism of the scenarios: (i) scenario off-road rate~(ORR) used in~\cite{tan2023language, ding2023realgen}, (ii) final displacement error~(FDE, $\mathrm{m}$) and (iii) comfort distance~(CFD) in~\cite{xu2023bits,zhong2023guided} to quantify the realism of the similarity in the smoothness of agents' trajectories in the generated scenarios. We standardize all the metrics among all the methods respectively and average them to get the \textbf{RS}, a higher-the-better score between 0 and 1: 
    \begin{equation*}
        \text{RS} = 1.0 - \frac{1}{3} \Big( \frac{\text{ORR} - \min(\text{ORR})}{\max(\text{ORR})-\min(\text{ORR})} + \frac{\text{FDE} - \min(\text{FDE})}{\max(\text{FDE})-\min(\text{FDE})} + \frac{\text{CFD} - \min(\text{CFD})}{\max(\text{CFD})-\min(\text{CFD})} \Big)
    \end{equation*}
    Specifically, FDE describes the trajectory closeness between the synthetic one and the original one, ORR describes how frequently the generated trajectories go off-road, while CFD measures the \textbf{smoothness} of the generated trajectories with their acceleration and jerk. All these raw metrics are lower the better, so after we revert it above, the resulting RS is a higher-the-better metric. 
    \item \textbf{Multi-objective optimization metrics}: with the \textbf{RS} and \textbf{CS},  we further quantify the optimality of the solution based on generational distance~(\textbf{GD}) and inverted generational distance~(\textbf{IGD}), the average minimum distance between the methods and Pareto frontier~\cite{coello2007evolutionary, yu2020gradient}: 
    \begin{equation*}    
    \text{GD} = \left( \frac{1}{|\mathcal{D}|} \sum_{\mathbf{d} \in \mathcal{D}} \min_{\mathbf{p} \in \mathcal{P}} \|\mathbf{a} - \mathbf{p}\|^q \right)^{\frac{1}{q}},
    \end{equation*}
    
    where $\|\cdot\|$ denotes the Euclidean distance, and $q$ is typically set to 2. Conversely, IGD measures the average distance from each solution in the Pareto frontier $\mathcal{P}$ to its nearest solution in the obtained set $\mathcal{D}$, and is defined as
    \begin{equation*}
    \text{IGD} = \left( \frac{1}{|\mathcal{P}|} \sum_{\mathbf{p} \in \mathcal{P}} \min_{\mathbf{d} \in \mathcal{D}} \|\mathbf{p} - \mathbf{d}\|^q \right)^{\frac{1}{q}}.
    \end{equation*}
    Both metrics provide insights into the convergence and diversity of the obtained solution set: lower values of GD indicate better convergence to the Pareto frontier. On the other hand, lower values of IGD suggest better coverage over the Pareto frontier. We visualize an example for GD and IGD in Figure~\ref{fig:moo_objective}. 
\end{itemize}
\vspace{-5mm}
\vspace{-5mm}

\newpage
\paragraph{Quantitative Analysis on the design Causal Masking Design} We also analyze the importance of different features w.r.t. the collision samples in the generated scenarios. The results show that TTC feature has the highest statistical correlation with the controllability score (i.e. the collision rate) in our setting. 

\begin{figure}[htbp!]
    \centering
    \includegraphics[width=0.9\linewidth]{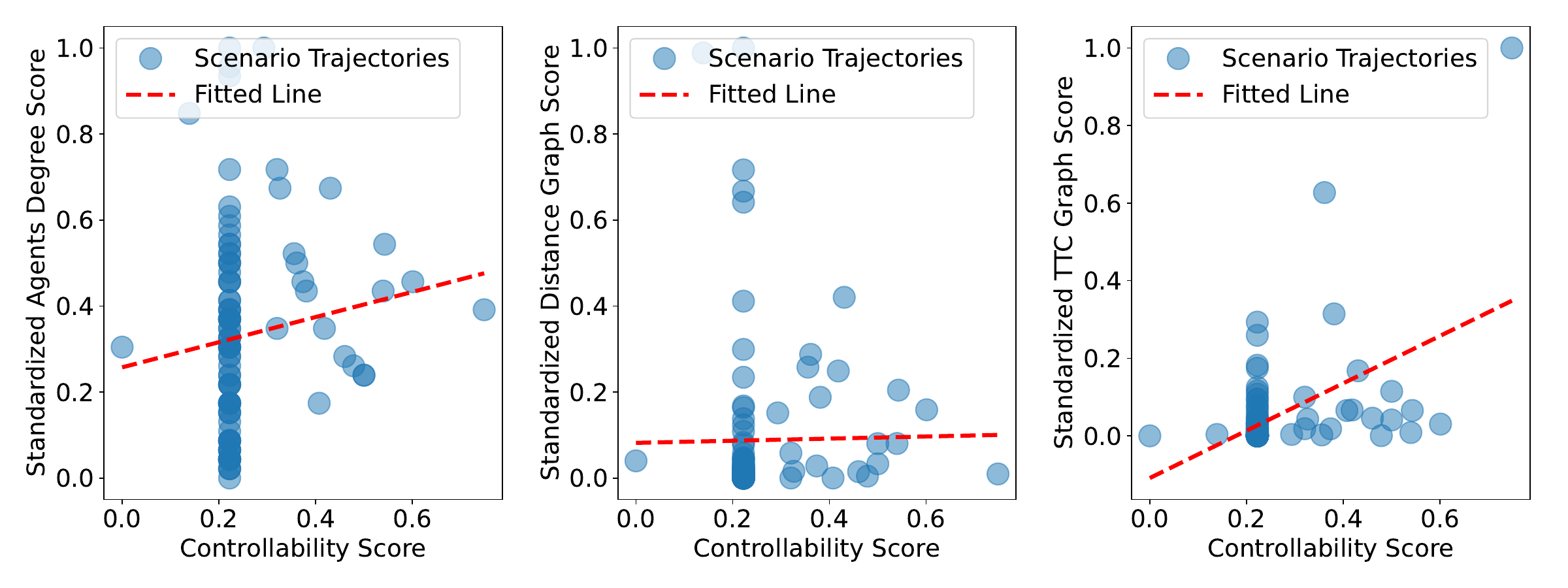}
    \caption{The number of cliques in the TTC graph is more informative causal features of safety-critical incidents (higher Pearson correlation) compared to Relative Distance and number of agents. }
    \label{fig:app_feat}
\end{figure}

\begin{table*}[htbp]
    \centering
    \caption{Correlation analysis between the collision accidents and different causal structure features: standardized clique score for TTC graph, standardized clique score for distance graph, and the standardized number of agents at the scene. We list the Pearson correlation $R^2$ between the standardized controllability score for each scenario, as well as the significance level of each feature (p-value)}
    \begin{tabular}{@{}lcc@{}}
        \toprule
        \textbf{Causal Structure Feature} & $R^{2}(\uparrow)$ & p-value ($\downarrow$) \\ 
        \midrule
        \#Cliques in Dist. graph & 0.01 & 0.89 \\
        \#Agents & 0.13 & 0.20 \\
        \textbf{\#Cliques in TTC graph (Ours)} & \textbf{0.49} & $\mathbf{2.2\times 10^{-7}}$ \\
        \bottomrule
    \end{tabular}
    \label{tab:correlation_analysis}
\end{table*}

\clearpage
\newpage
\subsection{Additional Qualitative Analysis over Scenarios}
In the following subsection, we present seven representative interactive scenarios that are safety-critical in urban traffic. We begin by analyzing the comparisons with baseline methods and highlighting the differences between distance-based graphs and TTC-based graphs. The results demonstrate that TTC-based graphs are generally sparser yet more informative, particularly for capturing safety-critical maneuvers.

Additionally, we provide examples of multi-agent, long-horizon trajectory generation for individual scenarios, showcasing the model's ability to handle complex interactions over extended time frames.

\subsubsection{Unprotected Left Turn}

\paragraph{Baseline Comparison} Below, we present the unprotected left-turn scenarios. The relational reasoning of the distance-based graph fails to capture the interaction between the two involved vehicles (11 and 14). We omit the multi-agent and long-horizon generation examples for this scenario, as these have already been analyzed in previous comparisons.

Among all the baselines, CTG, SimNet, and BITS closely follow the ground-truth trajectories, successfully generating a left-lane right turn without producing collision samples. In contrast, STRIVE generates unrealistic collisions with parked vehicles in the side lane. Notably, only CCDiff manages to produce realistic unprotected left-turn behaviors.
Only the TTC mask captures the interaction between agents 11 and 14. 

\begin{figure}[htbp]
    \centering
    \includegraphics[width=1.0\linewidth]{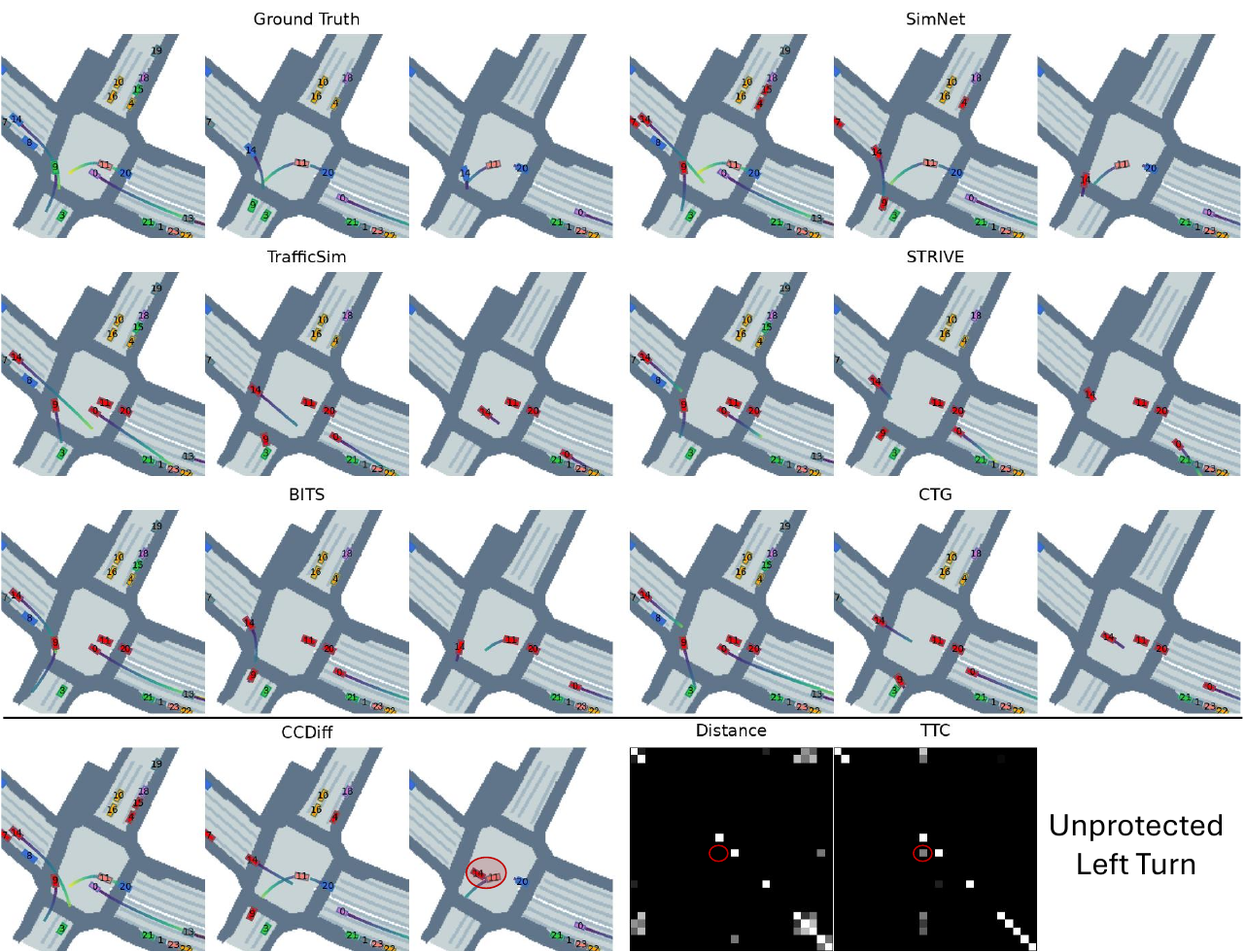}
    \caption{Qualitative of \method\ and baselines in unprotected left turn scenarios. }
    \label{fig:first-qualitative}
\end{figure}

\newpage
\subsubsection{Cross Traffic Violation} 

\paragraph{Baseline Comparison} A cross-traffic violation occurs when a vehicle at a T-intersection fails to yield the right of way to a vehicle approaching from a perpendicular direction. Such violations often result in side-impact collisions, particularly when the violating driver misjudges the speed or distance of the cross-traffic vehicle. In \method, agent 0 collides with agent 6, illustrating this scenario.

Among the baselines, BITS, TrafficSim, and CTG successfully avoid generating collision samples. However, SimNet also generates a collision between agent 0 and agent 6, failing to model the scenario accurately.
Both TTC and distance mask manage to capture the interaction between agents 0 and 6.

\begin{figure}[htbp]
    \centering
    \includegraphics[width=1.0\linewidth]{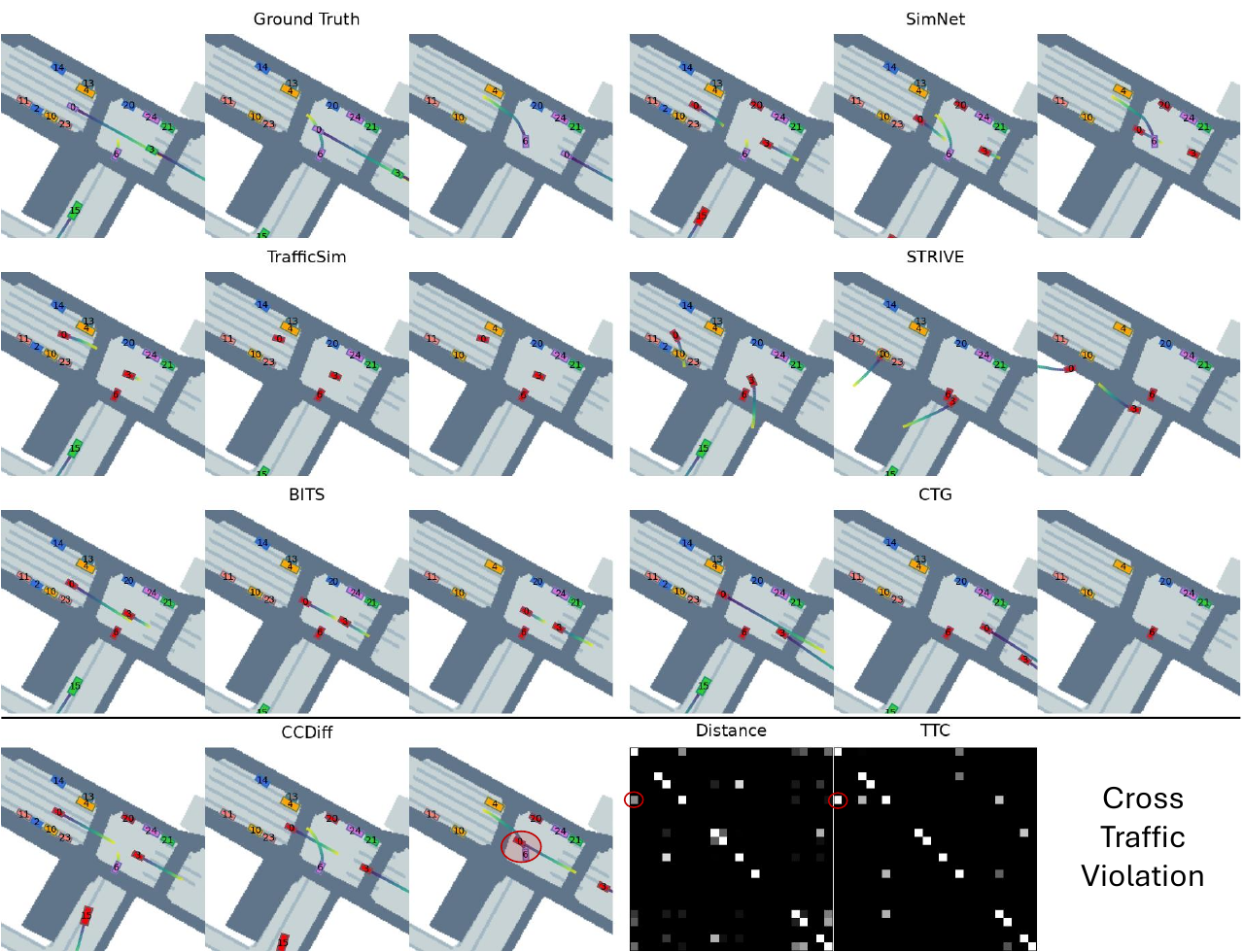}
    \caption{Qualitative of \method\ and baselines in cross traffic violation scenarios. }
    \label{fig:qualitative}
\end{figure}

\newpage
\paragraph{Multi-agent Generation} We compare the multi-agent generation results of \method\ with CTG. \method\ can consistently generate the cross traffic violation when the controllable agents $K\geq 2$. 
\begin{figure}[htbp]
    \centering
    \includegraphics[width=1.0\linewidth]{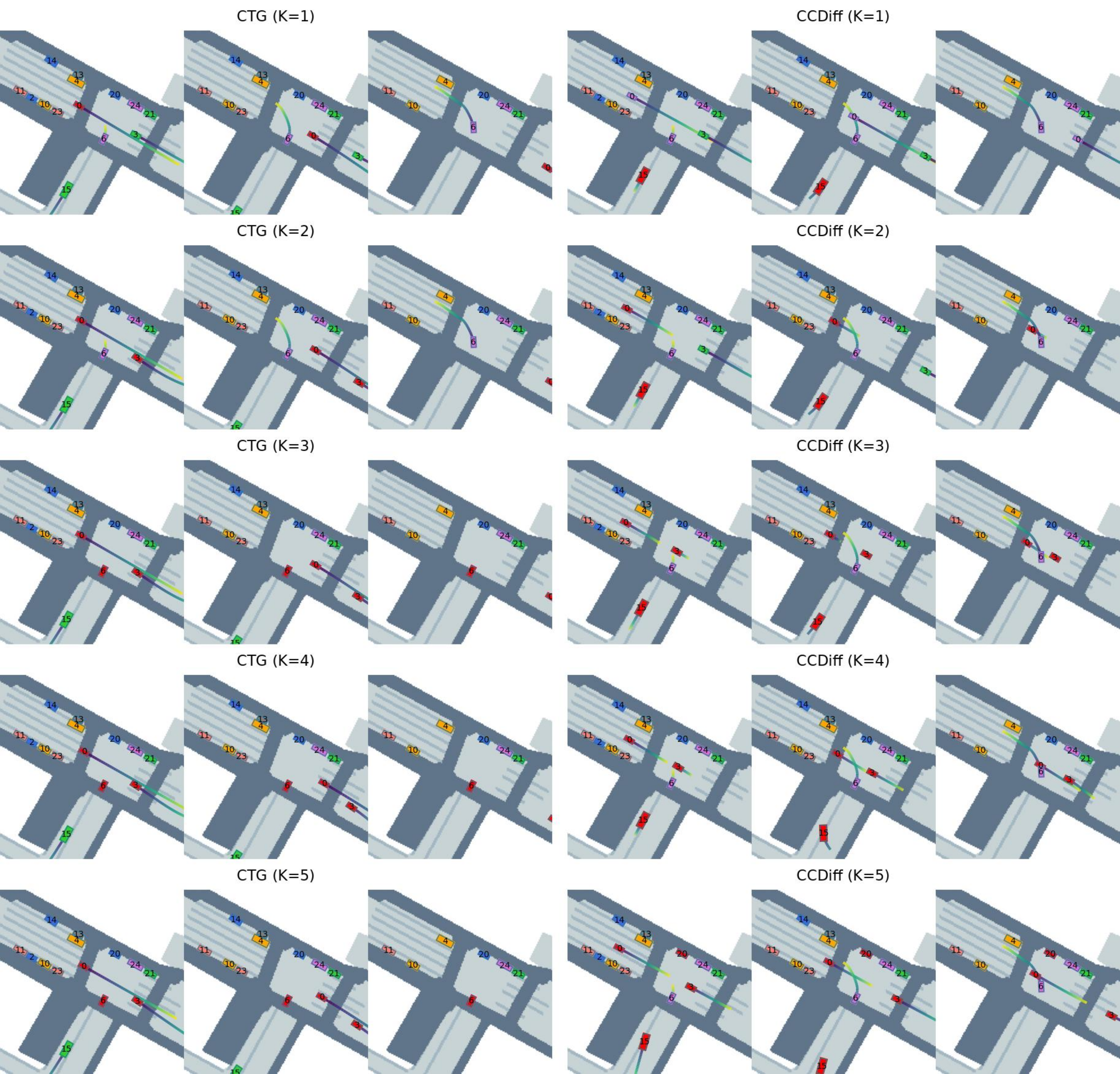}
    \caption{Qualitative comparison of \method\ and CTG under cross traffic violation generation under different sizes of controllable agents. }
    \label{fig:qualitative}
\end{figure}

\newpage
\paragraph{Long-horizon Generation}
We compare the long-horizon generation results of \method\ with CTG. \method\ can consistently generate the cross traffic violation even with a generation horizon $T> 2s$, yet CTG generated scenarios are more conservative. 

\begin{figure}[htbp]
    \centering
    \includegraphics[width=1.0\linewidth]{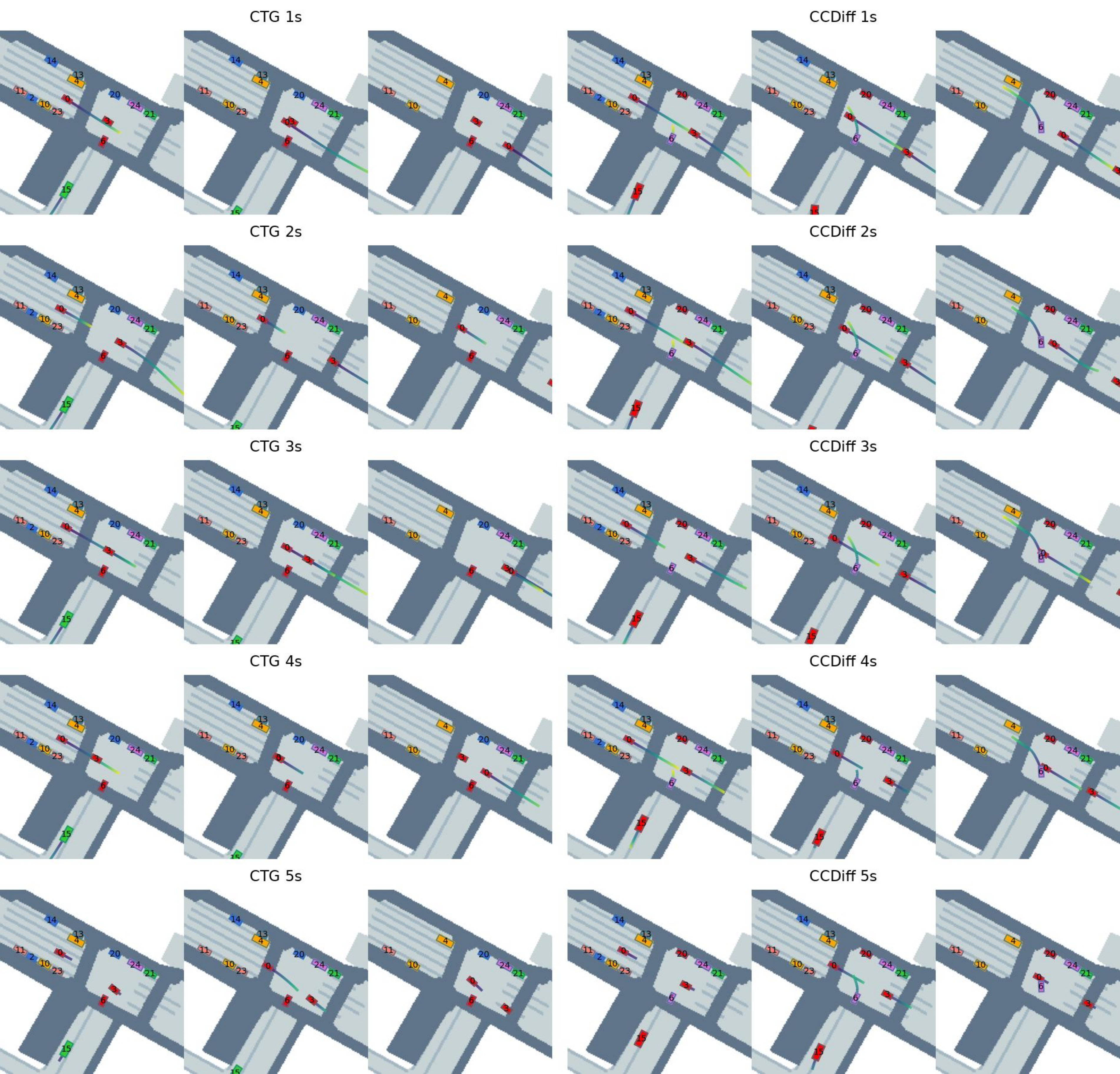}
    \caption{Qualitative comparison of \method\ and CTG under cross traffic violation generation under different generation horizons. }
    \label{fig:qualitative}
\end{figure}

\newpage
\subsubsection{Lane Cut-in}
\paragraph{Baseline Comparison}
A lane cut-in at an intersection occurs when a vehicle abruptly changes lanes or merges into another lane while navigating through or approaching an intersection, often without sufficient clearance or signaling. This maneuver typically forces other vehicles in the affected lane to brake suddenly or adjust their trajectory, increasing the risk of collisions or near-misses. In our case, agent 3 will suddenly cut in from the left lane to the right lane and collide with agent 0. 

Among all the baselines, CTG and SimNet generate some irregular behaviors and drive some of the controllable agents off-road. STRIVE generates relatively unrealistic right turn collision, and TrafficSim generates a wild unprotected left turn that is more unrealistic under this context. 
The TTC mask manages to capture the interaction between agents 0 and 3, while the distance mask misses it.

\begin{figure}[htbp]
    \centering
    \includegraphics[width=1.0\linewidth]{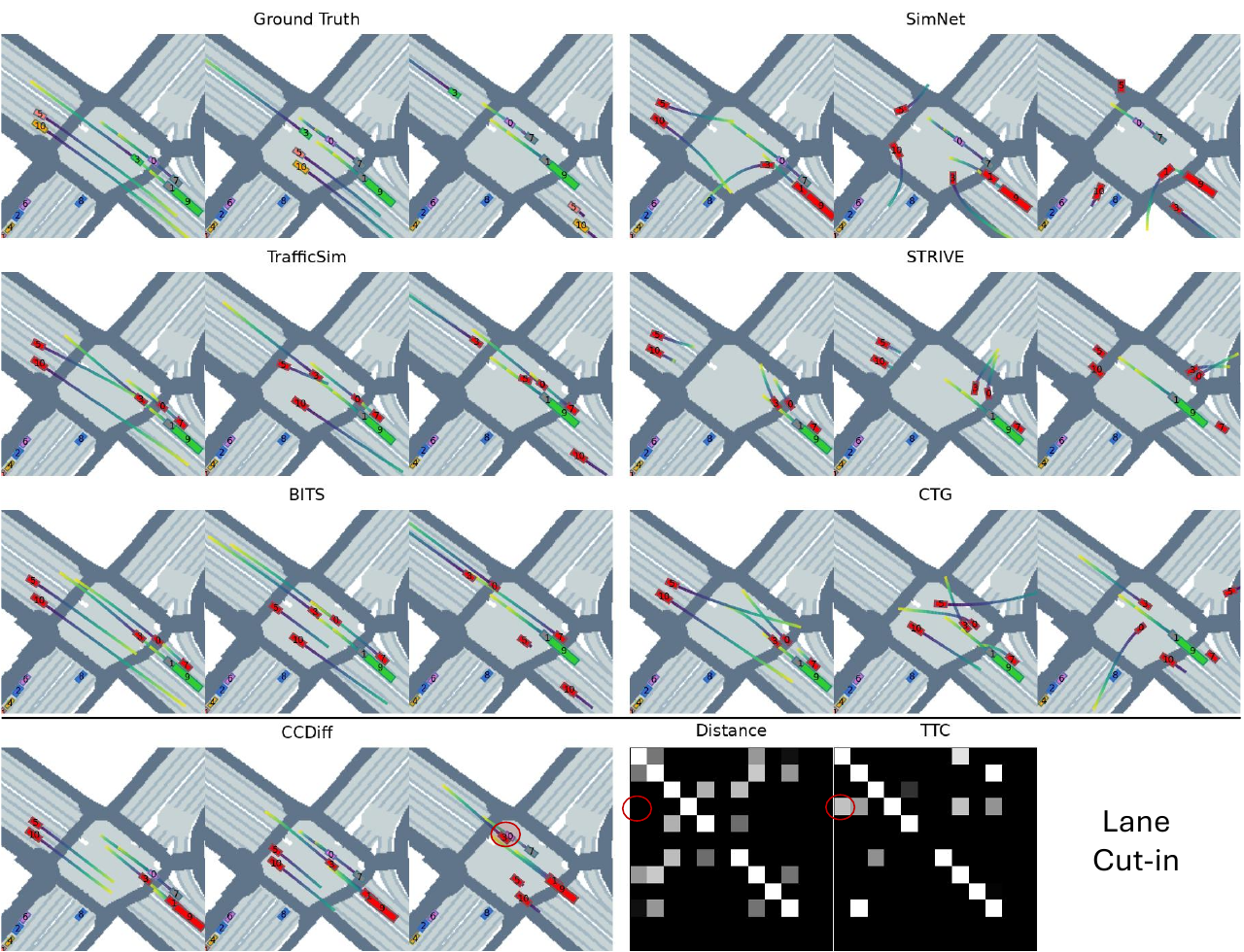}
    \caption{Qualitative of \method\ and baselines in lane cut-in scenarios.}
    \label{fig:qualitative}
\end{figure}

\newpage
\paragraph{Multi-agent Generation} We compare the multi-agent generation results of \method\ with CTG. \method\ can generate collision samples when $K=5$, yet the CTG generates very wild behaviors that are unrealistic from the ground-truth trajectories. 
\begin{figure}[htbp]
    \centering
    \includegraphics[width=1.0\linewidth]{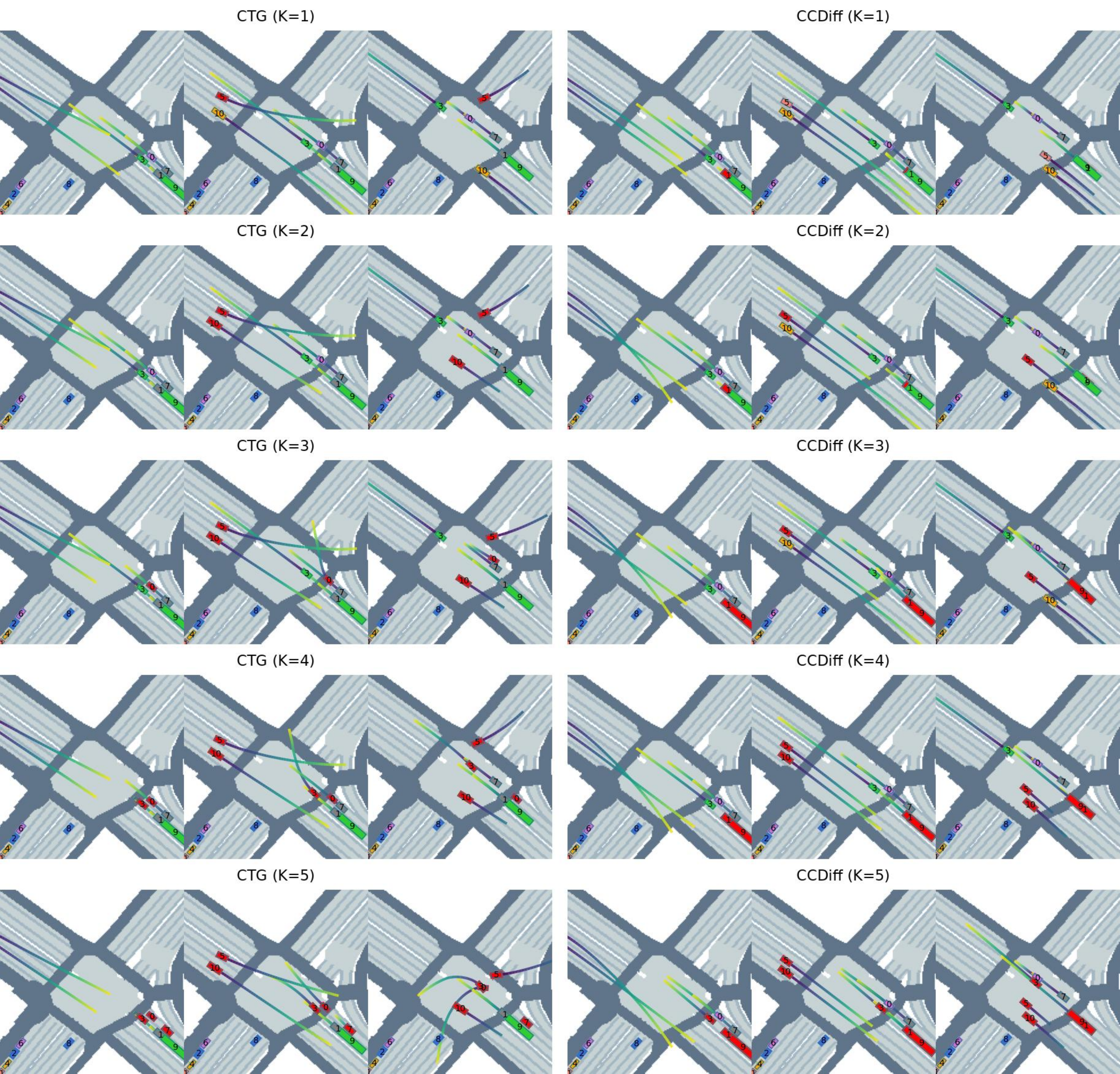}
    \caption{Qualitative comparison of \method\ and CTG under cross traffic violation generation under different sizes of controllable agents. }
    \label{fig:qualitative}
\end{figure}

\newpage
\paragraph{Long-horizon Generation}
We compare the long-horizon generation results of \method\ with CTG. \method\ can consistently generate the cut-in violation scenarios with the generation horizon $1s\leq T\leq 4s$. In contrast, CTG attempts to generate some unprotected left turn in this context but fails. 

\begin{figure}[htbp]
    \centering
    \includegraphics[width=1.0\linewidth]{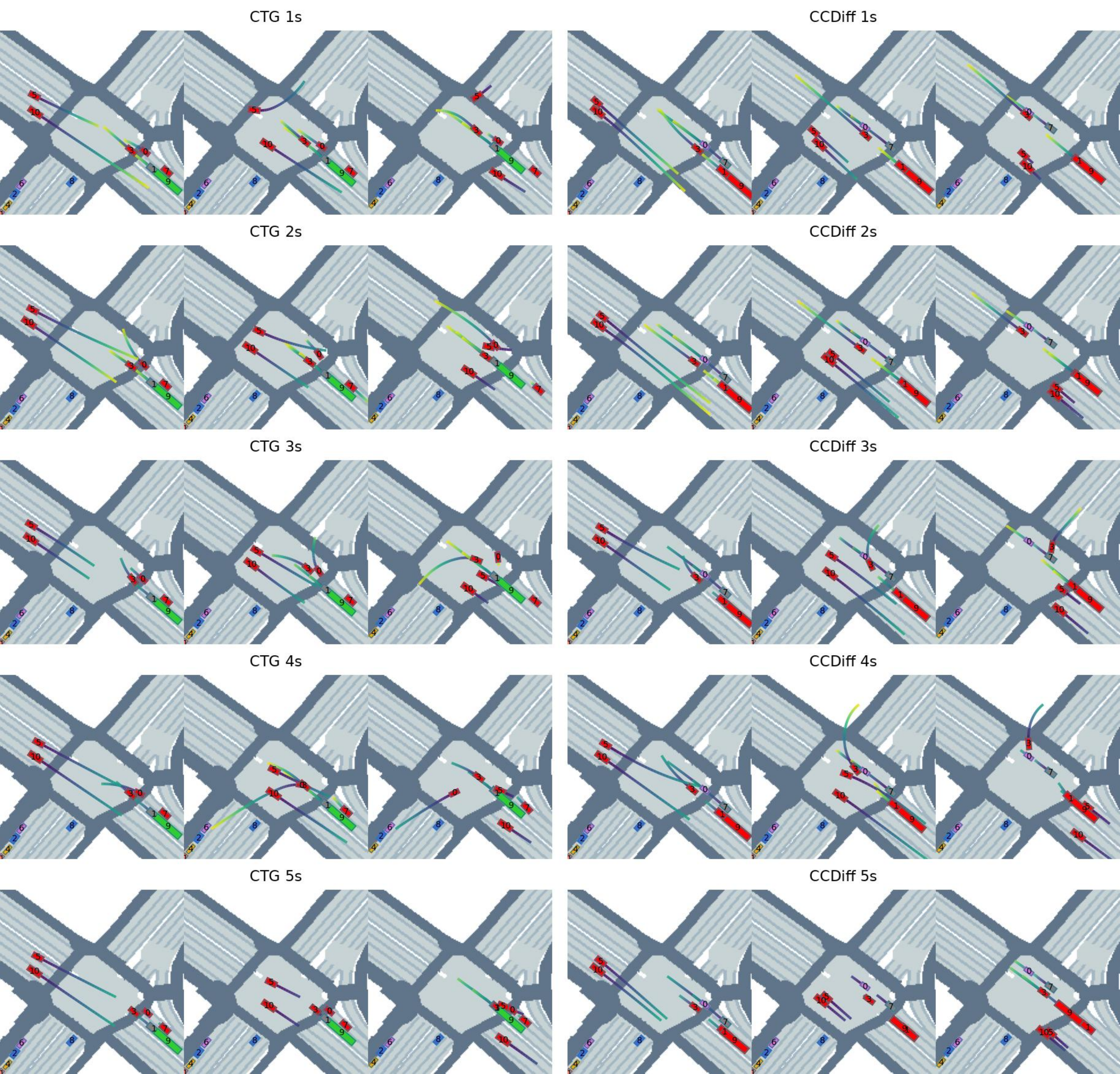}
    \caption{Qualitative comparison of \method\ and CTG under cross traffic violation generation under different generation horizons. }
    \label{fig:qualitative}
\end{figure}

\newpage
\subsubsection{Emergency Break}
\paragraph{Baseline Comparison}
The emergency break occurs when the middle vehicle (agent 0) brakes to keep distance from the forward vehicle (agent 9) suddenly, causing the trailing vehicle (agent 8) to collide with it due to insufficient stopping distance. 

Among all the baselines, STRIVE generates some irregular behaviors, which drive some of the controllable agents off-road. TrafficSim, BITS, and SimNet fail to generate safety-critical samples.  
Notably, although CTG also generates some collision samples, it accelerates the trailing vehicle 8 to collide with the middle vehicle 0, which does not break yet. 
In comparison, in our case, the middle vehicle 0 breaks and causes a collision with trailing vehicle 8 at normal speed, which is more realistic. 
Both the TTC mask and distance mask capture the interaction among agents 0, 8, and 9 in this scenario. 

\begin{figure}[htbp]
    \centering
    \includegraphics[width=1.0\linewidth]{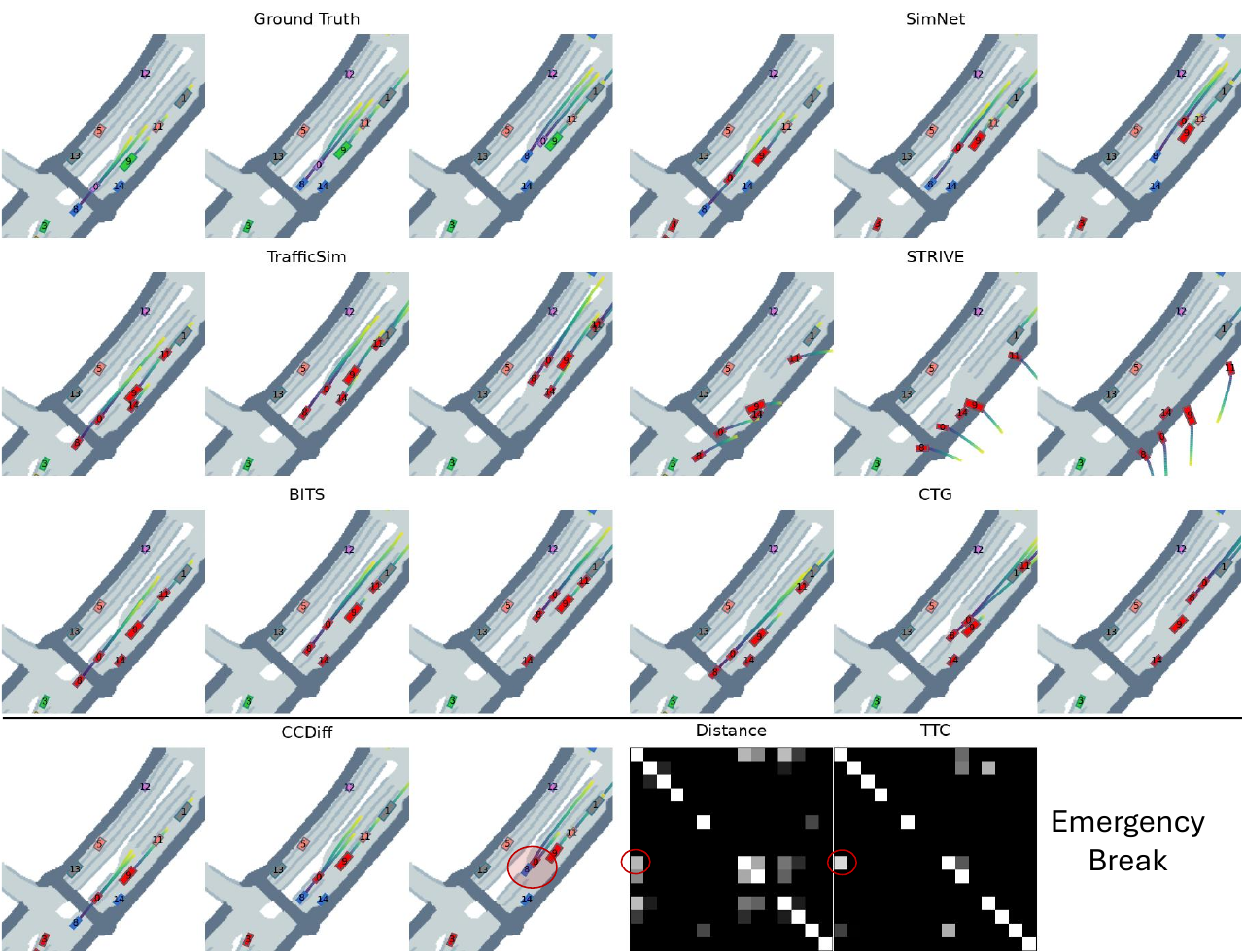}
    \caption{Qualitative of \method\ and baselines in the emergency break scenarios. }
\end{figure}

\newpage
\paragraph{Multi-agent Generation} We compare the multi-agent generation results of \method\ with CTG. \method\ can consistently generate safety-critical emergency breaking samples when $K\geq 2$, with a control of the most important vehicle 8 in this context. In contrast, CTG keeps accelerating the rear vehicle 8 instead of slowing down the middle vehicle 0. 
\begin{figure}[htbp]
    \centering
    \includegraphics[width=1.0\linewidth]{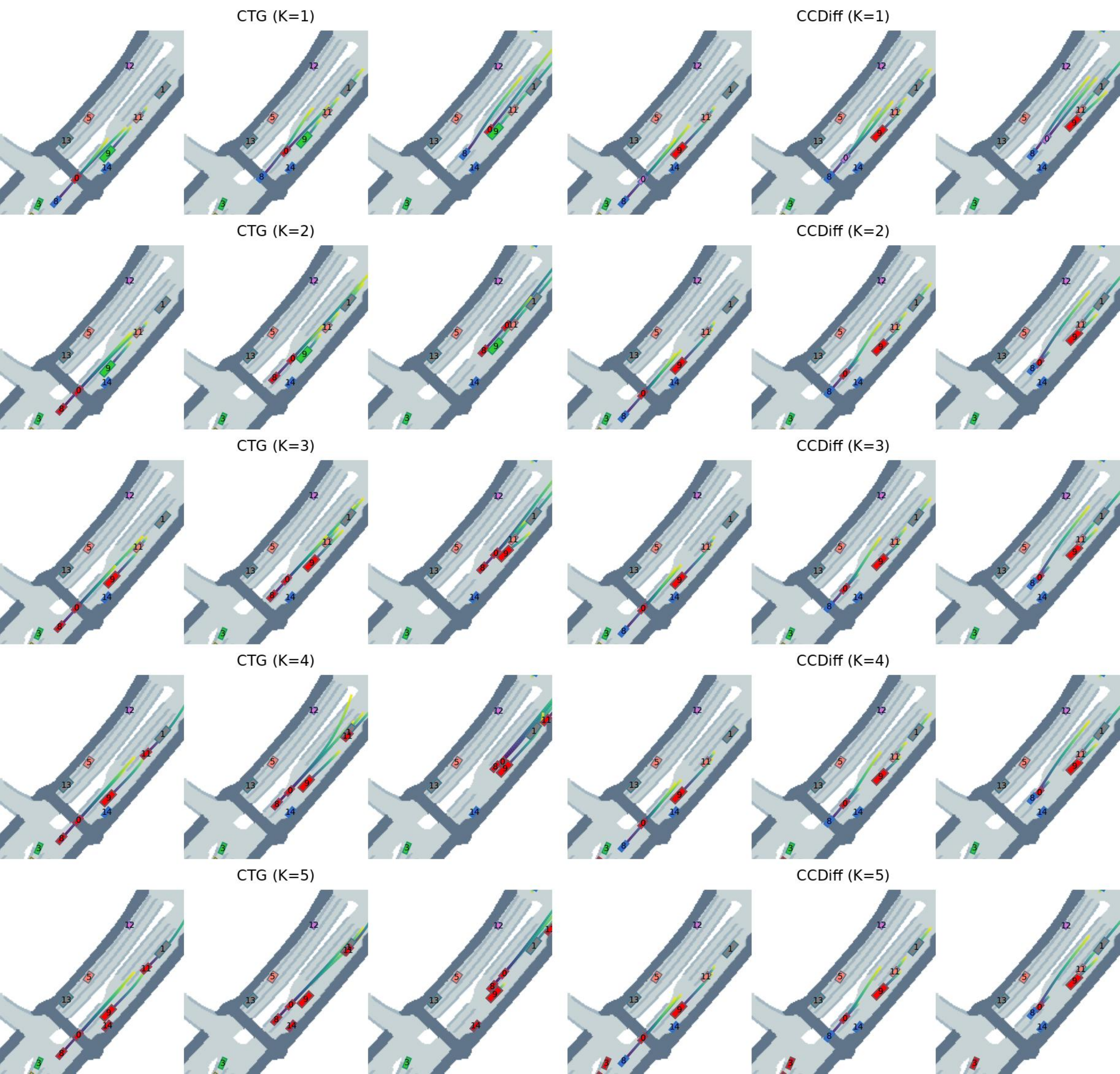}
    \caption{Qualitative comparison of \method\ and CTG under cross traffic violation generation under different sizes of controllable agents. }
    \label{fig:qualitative}
\end{figure}

\newpage
\paragraph{Long-horizon Generation}
We compare the long-horizon generation results of \method\ with CTG. \method\ can consistently generate the cut-in violation scenarios with all different lengths of the generation horizon $1s\leq T\leq 5s$. In contrast, CTG attempts to accelerate the vehicle in the middle and cannot generate any near-miss samples with longer generation horizons. 

\begin{figure}[htbp]
    \centering
    \includegraphics[width=1.0\linewidth]{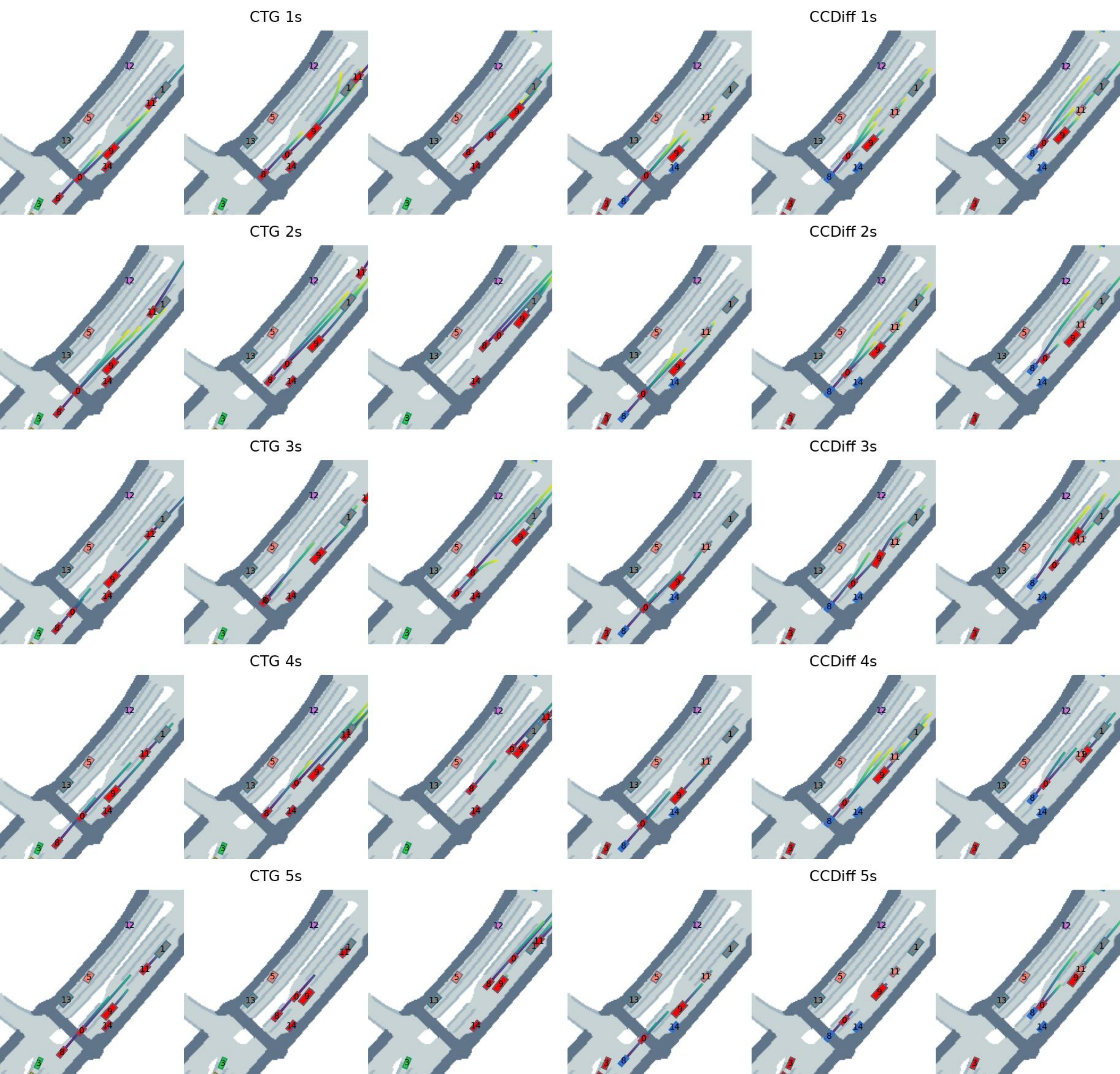}
    \caption{Qualitative comparison of \method\ and CTG under cross traffic violation generation under different generation horizons. }
    \label{fig:qualitative}
\end{figure}

\newpage
\subsubsection{Chain-reaction Crash}
\paragraph{Baseline Comparison}
A chain-reaction crash involving five vehicles (agents 1, 2, 5, 7, 8) occurs when a sudden stop or collision causes a cascade of impacts among closely spaced vehicles in the same lane. This happens before an intersection when vehicles fail to maintain a safe following distance, leading to multiple rear-end collisions.

Among all the baselines, SimNet and BITS fail to generate safety-critical scenarios. TrafficSim, STRIVE, and CTG generate collisions between agent 0 on the side lane with agent 2 with a very unrealistic cut-in behavior. 
In comparison, CCDiff generates realistic collisions where the trailing vehicles 1, 7, and 8 fail to break timely and collide with static front vehicle 5, waiting for the right turn of 2. 
Both TTC graph and distance graph captures the interaction of 5 and 7, 8. 
Yet distance-based graphs fail to capture the indirect interaction between 2 and 7, 8. 

\begin{figure}[htbp]
    \centering
    \includegraphics[width=1.0\linewidth]{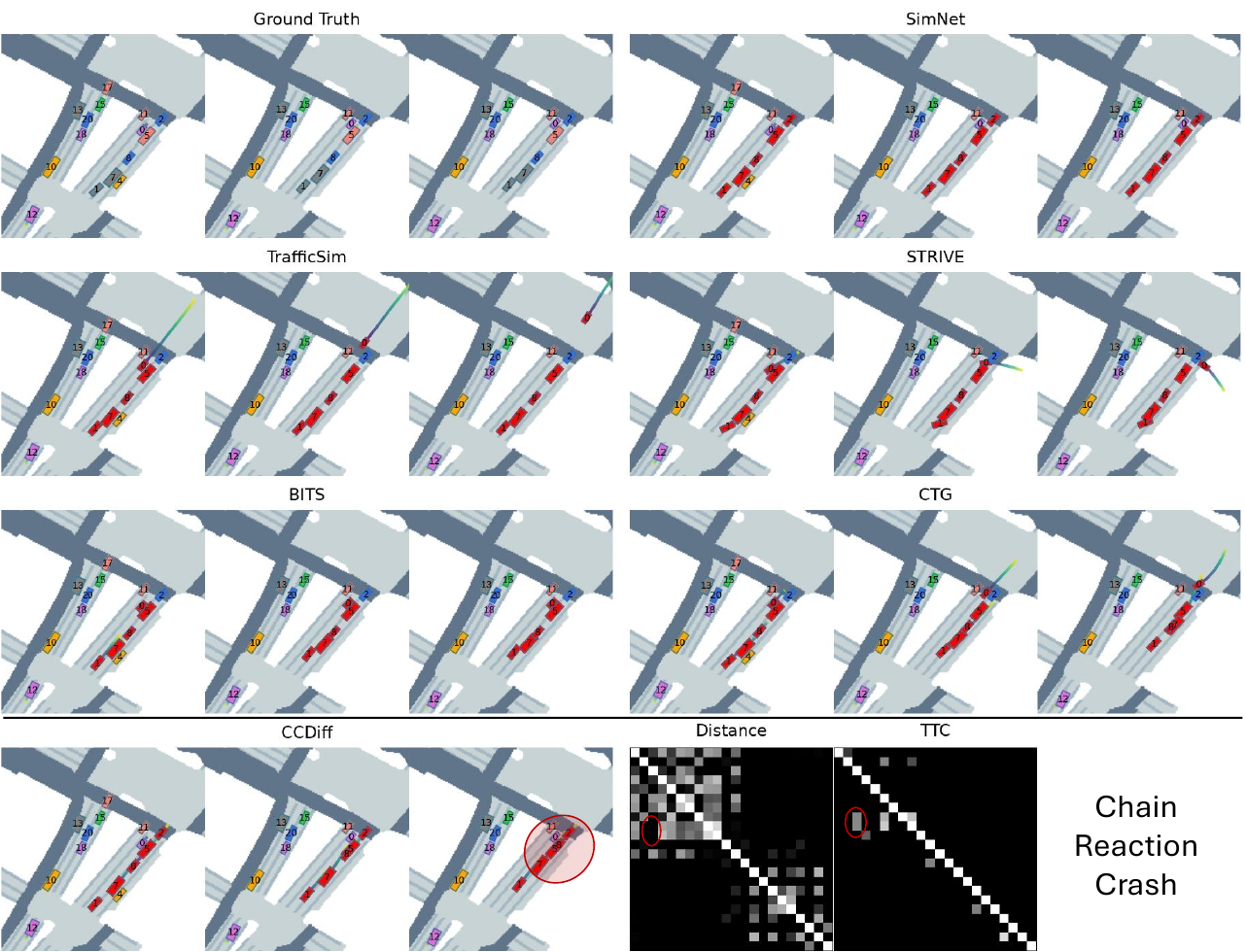}
    \caption{Qualitative of \method\ and baselines in the chain-reaction crash scenarios.}
\end{figure}

\newpage
\paragraph{Multi-agent Generation} We compare the multi-agent generation results of \method\ with CTG. \method\ can consistently generate safety-critical emergency breaking samples when $K\geq 3$, with a control of the most important vehicle 7, 8 in this context. In contrast, CTG keep accelerating the side-lane vehicle 0 or rear vehicle 1 in a very unrealistic way.  
\begin{figure}[htbp]
    \centering
    \includegraphics[width=1.0\linewidth]{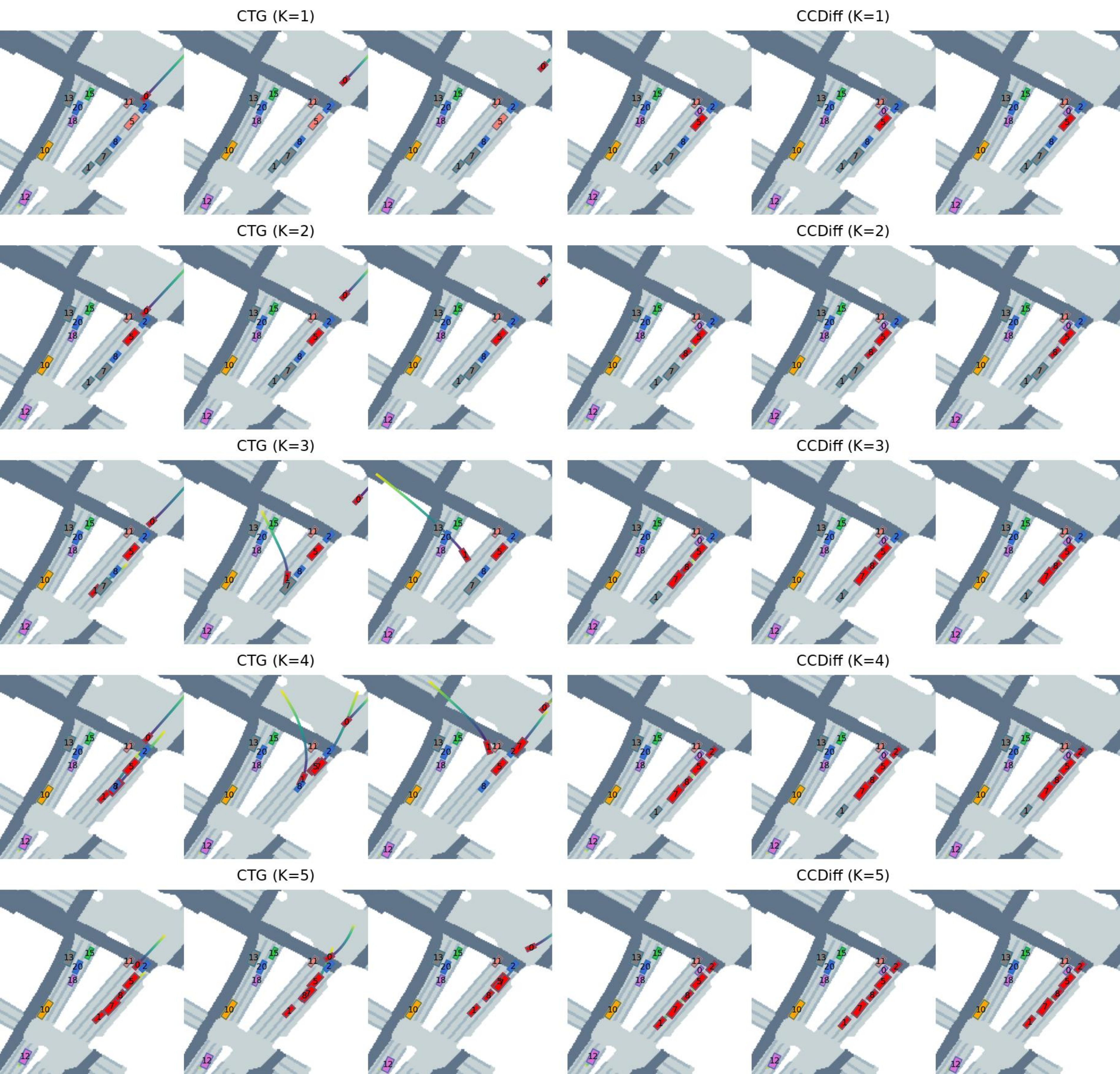}
    \caption{Qualitative comparison of \method\ and CTG under cross traffic violation generation under different sizes of controllable agents. }
    \label{fig:qualitative}
\end{figure}

\newpage
\paragraph{Long-horizon Generation}
We compare the long-horizon generation results of \method\ with CTG. \method\ can consistently generate the cut-in violation scenarios with all different lengths of the generation horizon $1s\leq T\leq 5s$. In contrast, the trajectories generated by CTG seem to diverge by a great deal when $T\geq 2s$. 

\begin{figure}[htbp]
    \centering
    \includegraphics[width=1.0\linewidth]{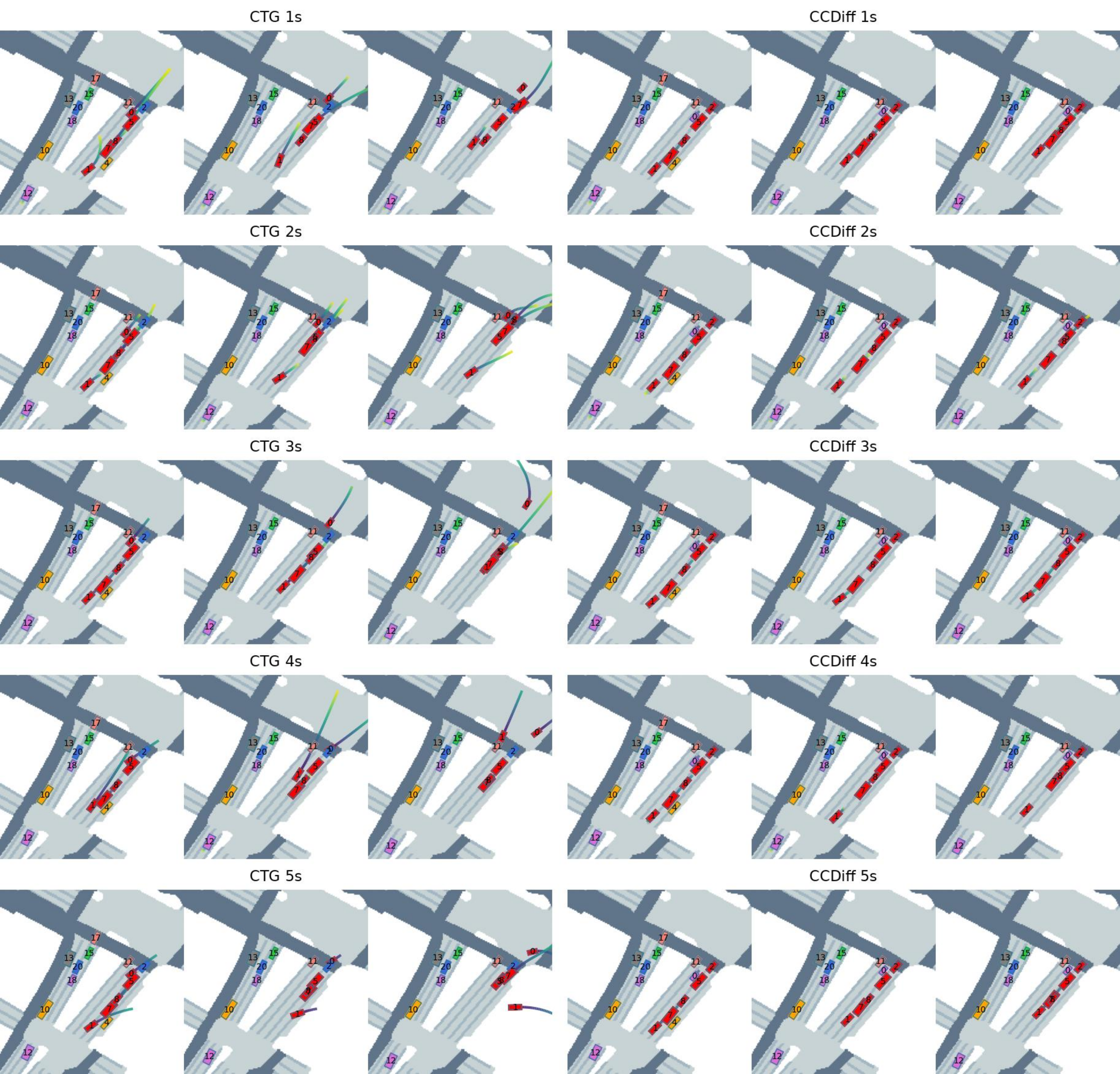}
    \caption{Qualitative comparison of \method\ and CTG under cross traffic violation generation under different generation horizons. }
    \label{fig:qualitative}
\end{figure}

\newpage
\subsubsection{Adjacent Left-turn Side-wipe}
\paragraph{Baseline Comparison}
An adjacent left-turn sideswipe occurs when two vehicles (agent 1, 11) in neighboring left-turn lanes collide as Agent 1 veers into Agent 11's path.

Among all the baselines, STRIVE and CTG generate the motions of 1 and 11 to the straight lane reverse lane.  
TrafficSim generates the motions of 1 and 11 to the straight lane. 
BITS generally follows the original history scenarios with a rear collision between agents 18 and 11. 
CCDiff drifts 1 a little bit and let it veer into the agent 11's path.

Both the Distance graph and the TTC graph could detect the close interaction between agents 1 and 11 in this case. 

\begin{figure}[htbp]
    \centering
    \includegraphics[width=1.0\linewidth]{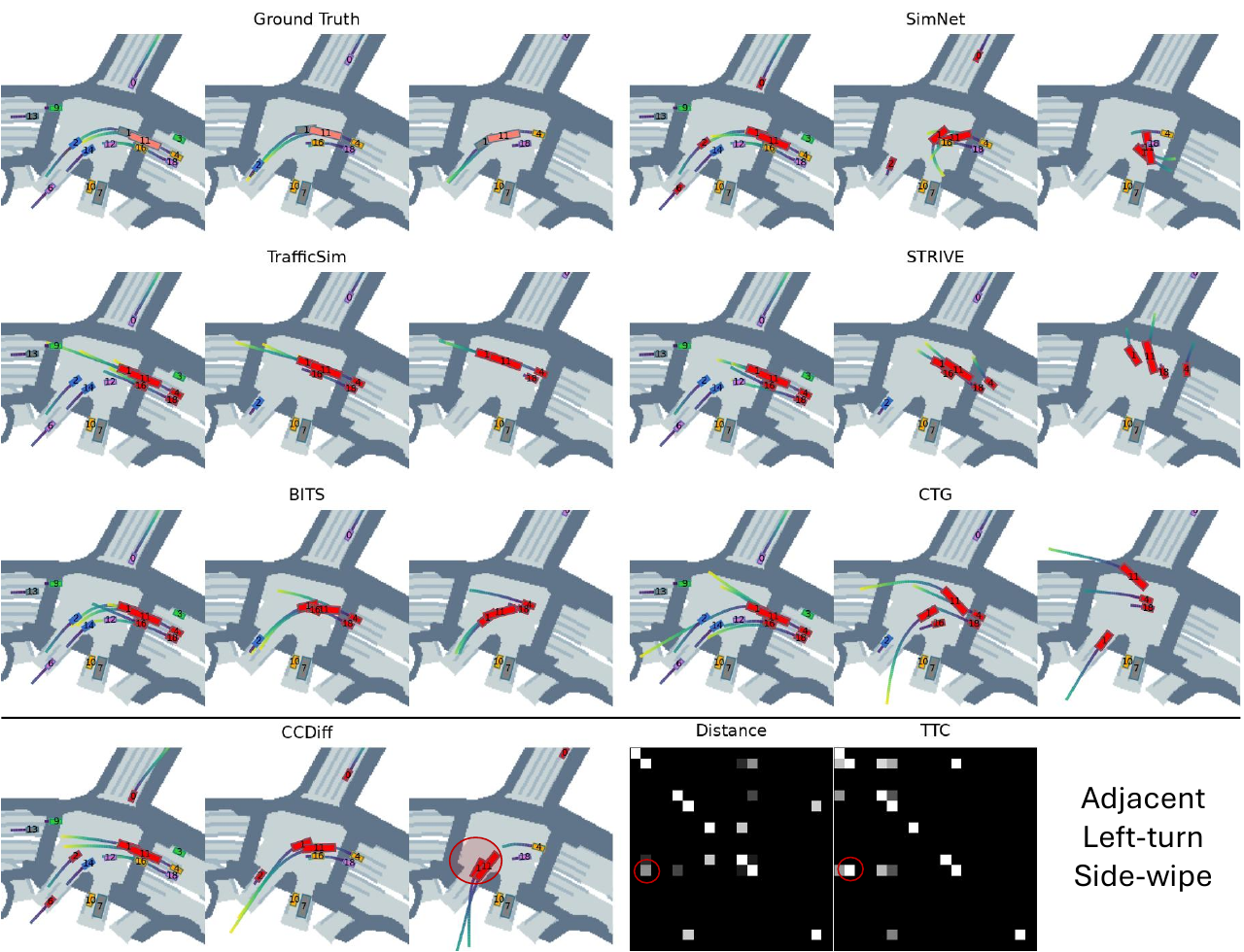}
    \caption{Qualitative of \method\ and baselines in the adjacent left-turn side-wipe scenario. }
\end{figure}

\newpage
\paragraph{Multi-agent Generation} 
We compare the multi-agent generation results of \method\ with CTG. \method\ consistently generates safety-critical emergency braking scenarios when $K\geq 3$, effectively controlling the behavior of the most critical vehicle, agent 1, in this context. In contrast, CTG fails to accurately model the scenario, allowing agent 11 to continue in the wrong direction and being unable to generate collision samples, even when more agents are controllable.
\begin{figure}[htbp]
    \centering
    \includegraphics[width=1.0\linewidth]{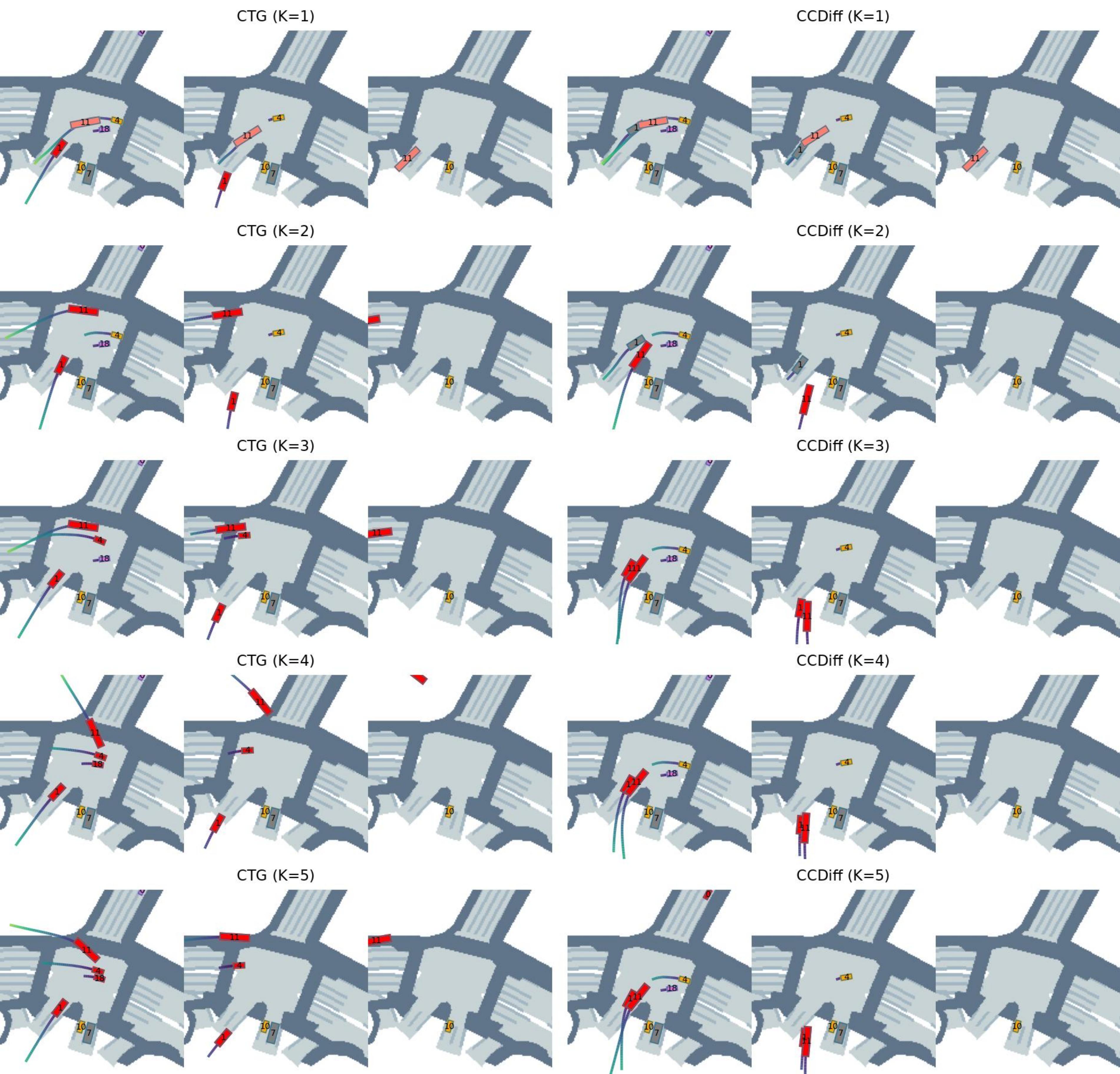}
    \caption{Qualitative comparison of \method\ and CTG under cross traffic violation generation under different sizes of controllable agents. }
    \label{fig:qualitative}
\end{figure}

\newpage
\paragraph{Long-horizon Generation}
We compare the long-horizon generation results of \method\ with CTG. \method\ can consistently generate the left-turn side-wipe scenarios, while CTG diverges and fails to generate collision samples at $T=3s, 4s$. 

\begin{figure}[htbp]
    \centering
    \includegraphics[width=1.0\linewidth]{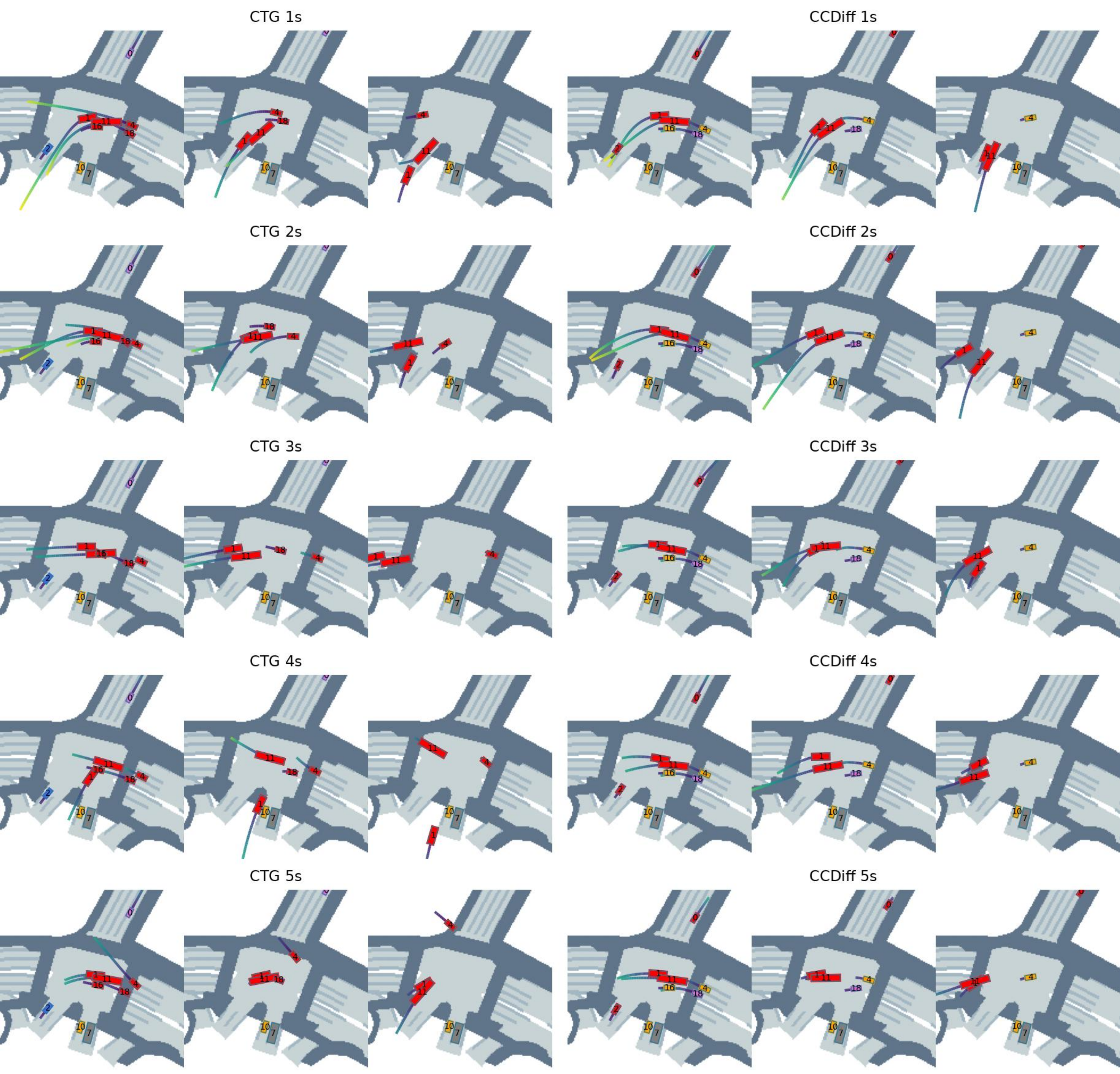}
    \caption{Qualitative comparison of \method\ and CTG under cross traffic violation generation under different generation horizons. }
    \label{fig:qualitative}
\end{figure}

\newpage
\subsubsection{Multi-vehicle Merge-in}
\paragraph{Baseline Comparison}

Multi-vehicle merge-in occurs when a vehicle from a side lane (agent 13) attempts to merge into a single-lane traffic flow (agents 6, 2, 29), causing disruptions or collisions involving three vehicles 2 and 29. 

Among all the baselines, SimNet does not generate collision samples, TrafficSim and CTG generate collision between 13 and 2 and manipulates the trajectory of 13 in an abrupt way. Our scenario just slows down agents 6 and 2 with an expectation of merge-in from agent 13, which causes the trailing agent 29 collides to agent 2. The generated final scenario of CCDiff have the closest layout with the ground-truth trajectories compared to other baselines. 

TTC mask in this case is more sparse with necessary information (agent 2 and 29) compared to the distance mask. 

\begin{figure}[htbp]
    \centering
    \includegraphics[width=1.0\linewidth]{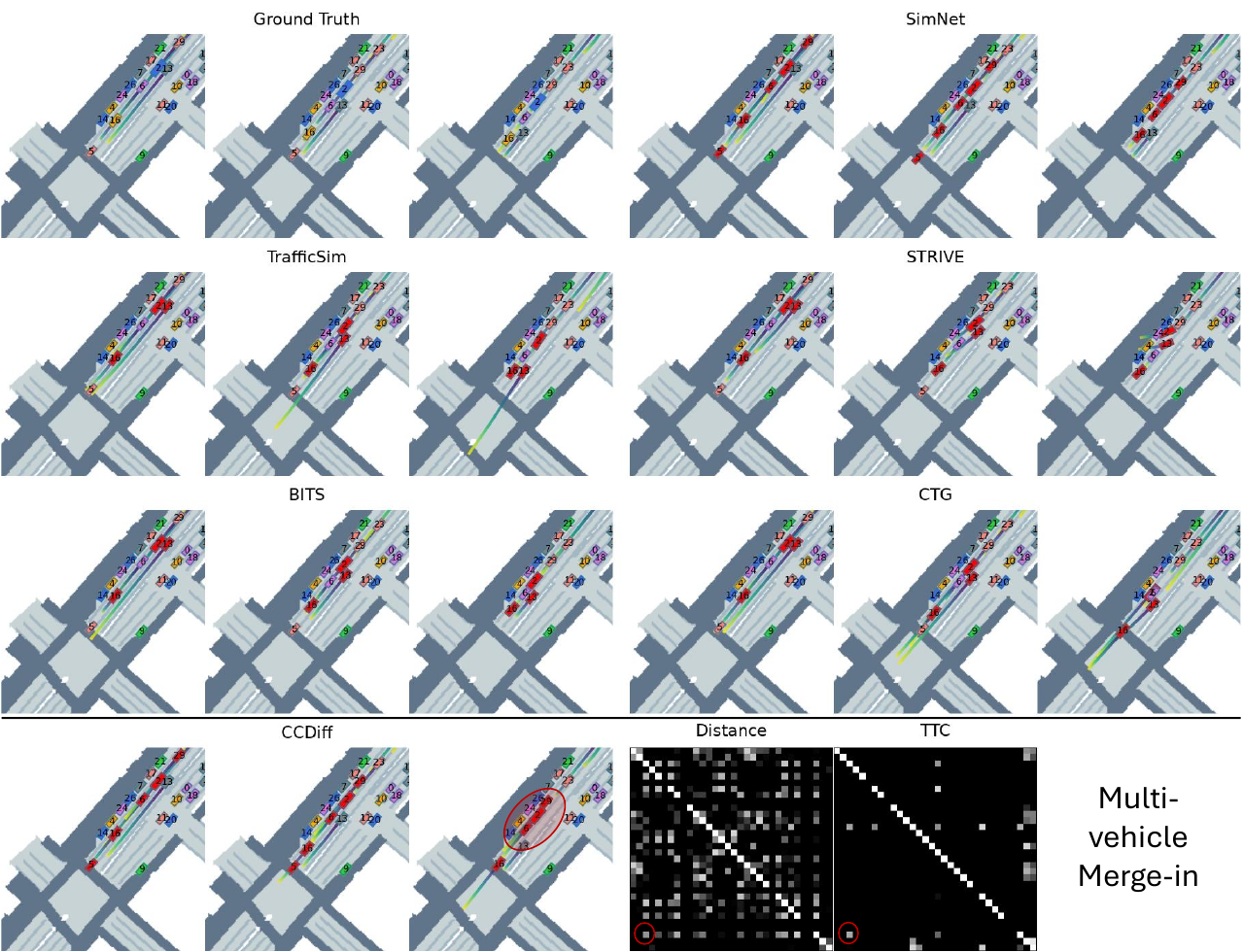}
    \caption{Qualitative of \method\ and baselines in the multi-vehicle lane merge-in scenarios. }
\end{figure}

\newpage
\paragraph{Multi-agent Generation} We compare the multi-agent generation results of \method\ with CTG. \method\ can consistently generate safety-critical emergency breaking samples when $K\geq 4$, with a control of the most important vehicle 2, 6 in this context. In contrast, CTG keeps accelerating the side-lane vehicle 13 without generating any meaningful near-miss samples. 

\begin{figure}[htbp]
    \centering
    \includegraphics[width=1.0\linewidth]{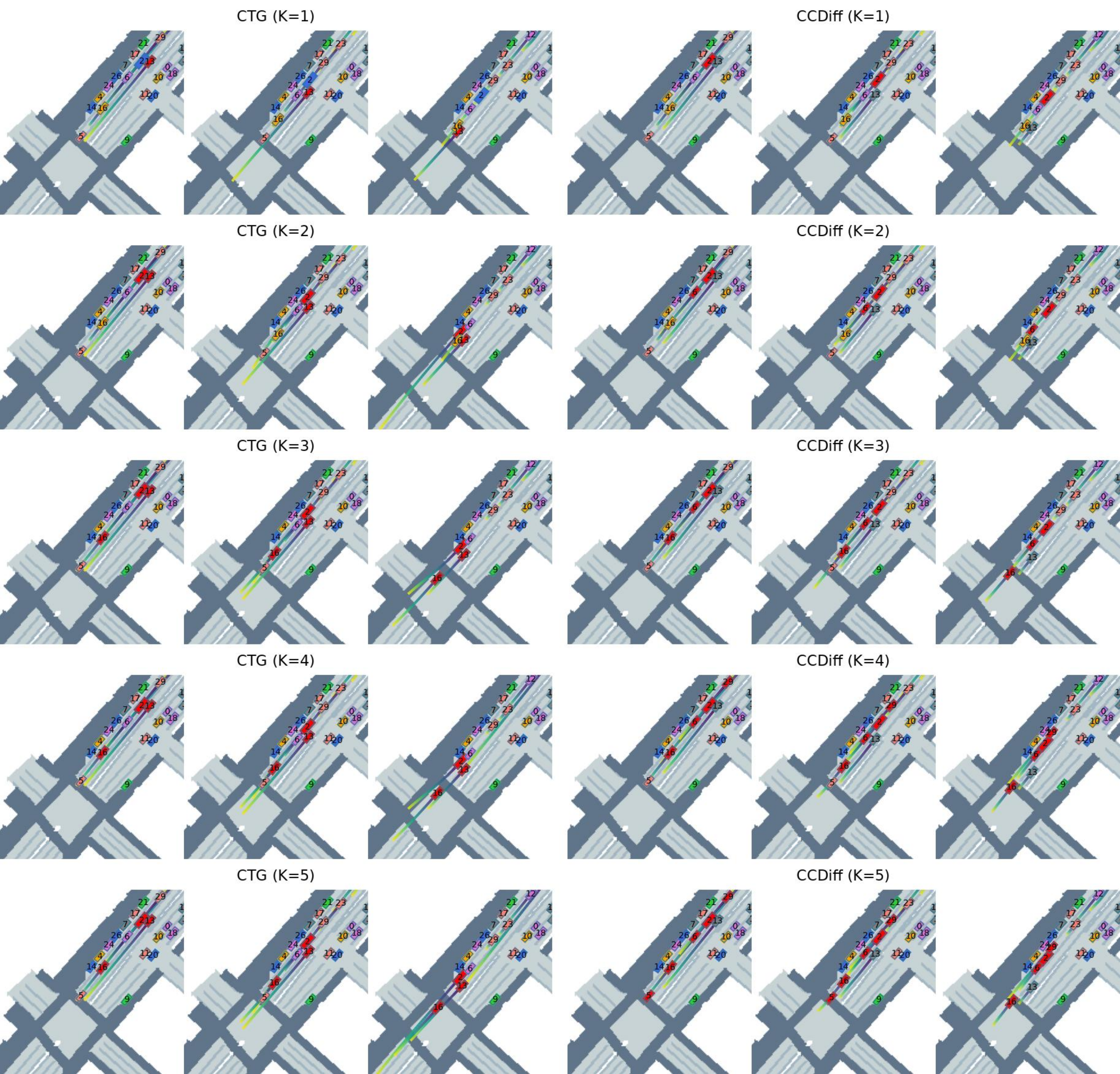}
    \caption{Qualitative comparison of \method\ and CTG under cross traffic violation generation under different sizes of controllable agents. }
    \label{fig:qualitative}
\end{figure}

\newpage
\paragraph{Long-horizon Generation}
We compare the long-horizon generation results of \method\ with CTG. \method\ can consistently generate the multi-vehicle merge-in collision scenarios with all different lengths of the generation horizon $1s\leq T\leq 5s$. In contrast, CTG generates some cut-in collisions between 13 and 6 when $T\geq 2$, which is more unrealistic given the ground-truth layouts. 

\begin{figure}[htbp]
    \centering
    \includegraphics[width=1.0\linewidth]{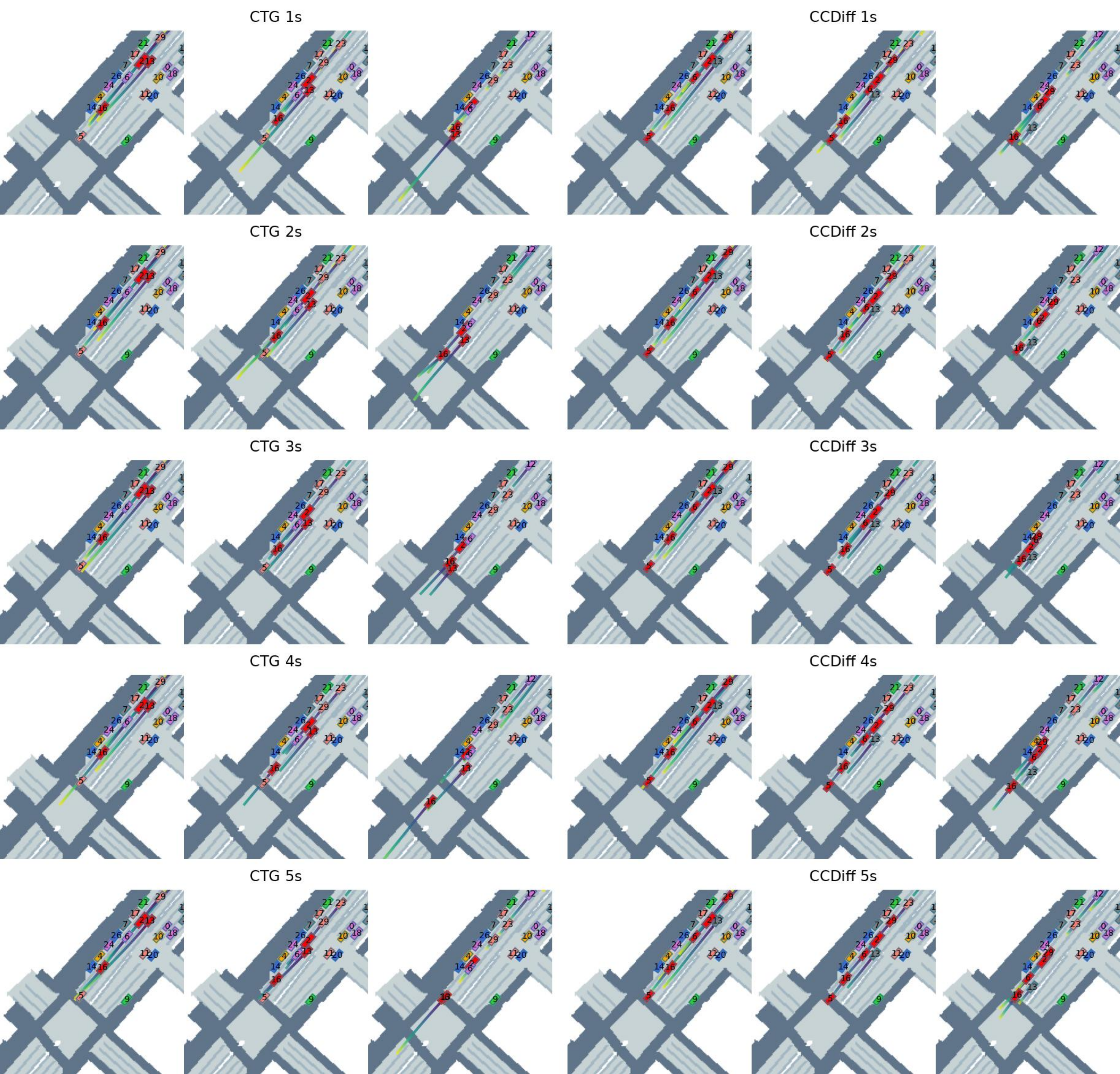}
    \caption{Qualitative comparison of \method\ and CTG under cross traffic violation generation under different generation horizons. }
    \label{fig:last-qualitative}
\end{figure}

\end{document}